\documentclass{article} 
\PassOptionsToPackage{numbers, compress}{natbib}
\usepackage{iclr2022_conference,times}

\usepackage[utf8]{inputenc} 
\usepackage[T1]{fontenc}    
\usepackage{url}            
\usepackage{hyperref}       
\usepackage{booktabs}       
\usepackage{amsfonts}       
\usepackage{nicefrac}       
\usepackage{microtype}      
\usepackage{xcolor}         

\usepackage{support-caption}
\usepackage{subcaption}
\usepackage{siunitx}
\usepackage{amsmath}
\usepackage{cleveref}
\input{llymacro}
\usepackage{xspace}
\usepackage{booktabs}
\usepackage{multirow}
\usepackage{amssymb}
\usepackage{colortbl}
\usepackage{wrapfig}
\usepackage{version}
\usepackage[normalem]{ulem}
\definecolor{tabgray}{gray}{0.90}
\usepackage{pdfpages}
\newcount\Comments  
\usepackage{color}
\definecolor{darkgreen}{rgb}{0,0.5,0}
\definecolor{darkblue}{rgb}{0,0,0.5}
\definecolor{purple}{rgb}{1,0,1}
\definecolor{gray}{rgb}{0.5,0.5,0.5}
\newcommand{\kibitz}[2]{\ifnum\Comments=0\textcolor{#1}{#2}\fi}

\ifnum\Comments=0\else\excludeversion{old}\fi
\ifnum\Comments=0\newenvironment{new}{\color{blue}}{}\else\fi

\usepackage{nicefrac}
\usepackage{enumitem}
\crefname{fact}{Fact}{Facts}

\def\Mweight{{\mathcal{M}_{\mathrm{WE}}}}
\def\Mmme{{\mathcal{M}_{\mathrm{MME}}}}
\def\dif{\rd}

\def\Lgd{\gL_{\mathrm{GD}}}
\def\Lcm{\gL_{\mathrm{CM}}}

\newcommand\approach{Diversity Regularized Training\xspace}
\newcommand\boldApproach{\textbf{D}iversity-\textbf{R}egularized \textbf{T}raining\xspace}
\newcommand\shortApproach{DRT\xspace}

\newcommand\ourEnsemble{Max-Margin Ensemble\xspace}
\newcommand\shortOurEnsemble{MME\xspace}
\newcommand\weightedEnsemble{Weighted Ensemble\xspace}
\newcommand\shortWeightedEnsemble{WE\xspace}

\newcommand\beforeProtocol{\textit{Ensemble-before-Smoothing}\xspace}
\newcommand\shortBeforeProtocol{EBS\xspace}
\newcommand\afterProtocol{\textit{Ensemble-after-Smoothing}\xspace}
\newcommand\shortAfterProtocol{EAS\xspace}

\newcommand\LHSloss{Gradient Diversity Loss\xspace}
\newcommand\shortLHSloss{GD Loss\xspace}
\newcommand\RHSloss{Confidence Margin Loss\xspace}
\newcommand\shortRHSloss{CM Loss\xspace}

\renewcommand{\paragraph}[1]{\textbf{#1}\quad}

\title{On the Certified Robustness for Ensemble Models and Beyond}
\author{Zhuolin Yang\textsuperscript{1}$^*$\quad  Linyi Li\textsuperscript{1}$^*$ \quad Xiaojun Xu\textsuperscript{1}\quad Bhavya Kailkhura\textsuperscript{2}\quad Tao Xie\textsuperscript{3}\quad Bo Li\textsuperscript{1}\\
\textsuperscript{1}University of Illinois Urbana-Champaign \\
\textsuperscript{2}Lawrence Livermore National Laboratory \quad \textsuperscript{3}Peking University\\
{\texttt{\{zhuolin5,linyi2,xiaojun3,lbo\}@illinois.edu}}  \\
 {\texttt{kailkhura1@llnl.gov}}
 \quad {\texttt{taoxie@pku.edu.cn}}\\
$^*$ Equal contribution
}

\iclrfinalcopy 
\begin{document}

\maketitle

\begin{abstract}
    Recent studies show that deep neural networks (DNN) are vulnerable to adversarial examples, which aim to mislead DNNs by adding perturbations with small magnitude. 
    To defend against such attacks, both empirical and theoretical defense approaches have been extensively studied for a \textit{single ML model}.
    In this work, we aim to analyze and provide the certified robustness for \textit{ensemble ML models}, together with the sufficient and necessary conditions of robustness for different ensemble protocols.
    Although ensemble models are shown more robust than a single model empirically;
    surprisingly, we find that in terms of the certified robustness the standard ensemble models only achieve marginal improvement compared to a single model.
    Thus, to explore the conditions that guarantee to provide certifiably robust ensemble ML models, we first prove that 
    \textit{diversified gradient} and \textit{large confidence margin} are sufficient and necessary conditions for certifiably robust ensemble models under the model-smoothness assumption.
    We then provide the bounded model-smoothness analysis based on the proposed \beforeProtocol strategy. We also prove
    that an ensemble model can \textit{always} achieve higher certified robustness than a single base model under mild conditions.
    Inspired by the theoretical findings, we propose the lightweight \approach~(\shortApproach) to train  certifiably robust ensemble ML models.
    Extensive experiments show that our \shortApproach enhanced ensembles can consistently achieve higher certified robustness than existing single and ensemble ML models, demonstrating the state-of-the-art certified $L_2$-robustness on MNIST, CIFAR-10, and ImageNet datasets.

\end{abstract}

\section{Introduction}
    \vspace{-1em}

    Deep neural networks~(DNN) have been widely applied in various applications, such as image classification~\citep{krizhevsky2009learning,he2016deep}, face recognition~\citep{sun2014deep}, and natural language processing~\citep{vaswani2017attention,devlin2019bert}.
    However, it is well-known that DNNs are vulnerable to adversarial examples~\citep{szegedy2013intriguing,carlini2017towards,xiao2018generating,xiao2018spatially,bhattad2019unrestricted, bulusu2020anomalous}, and it has raised great concerns especially when DNNs are deployed in safety-critical applications such as autonomous driving and facial recognition.
    
    \looseness=-1
    To defend against such attacks, several empirical defenses have been proposed~\citep{papernot2016distillation,madry2018towards}; however, many of them have been attacked again by strong adaptive attackers~\citep{athalye2018obfuscated,tramer2020adaptive}.
    To end such repeated game between the attackers and defenders, \textit{certified} defenses~\citep{wong2018provable,cohen2019certified} have been proposed to provide the robustness guarantees for given ML models, so that no additional attack can break the model under certain adversarial constraints.
    For instance, randomized smoothing has been proposed as an effective defense providing certified robustness~\citep{lecuyer2019certified,cohen2019certified,yang2020randomized}.
    Among different certified robustness approaches~\citep{weng2018towards,xu2020automatic,li2020sok,zhang2022boosting}, randomized smoothing provides a model-independent way to smooth a given ML model and achieves state-of-the-art certified robustness on large-scale datasets such as ImageNet.
    
    Currently, all the existing certified defense approaches  focus on the robustness of a single ML model. Given the observations that  ensemble ML models are able to bring additional benefits in standard learning~\citep{opitz1999popular,rokach2010ensemble}, in this work we aim to ask: \textit{Can an ensemble ML model provide additional benefits in terms of the certified robustness compared with a single model?} \textit{If so, what are the sufficient and necessary conditions to guarantee such certified robustness gain?} 
    
    Empirically, we first find that \textit{standard} ensemble models only achieve {marginally} higher certified robustness by directly appling randomized smoothing: 
    with $L_2$ perturbation radius $1.5$, a single model achieves certified accuracy as $21.9\%$, while the average aggregation based ensemble of three models achieves certified accuracy as $24.2\%$ on CIFAR-10~(\Cref{tab:cifartable}).
    Given such observations, next we aim to answer:
    \textit{How to improve the certified robustness of ensemble ML models? What types of conditions are required to improve the certified robustness for ML ensembles?}
    
    In particular, from the theoretical perspective, 
    we analyze the standard \weightedEnsemble~(\shortWeightedEnsemble) and \ourEnsemble~(\shortOurEnsemble) protocols, and prove the sufficient and necessary conditions for the certifiably robust ensemble models under model-smoothness assumption.
    Specifically, we prove that:
    (1)~an ensemble ML model is more certifiably robust than each single base model; 
    (2)~\textit{diversified gradients} and \textit{large confidence margins} of base models are the sufficient and necessary conditions for the certifiably robust ML ensembles. We show that these two key factors
    would lead to higher certified robustness for ML ensembles.
    We further propose \beforeProtocol as the model smoothing strategy and prove the bounded model-smoothness with such strategy, which realizes our model-smoothness assumption.

    
\begin{wrapfigure}{r}{0.45\textwidth}
      \vspace{-5.7mm}
  \begin{center}
    \includegraphics[width=0.37\textwidth]{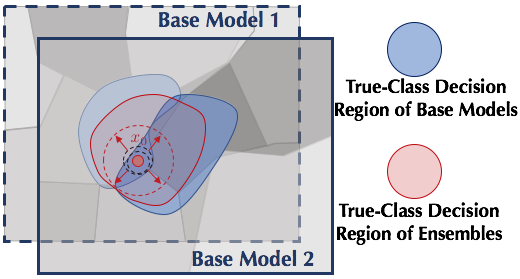}
  \end{center}
  \vspace{-4mm}
  \caption{\small Illustration of a robust ensemble.}
  \label{fig1}
  \vspace{-2.7mm}
\end{wrapfigure}

    Inspired by our theoretical analysis, we propose \boldApproach~(\shortApproach), a lightweight regularization-based ensemble training approach.
    \shortApproach is composed of two simple yet effective and general regularizers to promote the diversified gradients and large confidence margins respectively.
    \shortApproach can be easily combined with existing ML approaches for training smoothed models, such as Gaussian augmentation~\citep{cohen2019certified} and adversarial smoothed training~\citep{salman2019provably}, with negligible training time overhead while achieves \textit{significantly} higher certified robustness than state-of-the-art approaches consistently.
    
    
    We conduct extensive experiments on a wide range of datasets including MNIST, CIFAR-10, and ImageNet.
    The experimental results show that \shortApproach can achieve significantly higher certified robustness compared to baselines with similar training cost as training a single model.
    Furthermore, as \shortApproach is flexible to integrate any base models, by
    using the pretrained robust single ML models as base models, \shortApproach 
    achieves the highest certified robustness so far to our best knowledge.
    For instance, on CIFAR-10 under $L_2$ radius $1.5$, the \shortApproach-trained  ensemble with three base models improves the certified accuracy from SOTA $24.2\%$  to $30.3\%$; and under $L_2$ radius
    $2.0$, \shortApproach improves the certified accuracy from  SOTA $16.0\%$  to $20.3\%$.
    
{\bf \underline{Technical Contributions.} }
In this paper, we conduct the \textit{first} study for the sufficient and necessary conditions of certifiably robust ML ensembles and propose an efficient training algorithm \shortApproach to achieve the state-of-the-art certified robustness.
We make contributions on both theoretical
and empirical fronts.  
\begin{itemize}[leftmargin=0.8cm,itemsep=-0.5mm]
        \vspace{-0.8em}
        \item 
        We provide the \textit{necessary} and \textit{sufficient} conditions for robust ensemble ML models including \weightedEnsemble~(\shortWeightedEnsemble) and \ourEnsemble~(\shortOurEnsemble) under the model-smoothness assumption.
        In particular, we prove that the \textit{diversified gradients} and \textit{large confidence margins} of base models are the sufficient and necessary conditions of certifiably robust ensembles.
        We also prove the bounded model-smoothness via proposed \beforeProtocol strategy, which realizes our model-smoothness assumption.
        \item To analyze different ensembles, we prove that when the adversarial transferability among base models is low, \shortWeightedEnsemble is more robust than \shortOurEnsemble. We also prove that the ML ensemble is more robust than a single base model under the model-smoothness assumption.
        \item
        Based on the theoretical analysis of the sufficient and necessary conditions, we propose \shortApproach, a lightweight regularization-based training approach that can be easily combined with different training approaches and ensemble protocols with small training cost overhead.
        \item
        We conduct extensive experiments to evaluate the effectiveness of \shortApproach on various datasets, and we show that to the best of your knowledge, \shortApproach can achieve the \textit{highest} certified robustness, outperforming all existing baselines.
    \end{itemize}
    \vspace{-0.5em}
    
    \noindent \paragraph{Related work.}
    DNNs are known vulnerable to adversarial examples~\citep{szegedy2013intriguing}.
    To defend against such attacks,  several empirical defenses have been proposed~\citep{papernot2016distillation,madry2018towards}.
    For ensemble models, existing work mainly focuses on empirical robustness~\citep{pang2019improving,li2020boosting,cheng2021mixture}
        where the robustness is measured by accuracy under existing attacks and no certified robustness guarantee could be provided or enhanced;
    or certify the robustness for a standard weighted ensemble~\citep{zhang2019enhancing,liu2020enhancing}
        using either LP-based~\citep{zhang2018efficient} verification or randomized smoothing without considering the model diversity~\citep{liu2020enhancing} to boost their certified robustness.
    In this paper, we aim to prove that the diversified gradient and large confidence margin are the sufficient and necessary conditions for certifiably robust ensemble ML models.
    Moreover, to our best knowledge, we propose the \textit{first} training approach to boost the \textit{certified} robustness of ensemble~ML~models.
    
    Randomized smoothing~\citep{lecuyer2019certified,cohen2019certified} has been proposed to provide certified robustness for a single ML model.
        It achieved the state-of-the-art certified robustness on large-scale dataset such as ImageNet and CIFAR-10 under $L_2$ norm.
    Several approaches have been proposed to further improve it by:
    (1)~choosing different smoothing distributions for different $L_p$ norms~\citep{dvijotham2019framework,zhang2020black,yang2020randomized},
    and (2)~training more robust smoothed classifiers, using data augmentation~\citep{cohen2019certified}, unlabeled data~\citep{carmon2019unlabeled}, adversarial training~\citep{salman2019provably}, regularization~\citep{li2019certified,zhai2019macer}, and denoising~\citep{salman2020black}.
        In this paper, we compare and propose a suitable  smoothing strategy to improve the certified robustness of ML ensembles.
        
    
    \vspace{-0.75em}
\section{Characterizing ML Ensemble Robustness}
    \vspace{-0.75em}
    \label{sec:2}
    In this section, we prove the sufficient and necessary robustness conditions for both general  and smoothed ML ensemble models.
    Based on these robustness conditions, we discuss the key factors for improving the certified robustness of an ensemble, compare the robustness of ensemble models with single models, and outline several findings based on  additional theoretical analysis.
    
        \vspace{-0.25em}
    \subsection{Preliminaries}
        \vspace{-0.25em}
        \label{subsec:prelim}
        \paragraph{Notations.}
        Throughout the paper, we consider the classification task with $C$ classes.
        We first define the classification scoring function $f: \sR^d \to \vDelta^C$, which maps the input to a \emph{confidence vector}, and $f(\vx)_i$ represents the confidence for the $i$th class.
        We mainly focus on the confidence after normalization, i.e., $f(\vx) \in \vDelta^C = \{\vp \in \sR^C_{\ge 0}: \|\vp\|_1 = 1\}$  in the probability simplex.
        To characterize the \emph{confidence margin} between two classes, we define $f^{y_1/y_2}(\vx) := f(\vx)_{y_1} - f(\vx)_{y_2}$.
        The corresponding \emph{prediction} $F: \sR^d \to [C]$ is defined by $F(\vx) := \argmax_{i\in [C]} f(\vx)_i$.
        We are also interested in the \emph{runner-up prediction} 
        $F^{(2)}(\vx) := \argmax_{i\in [C]: i \neq F(\vx)} f(\vx)_i$.
        
        \paragraph{$r$-Robustness.}
        For brevity, we consider the model's certified robustness, against the $L_2$-bounded perturbations as defined below.
        Our analysis can be generalizable for $L_1$ and $L_\infty$ perturbations, leveraging existing work~\citep{li2019certified,yang2020randomized,levine2021improved}.
        \begin{definition}[$r$-Robustness]
            For a prediction function $F: \sR^d \to [C]$ and input $\vx_0$,  if all instance $\vx \in \{\vx_0 + \vdelta: \|\vdelta\|_2 {\color{blue} <} r\}$  satisfies $F(\vx) = F(\vx_0)$,  we say model $F$ is \emph{$r$-robust} (at point $\vx_0$).
            \label{def:r-robustness}
        \end{definition}
        
        \paragraph{Ensemble Protocols.}
        An ensemble model contains $N$ \emph{base models} $\{F_i\}_{i=1}^N$, where $F_i(\vx)$ and $F^{(2)}_i(\vx)$ are their top and runner-up predictions for given input $\vx$ respectively.
        The ensemble prediction is denoted by $\gM: \sR^d \to [C]$, which is computed based on outputs of base models following certain ensemble protocols.
        In this paper, we consider both Weighted Ensemble~(\shortWeightedEnsemble) and Maximum Margin Ensemble~(\shortOurEnsemble).
        
            \begin{definition}[\weightedEnsemble~(\shortWeightedEnsemble)]
                \label{def:weighted-ensemble}
                Given $N$ base models $\{F_i\}_{i=1}^N$, and the weight vector $\{w_i\}_{i=1}^N \in \sR^N_+$,
                the weighted ensemble $\Mweight$: $\sR^d \to [C]$ is defined by
                \begin{equation}
                    \small
                    \Mweight(\vx_0) := \argmax_{i\in [C]} \sum_{j=1}^N w_j f_j(\vx_0)_i.
                    \label{eq:weighted-ensemble}
                \end{equation}
            \end{definition}
            
            \begin{definition}[\ourEnsemble~(\shortOurEnsemble)]
                \label{def:maximum-margin-ensemble}
                Given $N$ base models $\{F_i\}_{i=1}^N$, for input $\vx_0$, the max-margin ensemble model $\Mmme: \sR^d \to [C]$ is defined by
                \begin{equation}
                        \small
                        \Mmme(\vx_0) := F_c(\vx_0) \quad 
                        \mathrm{where} \quad 
                        c = \argmax_{i \in [N]} \left( f_i(\vx_0)_{F_i(\vx_0)} - f_i(\vx_0)_{F_i^{(2)}(\vx_0)} \right).
                    \label{eq:maximum-margin-ensemble}
                \end{equation}
            \end{definition}
        
        The commonly-used \shortWeightedEnsemble~\citep{zhang2019enhancing,liu2020enhancing} sums up the weighted confidence  of base models $\{F_i\}_{i=1}^N$ with weight vector $\{w_i\}_{i=1}^N$, and predicts the class with the highest weighted confidence.
        The standard average ensemble can be viewed as a special case of \shortWeightedEnsemble~(where all $w_i$'s are equal).
        \shortOurEnsemble chooses the base model with the largest confidence margin between the top and the runner-up classes, which is a direct extension from max-margin training~\citep{huang2008maxi}.
        
        \paragraph{Randomized Smoothing.}
        Randomized smoothing~\citep{lecuyer2019certified,cohen2019certified} provides certified robustness by constructing a smoothed model from a given model.
        Formally, let $\varepsilon\sim \gN(0, \sigma^2 \mI_d)$ be a Gaussian random variable,
        for any given model $F: \sR^d \to [C]$~(can be an ensemble), we define \textit{smoothed confidence} function $g^\varepsilon_F: \sR^d \to \vDelta^C$ such that
        \begin{equation}
            \label{def:smoothed-conf-g}
            g^\varepsilon_F(\vx)_j := \underset{\varepsilon\sim\gN(0,\sigma^2 \mI_d)}{\E} \1[F(\vx + \varepsilon)=j] = \Pr_{\varepsilon\sim\gN(0,\sigma^2 \mI_d)} (F(\vx + \varepsilon) = j).
        \end{equation}
        Intuitively, $g^\varepsilon_F(\vx)_j$ is the probability of base model $F$'s prediction on the $j$th class given Gaussian smoothed input.
        The smoothed classifier $G^\varepsilon_F: \sR^d \to [C]$ outputs the class with highest smoothed confidence: $G^\varepsilon_F(\vx) := \argmax_{j\in [C]} g^\varepsilon_F(\vx)_j$.
        Let $c_A$ be the predicted class for input $\vx_0$, i.e., $c_A := G^\varepsilon_F(\vx_0)$.
        \citeauthor{cohen2019certified} show that  $G^\varepsilon_F$ is $(\sigma \Phi^{-1}(g^\varepsilon_F(\vx_0)_{c_A}))$-robust at input $\vx_0$, i.e., the certified radius is $\sigma \Phi^{-1}(g^\varepsilon_F(\vx_0)_{c_A})$ where $\Phi^{-1}$ is the inverse cumulative distribution function of standard normal distribution.
        In practice, we will leverage the smoothing strategy together with Monte-Carlo sampling to certify ensemble robustness.
        More details can be found in \Cref{adx:rand-smooth-background}.

    \subsection{Robustness Conditions for General Ensemble Models}
        \label{subsec:2-2}
       We will first provide sufficient and necessary conditions for robust ensembles under the model-smoothness assumption.
        
        \begin{definition}[$\beta$-Smoothness]
            \label{def:beta-smooth}
            A differentiable function $f:\,\sR^d \mapsto \sR^C$ is $\beta$-smooth, if for any $\vx_1,\,\vx_2 \in \sR^d$ and any output dimension $j \in [C]$,
            $
                \frac{\| \nabla_{\vx_1} f(\vx_1)_j - \nabla_{\vx_2} f(\vx_2)_j\|_2}{\|\vx_1 - \vx_2\|_2} \le \beta.
            $
        \end{definition}
                The definition of $\beta$-smoothness is inherited from optimization theory literature, and it is equivalent to the curvature bound in certified robustness literature~\citep{singla2020second}.
                $\beta$ quantifies the non-linearity of function $f$, where higher $\beta$ indicates more rigid functions/models and smaller $\beta$ indicates smoother ones.
                When $\beta = 0$ the function/model is linear.
        
        For \weightedEnsemble~(\shortWeightedEnsemble), we have the following robustness conditions.
        \begin{theorem}[Gradient and Confidence Margin Conditions for \shortWeightedEnsemble Robustness]
            \label{thm:gradient-based-sufficient-necessary-cond-weight-ensemble}
            Given input $\vx_0 \in \sR^d$ with ground-truth label $y_0 \in [C]$, and $\Mweight$ as a \shortWeightedEnsemble defined over base models $\{F_i\}_{i=1}^N$ with weights $\{w_i\}_{i=1}^N$.
            $\Mweight(\vx_0) = y_0$.
            All base models $F_i$'s are $\beta$-smooth.
            \begin{itemize}
                \vspace{-0.5em}
                \item (Sufficient Condition) The $\Mweight$ is $r$-robust at point $\vx_0$ if for any $y_i \neq y_0$,
                \begin{equation}
                    \small
                    \Big\|\sum_{j=1}^N w_j \nabla_\vx f_j^{y_0/y_i}(\vx_0) \Big\|_2 \le \frac 1 r \sum_{j=1}^N w_j f_j^{y_0/y_i}(\vx_0) - \beta r \sum_{j=1}^N w_j,
                    \label{eq:gradient-based-sufficient-cond-weight-ensemble}
                    \vspace{-1em}
                \end{equation}
                \item (Necessary Condition) If $\Mweight$ is $r$-robust at point $\vx_0$,  for any $y_i \neq y_0$,
                \begin{equation}
                    \small
                    \Big\|\sum_{j=1}^N w_j \nabla_\vx f_j^{y_0/y_i}(\vx_0) \Big\|_2 \le \frac 1 r \sum_{j=1}^N w_j f_j^{y_0/y_i}(\vx_0) + \beta r \sum_{j=1}^N w_j.
                    \label{eq:gradient-based-necessary-cond-weight-ensemble}
                    \vspace{-1em}
                \end{equation}
            \end{itemize}
        \end{theorem}
        The proof follows from Taylor expansion at $\vx_0$ and we leave the detailed proof in \Cref{subsec:a-2}.
        When it comes to \ourEnsemble~(\shortOurEnsemble), the derivation of robust conditions is more involved.
        In \Cref{thm:gradient-based-sufficient-necessary-cond-our-ensemble}~(\Cref{subsec:a-1-1}) we derive the robustness conditions for \shortOurEnsemble composed of two base models.
        The robustness conditions have highly similar forms as those for \shortWeightedEnsemble in \Cref{thm:gradient-based-sufficient-necessary-cond-weight-ensemble}.
        Thus, for brevity, we focus on discussing \Cref{thm:gradient-based-sufficient-necessary-cond-weight-ensemble} for \shortWeightedEnsemble hereinafter and similar conclusions  can be drawn for \shortOurEnsemble~(details are in \Cref{subsec:a-1-1}).
        
        To analyze \Cref{thm:gradient-based-sufficient-necessary-cond-weight-ensemble}, we define \emph{Ensemble Robustness Indicator}~(ERI) as such:
        \vspace{-0.5em}
        \begin{equation}
            \small
            I_{y_i} := \Big\|\sum_{j=1}^N w_j \nabla_\vx f_j^{y_0/y_i}(\vx_0) \Big\|_2 \Big/ \| \vw \|_1 
            -
            \frac{1}{r \|\vw\|_1} \sum_{j=1}^N w_j f_j^{y_0/y_i}(\vx_0).
            \label{eq:I}
            \vspace{-0.45em}
        \end{equation}
        ERI appears in both sufficient~(\Cref{eq:gradient-based-sufficient-cond-weight-ensemble}) and necessary~(\Cref{eq:gradient-based-necessary-cond-weight-ensemble}) conditions.
        In both conditions, \emph{smaller} ERI means \emph{more} certifiably robust ensemble.
        Note that we can analyze the robustness under different attack radius $r$ by directly varying $r$ in \Cref{eq:gradient-based-sufficient-cond-weight-ensemble,eq:gradient-based-necessary-cond-weight-ensemble}.
        When $r$ becomes larger, the gap between the RHS of two inequalities~($2\beta r \sum_{j=1}^N w_j$) also becomes larger, and thus it becomes harder to determine robustness via \Cref{thm:gradient-based-sufficient-necessary-cond-weight-ensemble}.
        This is because the first-order condition implied by \Cref{thm:gradient-based-sufficient-necessary-cond-weight-ensemble} becomes coarse when $r$ is large. 
        However, due to bounded $\beta$ as we will show, the training approach motivated by the theorem still empirically works well under large $r$. 
        
        \paragraph{Diversified Gradients.}
        The core of first term in ERI is the magnitude of the vector sum of gradients: $\|\sum_{j=1}^N w_j \nabla_\vx f_j^{y_0/y_i}(\vx_0) \|_2$.
        According to the law of cosines: $\|\va + \vb\|_2 = \sqrt{\|\va\|_2^2 + \|\vb\|_2^2 + 2\|\va\|_2\|\vb\|_2 \cos\langle \va, \vb\rangle}$,
        to reduce this term, we could either reduce the base models' gradient magnitude or diversify their gradients~(in terms of cosine similarity).
        Since simply reducing base models' gradient magnitude would hurt model expressivity~\citep{huster2018limitations}, during regularization the main functionality of this term would be promoting diversified gradients.
        
        \paragraph{Large Confidence Margins.}
        The core of second term in ERI is the confidence margin: $\sum_{j=1}^N w_j f_j^{y_0/y_i}(\vx_0)$.
        Due to the negative sign of second term in ERI, we need to increase this term, i.e., we need to increase confidence margins to achieve higher ensemble robustness.
        
        In summary, the diversified gradients and large confidence margins are the sufficient and necessary conditions for high certified robustness of ensembles.
        In \Cref{sec:DRT}, we will directly regularize these two key factors to promote certified robustness of ensembles.
        
        
        
        
        \paragraph{Impact of Model-Smoothness Bound $\beta$.}
        From \Cref{thm:gradient-based-sufficient-necessary-cond-weight-ensemble}, we observe that:
        (1)~if $\min_{y_i \neq y_0} I_{y_i} \le - \beta r$, $\Mweight$ is guaranteed to be $r$-robust~(sufficient condition);
        and (2)~if $\min_{y_i \neq y_0} I_{y_i} > \beta r$, $\Mweight$ cannot be $r$-robust~(necessary condition).
        However, if $\min_{y_i \neq y_0} I_{y_i} \in (-\beta r,\, \beta r]$, we only know $\Mweight$ is possibly $r$-robust.
        As a result, the model-smoothness bound $\beta$ decides the correlation strength between $\min_{y_i\neq y_0} I_{y_i}$ and the robustness of $\Mweight$:
        if $\beta$ becomes larger, $\min_{y_i\neq y_0} I_{y_i}$ is more likely to fall in $(-\beta r,\, \beta r]$, inducing an undetermined robustness status from \Cref{thm:gradient-based-sufficient-necessary-cond-weight-ensemble}, vice versa.
        Specifically, when $\beta = 0$, i.e., all base models are linear, the gap is closed and we can always certify the robustness of $\Mweight$ via comparing $\min_{y_i \neq y_0}$ with $0$.
        Similar observations can be drawn for \shortOurEnsemble.
        Therefore, to strengthen the correlation between $I_{y_i}$ and ensemble robustness, we would need model-smoothness bound $\beta$ to be small.
    
    \vspace{-0.5em}
    \subsection{Robustness Conditions for Smoothed Ensemble Models}
        \label{subsec:2-3}
        \vspace{-0.5em}
        
        Typically neural networks are nonsmooth or admit only coarse smoothness bounds~\citep{sinha2018certifying}, i.e., $\beta$ is large.
        Therefore, applying \Cref{thm:gradient-based-sufficient-necessary-cond-weight-ensemble} for normal nonsmooth models would lead to near-zero certified radius.
        Therefore, we propose soft smoothing to enforce the smoothness of base models.
        However, with the soft smoothed base models, directly applying \Cref{thm:gradient-based-sufficient-necessary-cond-weight-ensemble} to certify robustness is still practically challenging, since the LHS of \Cref{eq:gradient-based-sufficient-cond-weight-ensemble,eq:gradient-based-necessary-cond-weight-ensemble} involves gradient of the soft smoothed confidence.
        A precise computation of such gradient requires high-confidence estimation of high-dimensional vectors via sampling, which requires linear number of samples with respect to input dimension~\citep{mohapatra2020higher,salman2019provably} and is thus too expensive in practice. 
        To solve this issue, we then propose \beforeProtocol as the practical smoothing protocol, which serves as an approximation of soft smoothing, so as to leverage the randomized smoothing based techniques for certification.
        
        
        \paragraph{Soft Smoothing.}
        To impose base models' smoothness, we now introduce soft smoothing~\citep{aounan2020certifying}, which applies randomized smoothing over \emph{the confidence scores}.
        Given base model's confidence function $f: \sR^d \to \vDelta^C$~(see \Cref{subsec:prelim}), we define \emph{soft smoothed confidence} by $\bar g_f^\varepsilon : \vx \mapsto \E_\varepsilon f(\vx + \varepsilon)$.
        Note that \emph{soft} smoothed confidence is different from smoothed confidence $g^\varepsilon_F$ defined in \Cref{def:smoothed-conf-g}.
        We consider soft smoothing instead of classical smoothing in \Cref{def:smoothed-conf-g} since soft smoothing reveals differentiable and thus practically regularizable training objectives.
        The following theorem shows the smoothness bound for $\bar g_f^\varepsilon$.
        \begin{theorem}[Model-Smoothness Upper Bound for $\bar g_f^\varepsilon$]
            \label{thm:smoothness-bound}
            Let $\varepsilon \sim \gN(0,\sigma^2 \mI_d)$ be a Gaussian random variable, 
            then the soft smoothed confidence function $\bar g^\varepsilon_{f}$ is $(2/\sigma^2)$-smooth.
        \end{theorem}
        We defer the proof to \Cref{subsec:a-4}.
        The proof views the Gaussian smoothing as the Weierstrass transform~\citep{weierstrass1885analytische} of a function from $\sR^d$ to $[0,\,1]^C$, leverages the symmetry property, and bounds the absolute value of diagonal elements of the Hessian matrix.
        Note that a Lipschitz constant $\sqrt{2/(\pi\sigma^2)}$ is derived for smoothed confidence in previous work~\cite[Lemma 1]{salman2019provably}, which characterizes only the first-order smoothness property; while our bound in addition shows the second-order smoothness property.
        In \Cref{subsec:a-4}, we further show that our smoothness bound in \Cref{thm:smoothness-bound} is tight up to a constant factor.
        
        Now, we apply \shortWeightedEnsemble and \shortOurEnsemble protocols with these soft smoothness confidence $\{\bar g_i^\varepsilon(\vx_0)\}_{i=1}^N$ as base models' confidence scores, and obtain soft ensemble $\bar G^\varepsilon_{\Mweight}$ and $\bar G^\varepsilon_{\Mmme}$ respectively.
        Since each $\bar g_i^\varepsilon$ is $(2/\sigma^2)$-smooth, take \shortWeightedEnsemble as an example, 
        we can study the ensemble robustness with \Cref{thm:gradient-based-sufficient-necessary-cond-weight-ensemble}.
        %
        We state the full statement in \Cref{cor:gradient-based-sufficient-necessary-cond-smoothed-weight-ensemble}~(and in \Cref{cor:gradient-based-sufficient-necessary-cond-smoothed-our-ensemble} for \shortOurEnsemble) in \Cref{adxsubsec:robust-cond-smooth-ensemble}.
        From the corollary, we observe that the corresponding ERI for the soft smoothed \shortWeightedEnsemble can be written as
        \vspace{-0.5em}
        \begin{equation}
            \small
            \bar I_{y_i} := \Big\|\E_{\varepsilon} \nabla_\vx \sum_{j=1}^N w_j f_j^{y_0/y_i}(\vx_0+\varepsilon) \Big\|_2 \Big/ \| \vw \|_1 
            -
            \frac{1}{r \|\vw\|_1} \E_{\varepsilon} \sum_{j=1}^N w_j f_j^{y_0/y_i}(\vx_0+\varepsilon).
            \label{eq:Itilde}
            \vspace{-0.4em}
        \end{equation}
        We have following observations:
        (1)~unlike for standard models with unbounded $\beta$, for the smoothed ensemble models, this ERI~(\Cref{eq:Itilde}) would have \textit{guaranteed} correlation
        with the model robustness since $\beta = \Theta(1/\sigma^2)$ is bounded and can be controlled by tuning $\sigma$ for smoothing. 
        (2)~we can still control  ERI by diversifying gradients and ensuring large confidence margins as discussed in \Cref{subsec:2-2}, but need to compute on the noise augmented input $\vx_0+\varepsilon$ instead of original input $\vx_0$.
        
        \paragraph{Towards Practical Certification.}
        As outlined at the beginning of this subsection, even with smoothed base models, certifying robustness using \Cref{thm:gradient-based-sufficient-necessary-cond-weight-ensemble} is practically difficult.
        Therefore, we introduce \beforeProtocol strategy as below to construct $G^\varepsilon_{\Mweight}$ and $G^\varepsilon_{\Mmme}$ as approximations of soft ensemble $\bar G^\varepsilon_{\Mweight}$ and $\bar G^\varepsilon_{\Mmme}$ respectively.
        
        \begin{definition}[\newterm{\beforeProtocol~(\shortBeforeProtocol)}]
            \label{def:ebs}
            Let $\gM$ be an ensemble model over base models $\{F_i\}_{i=1}^N$ and
            $\rvepsilon$ be a random variable.
            The \shortBeforeProtocol strategy construct smoothed classifier $G^\rvepsilon_{\gM}: \sR^d \to [C]$ that picks the class with highest smoothed confidence of $\gM$: $G^\rvepsilon_{\gM}(\vx) := \arg\max_{j\in [C]} g^\rvepsilon_{\gM}(\vx)_j$.
        \end{definition}
        
        Here $\gM$ could be either $\Mweight$ or $\Mmme$.
        \shortBeforeProtocol aims to approximate the soft smoothed ensemble.
        Formally, use \shortWeightedEnsemble as an example, we let $f_{\Mweight} := \frac{\sum_{j=1}^N w_j f_j}{\|\vw\|_1}$ to be \shortWeightedEnsemble ensemble's confidence, then
        \vspace{-0.5em}
        \begin{equation}
            \small
            g^\varepsilon_{\Mweight}(\vx)_i = \E_\varepsilon \1[\Mweight(\vx+\varepsilon)=i] \approx
            \E_\varepsilon f_{\Mweight}(\vx+\varepsilon)_i
            = \dfrac{\sum_{j=1}^N w_j (\bar g_{f_j}^\varepsilon)_i}{\sum_{j=1}^N w_j} = \bar g^\varepsilon_{\Mweight}(\vx)_i
            \label{eq:approximation}
            \vspace{-0.3em}
        \end{equation}
        where LHS is the smoothed confidence of \shortBeforeProtocol ensemble and RHS is the soft smoothed ensemble's confidence.
        Such approximation is also adopted in existing work~\citep{salman2019provably,zhai2019macer,aounan2020certifying} and shown effective and useful.
        Therefore, our robustness analysis of soft smoothed ensemble still applies with EBS and we can control ERI in \Cref{eq:Itilde} to improve the certified robustness of \shortBeforeProtocol ensemble.
        For \shortBeforeProtocol ensemble, we can leverage randomized smoothing based techniques to compute the robustness certification~(see \Cref{adx-prop:rob-radii-before} in \Cref{adx-sec:2}).

        \vspace{-0.2em}
    \subsection{Additional Properties of ML Ensembles}
            \vspace{-0.5em}
            \paragraph{Comparison between Ensemble and Single-Model Robustness.}
            In \Cref{subsec:a-1}, we show \Cref{cor:robust-radius-lower-bound-for-ensemble-constructed-from-single-models}, a corollary of \Cref{thm:gradient-based-sufficient-necessary-cond-weight-ensemble}, which indicates that when the base models are smooth enough, 
            both \shortWeightedEnsemble and \shortOurEnsemble ensemble models are more certifiably robust than the base models.
            This aligns with our empirical observations~(see \Cref{tab:mnisttable} and \Cref{tab:cifartable}), though \emph{without} advanced training approaches such as \shortApproach, the improvement of robustness brought by ensemble itself is  marginal.
             In \Cref{subsec:a-1}, we also show larger number of base models $N$ can lead to better certified robustness.
        
            \vspace{-0.5em}
            \paragraph{Comparison between \shortWeightedEnsemble and \shortOurEnsemble Robustness.}
            Since in actual computing, the certified radius of a smoothed model is directly correlated with the probability of correct prediction under smoothed input~(see \Cref{eq:rand-smooth-vanilla-r} in \Cref{adx:rand-smooth-background}), we study the robustness of both \shortWeightedEnsemble and \shortOurEnsemble along with single models from the statistical robustness perspective in \Cref{adx-sec:3}.
            From the study, we have the following theoretical observations verified by numerical experiments:
            (1)~\shortOurEnsemble is more robust when the adversarial transferability is high; while \shortWeightedEnsemble is more robust when the adversarial transferability is low.
            (2)~If we further assume that $f_i(x_0 + \varepsilon)_{y_0}$ follows marginally uniform distribution, when the number of base models $N$ is sufficiently large,  \shortOurEnsemble is always more certifiably robust. 
            \Cref{adx-subsec:3-4} entails the numerical evaluations that verify our theoretical conclusions.

        \vspace{-0.5em}
\section{\boldApproach}
        \vspace{-0.5em}

    \label{sec:DRT}
    
    Inspired by the above key factors in the sufficient and necessary conditions for the certifiably robust ensembles, we propose the \boldApproach (\shortApproach).
    In particular, let $\vx_0$ be a training sample, DRT contains the following two regularization terms in the objective function to minimize:
    \begin{itemize}[leftmargin=0.5cm,noitemsep]
        \item \LHSloss~(\shortLHSloss):
        \vspace{-2.2em}
        \begin{equation}
        \begin{small}
            \hspace{16.35em}
            \Lgd(\vx_0)_{ij} = \big\|\nabla_\vx f_i^{y_0/y_i^{(2)}}(\vx_0) + \nabla_\vx f_j^{y_0/y_j^{(2)}}(\vx_0)\big\|_2.
            \label{eq:loss1}
        \end{small}
        \end{equation}
        \vspace{0.2em}
        \item \RHSloss~(\shortRHSloss):
        \vspace{-2.3em}
        \begin{equation}
        \begin{small}
            \hspace{16.75em}
            \Lcm(\vx_0)_{ij} = f_i^{y_i^{(2)}/y_0}(\vx_0) + f_j^{y_j^{(2)}/y_0} (\vx_0).
            \label{eq:loss2}
        \end{small}
        \end{equation}
    \end{itemize}
    \vspace{-1em}
    In \Cref{eq:loss1,eq:loss2},  $y_0$ is the ground-truth label of  $\vx_0$, and $y_i^{(2)}$~(or $y_j^{(2)}$) is the runner-up class of base model $F_i$~(or $F_j$). 
    Intuitively, for each model pair $(F_i,F_j)$ where $i,j\in [N]$ and $i\neq j$, the GD loss 
    promotes the \textit{diversity of gradients} between the base model $F_i$ and $F_j$.
        Note that the gradient computed here is actually the gradient difference between different labels.
        As our theorem reveals, it is the gradient difference between different labels instead of pure gradient itself that matters, which improves existing understanding of gradient diversity~\citep{pang2019improving,demontis2019adversarial}.
        Specifically, the GD loss encourages both large gradient diversity and small base models' gradient magnitude in a naturally balanced way, and encodes the interplay between gradient magnitude and direction diversity.
        In contrast, solely regularizing the base models' gradient would hurt the model's benign accuracy, and solely regularizing gradient diversity is hard to realize due to the boundedness of cosine similarity.
    %
    The CM loss encourages the \textit{large margin} between the true and runner-up classes for base models.
    Both regularization terms are directly motivated by theoretical analysis in \Cref{sec:2}. 
    
    For each input $\vx_0$ with ground truth $y_0$, we use $\vx_0 + \rvepsilon$ with $\rvepsilon \sim \gN(0,\sigma^2 \mI_d)$ as training input for each base model~(i.e., Gaussian augmentation). 
    We call two base models $\big(F_i, F_j\big)$ a \emph{valid} model pair at $(\vx_0, y_0)$ if both $F_i(\vx_0+\rvepsilon)$ and $F_j(\vx_0+\rvepsilon)$ equal to $y_0$. For every valid model pair, we apply DRT: \shortLHSloss and \shortRHSloss with $\rho_1$ and $\rho_2$ as the weight hyperparameters as below.
        \begin{equation*}
        \vspace{-0.5mm}
        \begin{aligned}
            \mathcal{L}_{\text{train}} = & \sum_{i\in [N]} \mathcal{L}_{\mathrm{std}}(\vx_0 + \rvepsilon, y_0)_i 
             +
            \rho_1 \sum_{\substack{i,j \in [N], i\neq j\\ (F_i,\,F_j)\text{ is valid}}} 
            \Lgd(\vx_0 + \rvepsilon)_{ij} 
            +
            \rho_2 \sum_{\substack{i,j\in [N], i \neq j\\ (F_i,\,F_j)\text{ is valid}}} 
            \Lcm(\vx_0 + \rvepsilon)_{ij}.
        \end{aligned}
        \vspace{-0.85mm}
        \end{equation*}
        The standard training loss $\mathcal{L}_{\mathrm{std}}(\vx_0+\rvepsilon, y_0)_i$ of each base model $F_i$ is either cross-entropy loss \citep{cohen2019certified}, or adversarial training loss \citep{salman2019provably}.
        This standard training loss will help to produce sufficient valid model pairs with high benign accuracy for robustness regularization.
        Specifically, as discussed in \Cref{subsec:2-3}, we compute $\Lgd$ and $\Lcm$ on the  noise augmented inputs $(\vx_0+\varepsilon)$ instead of $\vx_0$  to improve the certified robustness for the \emph{smoothed} ensemble.

    \vspace{-0.5em}
    \paragraph{Discussion.}
        To our best knowledge, this is the \textit{first} training approach that is able to promote the certified robustness of ML ensembles, while existing work either only provide  empirical robustness without guarantees~\citep{pang2019improving,kariyappa2019improving,yang2020dverge,yang2021trs}, or tries to only optimize the weights of \weightedEnsemble~\citep{zhang2019enhancing,liu2020enhancing}.
        We should notice that, though concepts similar with the gradient diversity have been explored in empirically robust ensemble training (e.g., ADP~\citep{pang2019improving}, GAL~\citep{kariyappa2019improving}), directly applying these regularizers cannot train models with high \textit{certified robustness} due to the lack of theoretical guarantees in their design. We indicate this through  ablation studies in \Cref{adx-subsec:adpgal}.
        For the design of \shortApproach, we also find that there exist some variations.
        We analyze them and show that the current design is usually better based on the analysis in \Cref{adxsec:alternative-design}.
        Our approach is generalizable for other $L_p$-bounded perturbations such as $L_1$ and $L_\infty$ leveraging existing work~\citep{li2019certified,lecuyer2019certified,yang2020randomized,levine2021improved}.

    \vspace{-0.5em}
\section{Experimental Evaluation}
    \vspace{-0.5em}
    \label{sec:5}
    
    To make a thorough comparison with existing certified robustness approaches, we evaluate  DRT on different datasets including MNIST~\citep{lecun2010mnist},  CIFAR-10~\citep{krizhevsky2009learning}, and ImageNet~\citep{deng2009imagenet}, based on both MME and WE  protocols. 
    Overall, we show that the DRT enabled ensemble  outperforms all baselines in terms of certified robustness under different settings. 
    

    \vspace{-0.55em}
\subsection{Experimental Setup}
    \vspace{-0.55em}
\label{sec:implementation-and-evaluation}
\textbf{Baselines.} 
We consider the following state-of-the-art baselines for certified robustness:
\textbf{Gaussian smoothing}~\citep{cohen2019certified},
\textbf{SmoothAdv}~\citep{salman2019provably},
\textbf{MACER}~\citep{zhai2019macer},
\textbf{Stability}~\citep{li2019certified},
and \textbf{SWEEN}~\citep{liu2020enhancing}.
Detail description of these baselines can be found in \Cref{adx-subsec:exp}.
We follow the configurations of baselines, and compare DRT-based ensemble with Gaussian Smoothing, SmoothAdv, and MACER on all datasets, and in addition compare it with other baselines on MNIST and CIFAR-10 considering the training efficiency.
There are other baselines, e.g., \citep{jeong2020consistency}. However, SmoothAdv performs consistently better across different datasets, so we mainly consider SmoothAdv as our strong baseline.


\textbf{Models.} For base models in our ensemble, we follow the configurations used in baselines: LeNet~\citep{lecun1998gradient}, ResNet-110, and ResNet-50~\citep{he2016deep} for MNIST, CIFAR-10, and ImageNet datasets respectively. 
Throughout the experiments, we use $N=3$ base models to construct the ensemble for demonstration. 
We expect more base models would yield higher ensemble robustness.

    \textbf{Training Details.}
    We follow \Cref{sec:DRT} to train the base models. We combine \shortApproach with Gaussian smoothing and SmoothAdv (i.e., instantiating $\gL_{\mathrm{std}}$ by either cross-entropy loss \citep{cohen2019certified,yang2020randomized} or adversarial training loss \citep{salman2019provably}).
    We leave training details along with hyperparametes in \Cref{adx-subsec:exp}.

        \begin{figure}[!t]
            \centering
            \vspace{-7mm}
            \includegraphics[width=0.86\linewidth]{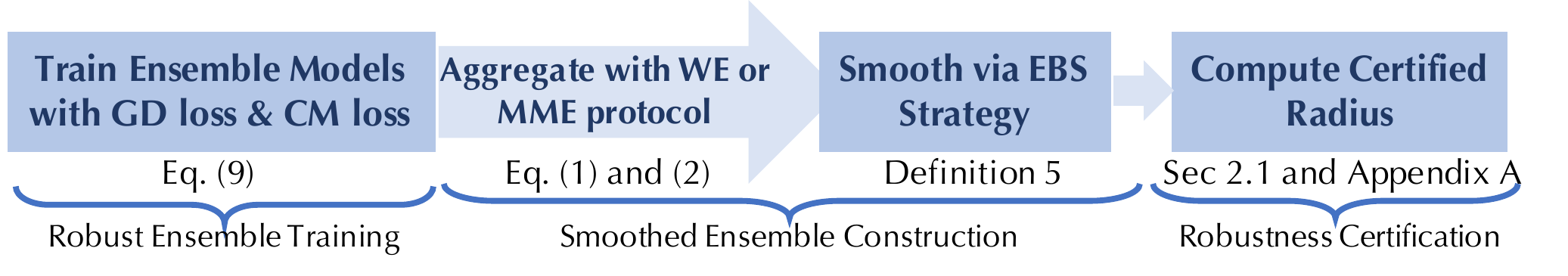}
            \vspace{-1mm}
            \caption{\small Pipeline for DRT-based ensemble.}
            \label{fig:pip_robust}
            \vspace{-2mm}
        \end{figure}
        
    \textbf{Pipeline.}
        After the base models are trained with \shortApproach, we aggregate them to form the ensemble $\gM$, using either \shortWeightedEnsemble or \shortOurEnsemble protocol~(see \Cref{def:weighted-ensemble,def:maximum-margin-ensemble}).
        If we use \shortWeightedEnsemble, to filter out the effect of different weights, we adopt the average ensemble where all weights are equal.
        We also studied how optimizing weights can further improve the certified robustness in \Cref{adx-subsec:optimize-weights}.
        Then, we leverage \beforeProtocol strategy to form a smoothed ensemble~(see \Cref{def:ebs}).
        Finally, we compute the certified robustness for the smoothed ensemble
        based on Monte-Carlo sampling with high-confidence~($99.9\%$). The training pipeline is shown in~\Cref{fig:pip_robust}. 
        
        
        \textbf{Evaluation Metric.}
        We report the standard \emph{certified accuracy} under different $L_2$ radii $r$'s as our evaluation metric following existing work~\citep{cohen2019certified,yang2020dverge,zhai2019macer,jeong2020consistency}.
        More evaluation details are in \Cref{adx-subsec:exp}.

\vspace{-0.9em}
\subsection{Experimental Results}
\vspace{-0.7em}


Here we consider ensemble models consisting of three base models.
We show that 1) DRT-based ensembles outperform the SOTA baselines significantly especially under large perturbation radii; 2) smoothed ensembles are always more certifiably robust than each base model (\Cref{cor:robust-radius-lower-bound-for-ensemble-constructed-from-single-models} in \Cref{subsec:a-1}); 3) applying DRT for either MME or WE ensemble protocols achieves similar and consistent improvements on certified robustness.
\begin{table*}[t]
\small
\centering
\vspace{-0.25mm}
\caption{\small {Certified accuracy} under different radii on MNIST dataset. The grey rows present the performance of the proposed \shortApproach approach. The brackets show the base models we use.}
\vspace{-3mm}
\resizebox{0.9\textwidth}{!}{
\begin{tabular}{l|c|c|c|c|c|c|c|c|c|c|c}
\toprule 
                     \multicolumn{1}{l|}{Radius $r$}
                     & \multicolumn{1}{l|}{$0.00$} & \multicolumn{1}{l|}{$0.25$} & \multicolumn{1}{l|}{$0.50$} & \multicolumn{1}{l|}{$0.75$} & \multicolumn{1}{l|}{$1.00$} & \multicolumn{1}{l|}{$1.25$} & \multicolumn{1}{l|}{$1.50$} &
                     \multicolumn{1}{l|}{$1.75$} &
                     \multicolumn{1}{l|}{$2.00$} &
                     \multicolumn{1}{l|}{$2.25$} &
                     \multicolumn{1}{l}{$2.50$} \\ \midrule
Gaussian~\citep{cohen2019certified}               & 99.1                         & 97.9                         & 96.6                         & 94.7                         & 90.0                         & 83.0                         & 68.2  & 46.6 & 33.0 & 20.5 & 11.5                         \\
SmoothAdv~\citep{salman2019provably}        & 99.1                         & 98.4                         & 97.0                         & 96.3                         & 93.0                         & 87.7                         & 80.2   & 66.3 & 43.2 & 34.3 & 24.0                          \\
MACER~\citep{zhai2019macer}              & 99.2   & 98.5   & 97.4   & 94.6   & 90.2   & 83.5   & 72.4   & 54.4   & 36.6   & 26.4   & 16.5   \\
Stability~\citep{li2019certified}          & 99.3   & \textbf{98.6}   & 97.1   & 93.8   & 90.7   & 83.2   & 69.2   & 46.8   & 33.1   & 20.0   & 11.2   \\
SWEEN (Gaussian)~\citep{liu2020enhancing}      & 99.2   & 98.4   & 96.9   & 94.9   & 90.5   & 84.4   & 71.1   & 48.9   & 35.3 & 23.7 & 12.8   \\
SWEEN (SmoothAdv)~\citep{liu2020enhancing}      & 99.2   & 98.2   & 97.4   & 96.3   & 93.4 & 88.1   & 81.0 & 67.2   & 44.5   & 34.9   & 25.0   \\

\hline
MME (Gaussian) & 99.2                         & 98.4                         & 96.8                         & 94.9                         & 90.5                         & 84.3                         & 69.8      & 48.8 & 34.7 & 23.4 & 12.7                       \\
\rowcolor{tabgray} DRT + 
MME (Gaussian) & \textbf{99.5}                         & \textbf{98.6}                         & 97.5                            & 95.5                         & 92.6                         & 86.8                         & 76.5       & 60.2 & 43.9 & 36.0 & 29.1                      \\

MME (SmoothAdv)  & 99.2                         & 98.2                        & 97.3                         & 96.4                         & {93.2}                         & 88.1                         & 80.6  & 67.9 & 44.8 & 35.0 & 25.2                           \\
\rowcolor{tabgray} DRT + 
MME (SmoothAdv)  & 99.2                         & 98.4                         & \textbf{97.6}                         & \textbf{96.7}                         & 93.1                         & \textbf{88.5}                         & 83.2  & 68.9 & 48.2 & \bf 40.3 & 34.7                           \\

\hline
WE (Gaussian)  & 99.2                         & 98.4                         & 96.9                         & 94.9                         & 90.6                         & 84.5                         & 70.4 & 49.0 & 35.2 & 23.7 & 12.9   \\
\rowcolor{tabgray}DRT + WE (Gaussian)  & \textbf{99.5}                         & \textbf{98.6}                         & 97.4                         & 95.6                         & 92.6                         & 86.7                         & 76.7         & 60.2 & 43.9 & 35.8 & 29.0                   \\
WE  (SmoothAdv)  & 99.1                         & 98.2                         & 97.4                         & 96.4                         & \textbf{93.4}                         & 88.2                         & 81.1   & 67.9 & 44.7 & 35.2 & 24.9                         \\
\rowcolor{tabgray} DRT + WE (SmoothAdv)   & 99.1                         & 98.4                         & \textbf{97.6}                         & \textbf{96.7}                         & \textbf{93.4}                         & \textbf{88.5}                         & \textbf{83.3}   & \textbf{69.6} & \textbf{48.3} & 40.2 & \textbf{34.8}                          \\

\bottomrule
\end{tabular}}
\vspace{-2em}
\label{tab:mnisttable}
\end{table*}


\begin{table*}[t]
\small
\centering
\vspace{-4mm}
\caption{\small {Certified accuracy} under different radii on CIFAR-10 dataset. The grey rows present the performance of the proposed \shortApproach approach. The brackets show the base models we use.} 
\label{tab:cifartable}
\vspace{-3mm}
\resizebox{0.9\textwidth}{!}{
\begin{tabular}{l|c|c|c|c|c|c|c|c|c}
\toprule 
                     \multicolumn{1}{l|}{Radius $r$}
                     & \multicolumn{1}{l|}{$0.00$} & \multicolumn{1}{l|}{$0.25$} & \multicolumn{1}{l|}{$0.50$} & \multicolumn{1}{l|}{$0.75$} & \multicolumn{1}{l|}{$1.00$} & \multicolumn{1}{l|}{$1.25$} & \multicolumn{1}{l|}{$1.50$} &
                     \multicolumn{1}{l|}{$1.75$} &
                     \multicolumn{1}{l}{$2.00$} \\ \midrule
Gaussian~\citep{cohen2019certified}               & 78.9                         & 64.4                         & 47.4                         & 33.7                         & 23.1                         & 18.3                         & 13.6  & 10.5 & 7.3         \\
SmoothAdv~\citep{salman2019provably}        & 68.9                         & 61.0                         & 54.4                         & 45.7                         & 34.8                         & 28.5                         & 21.9   & 18.2 & 15.7                 \\
MACER~\citep{zhai2019macer}       & 79.5   & 68.8   & 55.6   & 42.3   & 35.0   & 27.5   & 23.4   & 20.4   & 17.5   \\
Stability~\citep{li2019certified}  & 72.4   & 58.2   & 43.4   & 27.5   & 23.9   & 16.0   & 15.6   & 11.4   & 7.8    \\
SWEEN (Gaussian)~\citep{liu2020enhancing}      & 81.2   & 68.7   & 54.4   & 38.1   & 28.3   & 19.6   & 15.2   & 11.5   & 8.6   \\
SWEEN (SmoothAdv)~\citep{liu2020enhancing}      & 69.5   & 62.3   & 55.0   & 46.2   & 35.2   & 29.5   & 22.4   & 19.3   & 16.6   \\

\hline
MME (Gaussian) & 80.8                         & 68.2                         & 53.4                         & 38.4                         & 29.0                         & 19.6                        & 15.6      & 11.6 & 8.8                      \\

\rowcolor{tabgray} DRT + 
MME (Gaussian) & 81.4                         & \textbf{70.4}                         & 57.8                         & 43.8                         & 34.4                         & 29.6                        & 24.9      & 20.9 & 16.6                      \\
MME (SmoothAdv)  & 71.4                         & 64.5                         &       57.6                   & 48.4                         & 36.2                         & 29.8                         & 23.9  & 19.5 & 16.2                            \\
\rowcolor{tabgray} DRT + 
    MME (SmoothAdv)  & 72.6                        & 67.2                         &  \textbf{60.2}                        & 50.4                         & 39.4                         & 35.8                         & \textbf{30.4}                         & 24.0    & 20.1                            \\

\hline
WE (Gaussian)  & 80.8                         & 68.4                         & 53.6                         & 38.4                         & 29.2                         & 19.7                         & 15.9 & 11.8 & 8.9    \\
\rowcolor{tabgray} DRT + WE (Gaussian)  & \bf 81.5                         & \textbf{70.4}                         & 57.9                         & 44.0                         & 34.2                         & 29.6                         & 24.9         & 20.8 & 16.4                   \\
WE  (SmoothAdv)  & 71.8                         & 64.6                         &  57.8                         & 48.5                         & 36.2                         & 29.6                         & 24.2    & 19.6 & 16.0                         \\
\rowcolor{tabgray} DRT + WE (SmoothAdv)   & 72.6                         & 67.0                         & \textbf{60.2}                         & \textbf{50.5}                         & \textbf{39.5}                         & \textbf{36.0}                         & 30.3    & \textbf{24.1} & \textbf{20.3}                        \\
\bottomrule
\end{tabular}}
\vspace{-2em}
\end{table*}

\noindent\textbf{Certified Robustness of DRT with Different Ensemble Protocols.} 
The evaluation results on MNIST, CIFAR-10, ImageNet are shown in Tables~\ref{tab:mnisttable},~\ref{tab:cifartable},~\ref{tab:imagenettable} respectively. It is clear that though the certified accuracy of a single model can be improved by directly applying either MME or WE ensemble training (proved in \Cref{cor:robust-radius-lower-bound-for-ensemble-constructed-from-single-models}), such improvements are usually negligible (usually less than $2\%$). 
In contrast, in all tables we find DRT provides significant gains on certified robustness for both \shortOurEnsemble and \shortWeightedEnsemble~(up to over $16\%$ as \Cref{tab:mnisttable} shows).

From Tables~\ref{tab:mnisttable} and~\ref{tab:cifartable} on MNIST and CIFAR-10, we find that compared with all baselines, DRT-based ensemble
achieves the highest robust accuracy, and the performance gap is more pronounced on large radii (over $8\%$ for $r=2.50$ on MNIST and $6\%$ for $r=1.50$ on CIFAR-10). 
We also demonstrate the scalability of DRT by training on ImageNet, and 
Table~\ref{tab:imagenettable} shows that DRT achieves the highest certified robustness under large radii.
It is clear that DRT can be easily combined with existing training approaches (e.g. Gaussian smoothing or SmoothAdv), boost their certified robustness, and set the state-of-the-art results to the best of our knowledge.

\looseness=-1
To evaluate the computational cost of DRT, we 
analyze the theoretical complexity in \Cref{adxsec:alternative-design} and
compare the efficiency of different methods in practice
in \Cref{adx-subsec:mnist,adx-subsec:cifar}. 
In particular, we show that DRT with Gaussian Smoothing base models even achieves around two times speedup compared with SmoothAdv with comparable or even higher certified robustness, since DRT does not require adversarial training. More discussions about hyper-parameters settings for DRT can be found in \Cref{adx-subsec:exp}. In \Cref{adx-subsec:adpgal}, we also show that our proposed DRT approach could achieve $6\%\sim10\%$ higher certified accuracy compared to adapted ADP~\citep{pang2019improving} and GAL~\citep{kariyappa2019improving} training on large radii for both MNIST and CIFAR-10 datasets.

\begin{wraptable}{r}{0.575\textwidth}
\small
\centering
    \vspace{-1.5em}
\caption{\small {Certified accuracy} under different radii on ImageNet dataset. The grey rows present the performance of the proposed \shortApproach approach. The brackets show the base models we use.}
    \vspace{-0.5em}
\scalebox{0.65}{
\begin{tabular}{l|c|c|c|c|c|c|c}
\toprule
   Radius $r$                  & \multicolumn{1}{l|}{$0.00$} & \multicolumn{1}{l|}{$0.50$} & \multicolumn{1}{l|}{$1.00$} & \multicolumn{1}{l|}{$1.50$} & \multicolumn{1}{l|}{$2.00$} & \multicolumn{1}{l|}{$2.50$} & \multicolumn{1}{l}{$3.00$} \\ \midrule
Gaussian~\citep{cohen2019certified}               & 57.2                         & 46.2                         &        37.0                  & 29.2                         & 19.6                         & 15.2                         & 12.4                           \\
SmoothAdv~\citep{salman2019provably}        &            54.6              & 49.0                         & 43.8                         & 37.2                         & 27.0                         & 25.2                         & 20.4                           \\
MACER~\citep{zhai2019macer}        &         \bf    68.0              & \bf 57.0                         & 43.0                         & 31.0                        & 25.0                         & 18.0                         & 14.0                       \\
SWEEN (Gaussian)~\citep{liu2020enhancing}      & 58.4   & 47.0   & 37.4   & 29.8   & 20.2   & 15.8   & 12.8   \\
SWEEN (SmoothAdv)~\citep{liu2020enhancing}      & 55.2   & 50.0   & 44.2   & 37.8   & 27.6   & 26.6   & 21.6   \\
\hline
MME (Gaussian) & 58.0                         & 47.2                         & 38.8                         & 31.2                   & 21.4                         & 16.4                         & 14.2                           \\
\rowcolor{tabgray} DRT + 
MME (Gaussian) & 52.2                         & 46.8                         & 42.4                         & 34.2                         & 24.0                         & 19.6                         & 18.0                           \\

MME (SmoothAdv)  & 55.0                         & 50.2                         & 44.2                         & 38.6                    & 27.4     & 26.4                         & 21.6                                                \\
\rowcolor{tabgray} DRT + 
MME (SmoothAdv)  & 49.8                         & 46.8                         & \bf 44.4                         & \textbf{39.8}                         & 30.2                         & 28.2                         & \textbf{23.4}                           \\
\hline
WE (Gaussian)  & 58.2                         & 47.2                         & 38.6                         & 31.2                         & 21.6                         & 17.0                         & 14.4  \\

\rowcolor{tabgray} DRT + WE (Gaussian)  & 52.2                         & 46.8                         & 41.8                         & 33.6                         & 24.2                         & 19.8                         & 18.4                           \\

WE  (SmoothAdv)  & 55.2                         &  50.2                         & \bf 44.4                         & 38.6                         & 28.2                         & 26.2                         & 22.0                           \\
\rowcolor{tabgray} DRT + WE (SmoothAdv)   & 49.8                         & 46.6                         & \bf 44.4                         & 38.8                         & \textbf{30.4}                         & \textbf{29.0}                         & 23.2                           \\
\bottomrule
\end{tabular}}
\label{tab:imagenettable}

\end{wraptable}

\begin{wrapfigure}{r}{0.56\textwidth}
    \vspace{-1em}
    \begin{subfigure}{.27\textwidth}
    
    \includegraphics[width=\textwidth]{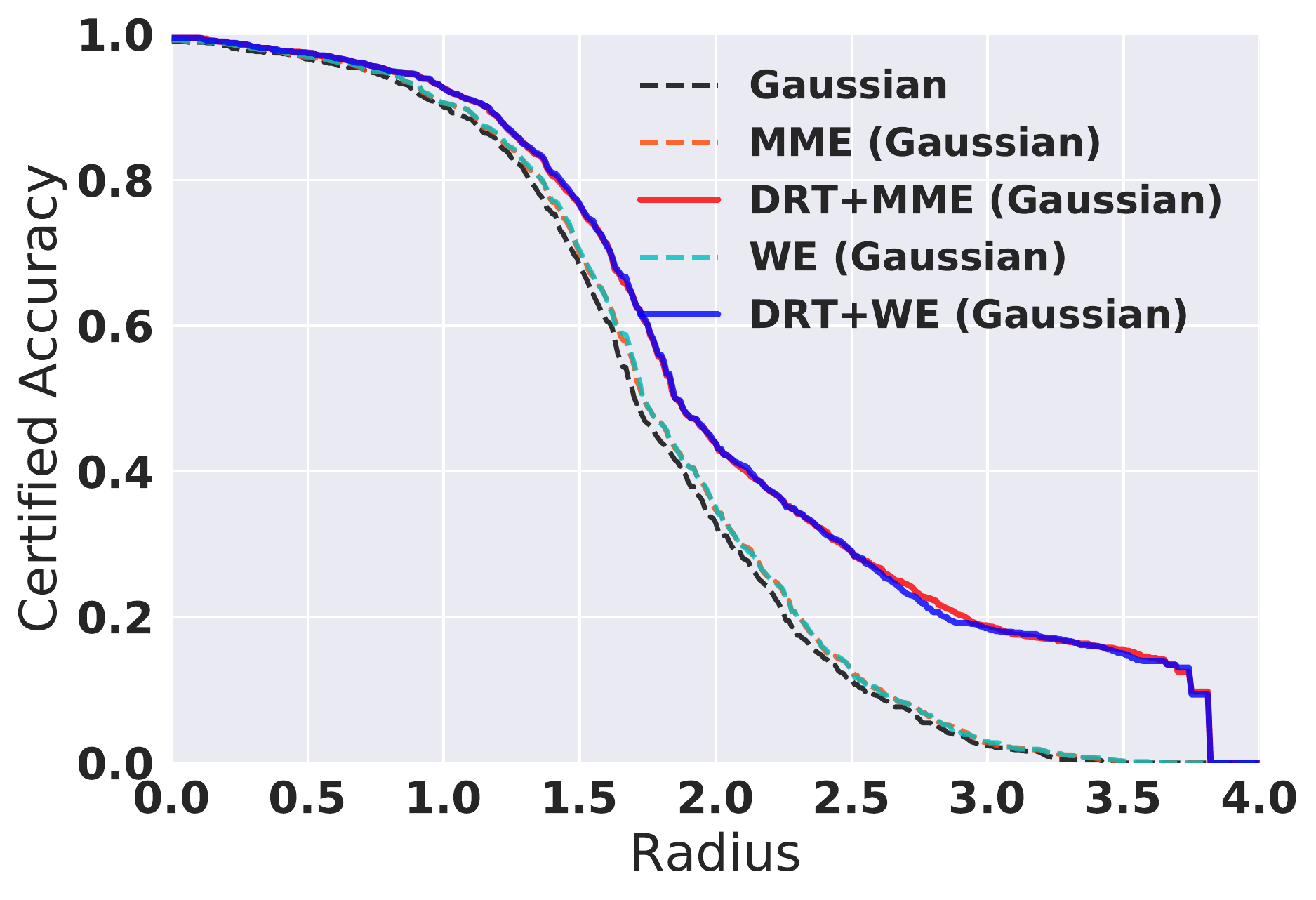}
    \caption{MNIST}
    \end{subfigure}
   \begin{subfigure}{.27\textwidth}
    \includegraphics[width=\textwidth]{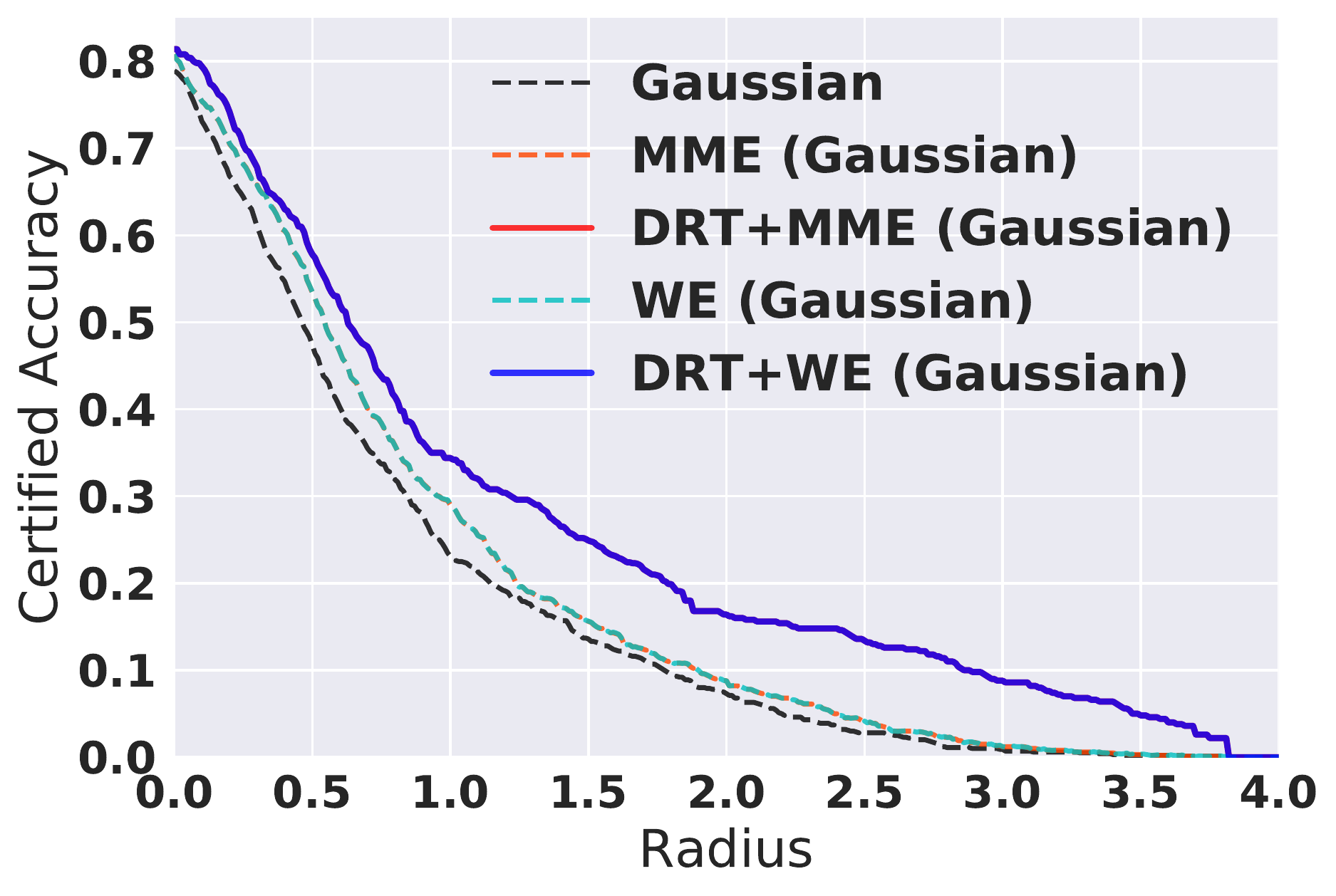}
    \caption{CIFAR-10}
    \end{subfigure}
    \vspace{-0.5em}
    \caption{\small  {Certified accuracy} for ML ensembles with Gaussian smoothed base models, under smoothing parameter $\sigma \in \{0.25, 0.50, 1.00\}$ on (Left) MNIST; (Right) CIFAR-10. }
    \label{fig:maincurve}
    \vspace{-1.5em}
\end{wrapfigure}

\noindent\textbf{Certified Accuracy with Different Perturbation Radius.}
We visualize the trend of certified accuracy along with different perturbation radii in Figure~\ref{fig:maincurve}. For each radius $r$, we present the best certified accuracy among different smoothing parameters $\sigma\in \{0.25, 0.50, 1.00\}$. We notice that while simply applying MME or WE protocol could  slightly improve the certified accuracy, DRT could significantly boost the certified accuracy under different radii. We also present the trends of different smoothing parameters separately in \Cref{adx-subsec:exp} which lead to similar conclusions. 

\noindent\textbf{Effects of GD and CM Losses in DRT.}
To explore the effects of individual Gradient Diversity and Confidence Margin Losses in DRT, we set $\rho_1$ or $\rho_2$ to 0 separately and tune the other for evaluation on MNIST and CIFAR-10. The full results are shown in \Cref{adx-subsec:seperate}. 
We observe that both GD and CM losses have positive effects on improving the certified accuracy, and GD plays a major role on larger radii. By combining these two regularization losses as DRT does, the ensemble model achieves the highest certified accuracy under all radii.

    \vspace{-0.5em}
\section{Conclusion}
    \vspace{-0.5em}

In this paper, we explored and characterized the robustness conditions for certifiably robust ensemble ML models theoretically, and proposed \shortApproach for training a robust ensemble. 
    Our analysis provided the justification of the regularization-based training approach \shortApproach.
Extensive experiments showed that \shortApproach-enhanced ensembles achieve the highest certified robustness compared with existing baselines.

\paragraph{Ethics Statement.}
In this paper, we characterized the robustness conditions for certifying ML ensemble robustness.
Based on the analysis, we propose \shortApproach to train a certifiably robust ensemble.
On the one hand, the training approach boosts the certified robustness of ML ensemble, thus significantly reducing the security vulnerabilities of ML ensemble.
On the other hand, the trained ML ensemble can only guarantee its robustness under specific conditions of the attack.
Specifically, we evaluate the trained ML ensemble on the held-out test set and constrain the attack to be within predefined $L_2$ distance from the original input.
We cannot provide robustness guarantee for all possible real-world inputs.
Therefore, users should be aware of such limitations of \shortApproach-trained ensembles, and should not blindly rely on the ensembles when the attack can cause large deviations measured by $L_2$ distance.
As a result, we encourage researchers to understand the potential risks, and evaluate whether our attack constraints align with their usage scenarios when applying our \shortApproach approach to real-world applications.
We do not expect any ethics issues raised by our work.

\paragraph{Reproducibility Statement.}
All the theorem statements are substantiated with rigorous proofs in \Cref{adxsec:sec2-theorem,adx-sec:2,adx-sec:3}.
In \Cref{adx-subsec:exp}, we list the details and hyperparameters for reproducing all experimental results.
Our evaluation is conducted on commonly accessible MNIST, CIFAR-10, and ImageNet datasets.
Finally, we upload the source code as the supplementary material for  reproducibility purpose.

\subsection*{Acknowledgements} 
This work was performed under the auspices of the U.S. Department of Energy by the Lawrence Livermore National Laboratory under Contract No. DE-AC52-07NA27344 and LLNL LDRD Program Project No. 20-ER-014. This work is partially supported by the NSF grant No.1910100, NSF CNS 20-46726 CAR, Alfred P. Sloan Fellowship, and Amazon Research Award.


\bibliography{bib}
\bibliographystyle{iclr2022_conference}

\newpage
\appendix

    In \Cref{adx:rand-smooth-background}, we provide more background knowledge about randomized smoothing.
    In \Cref{adxsec:sec2-theorem}, we first discuss the direct connection between the definition of $r$-robustness and the robustness certification of randomized smoothing, then prove the robustness conditions and the comparison results presented in \Cref{sec:2}.
    In \Cref{adx-sec:2}, we formally define, discuss, and theoretically compare the smoothing strategies for ensembles.
    In \Cref{adx-sec:3}, we characterize the robustness of smoothed ML ensembles from statistical robustness perspective which is directly related to the robustness certification of randomized smoothing.
    In \Cref{adxsec:alternative-design}, we present and analyze some alternative designs of \shortApproach.
    In \Cref{adx-subsec:exp}, we show the detailed experimental setup and full experiment results.
    Finally, in \Cref{adx-subsec:ablation}, we conduct abalation studies on the effects of Gradient Diversity and Confidence Margin Losses in DRT in \Cref{adx-subsec:seperate}, certified robustness of single base model within DRT-trained ensemble in \Cref{adx-subsec:singlemodels}, and investigate how optimizing ensemble weights can further improve the certified robustness of DRT-trained ensemble in \Cref{adx-subsec:optimize-weights}. We also analyze other gradient diversity promoted regularizers' performance and compare them with \shortApproach in \Cref{adx-subsec:adpgal}. 

\section{Background: Randomized Smoothing}

    \label{adx:rand-smooth-background}
    
    \citeauthor{cohen2019certified}~\citep{cohen2019certified} leverage Neyman-Pearson Lemma~\cite{neyman1933ix} to provide a computable robustness certification for the smoothed classifier.
    \begin{lemma}[Robustness Certificate of Randomized Smoothing; \citep{cohen2019certified}]
        \label{lem:rand-smooth-vanilla-certify}
        At point $\vx_0$, let random variable $\varepsilon\sim\gN(0,\sigma^2\mI_d)$, a smoothed model $G^\varepsilon_F$ is $r$-robust where
        \begin{equation}
            r = \frac{\sigma}{2} \left(
                \Phi^{-1}\left( g^\varepsilon_F(\vx_0)_{c_A} \right)
                -
                \Phi^{-1}\left( g^\varepsilon_F(\vx_0)_{c_B} \right)
            \right),
            \label{eq:rand-smooth-vanilla-r}
        \end{equation}
        where $c_A = G^\varepsilon_F(\vx_0)$ and $c_B = G^{\varepsilon(2)}_F(\vx_0)$ are the top and runner-up class respectively, and $\Phi^{-1}$ is the inverse cumulative distribution function~(CDF) of standard normal distribution.
    \end{lemma}
    In practice, for ease of sampling, the standard way of computing the certified radius is using the lower bound of \Cref{eq:rand-smooth-vanilla-r}: $r = \sigma \Phi^{-1}(g^\varepsilon_F(\vx_0)_{c_A})$.
    Now we only need to figure out $g^\varepsilon_F(\vx_0)_{c_A}$.
    The common way is to use Monte-Carlo sampling together with binomial confidence interval~\cite{cohen2019certified,yang2020randomized,zhai2019macer,jeong2020consistency}.
    Concretely, from definition $g^\varepsilon_F(\vx_0)_{c_A} = \Pr_{\varepsilon\sim\gN(0,\sigma^2\mI_d)} (F(\vx_0+\varepsilon)=c_A)$,
    we sample $n$ Gaussian noises: $\varepsilon_1,\varepsilon_2,\dots,\varepsilon_n \sim \gN(0,\sigma^2\mI_d)$ and compute the empirical mean:
    $\hat g^\varepsilon_F(\vx_0)_{c_A} = \frac{1}{n} \sum_{i=1}^n \1[F(\vx_0 + \varepsilon_i) = c_A]$.
    The binomial testing~\cite{clopper1934use} then gives a high-confidence lower bound of $g^\varepsilon_F(\vx_0)_{c_A}$ based on $\hat g^\varepsilon_F(\vx_0)_{c_A}$.
    We follow the setting in the literature: sample $n=10^5$ samples and set confidence level be $99.9\%$~\cite{cohen2019certified,yang2020randomized,zhai2019macer,jeong2020consistency}.
    More details are available in \Cref{adx-subsec:exp}.
    Note that $F$ could be either single model or any ensemble models with \beforeProtocol strategy.


\section{Detailed Analysis and Proofs in \texorpdfstring{\Cref{sec:2}}{Section 2}}
    
    \label{adxsec:sec2-theorem}
    
    In this appendix, we first show the omitted theoretical results in \Cref{sec:2}, which are the robustness conditions for \ourEnsemble~(\shortOurEnsemble) and the comparison between the robustness of ensemble model and single model.
    We then present all proofs for these theoretical results.
    
    \subsection{Detailed Theoretical Results and Discussion}
        \label{subsec:a-1}
        Here we present the theoretical results omitted from \Cref{sec:2} along with some discussions.
        
        \subsubsection{Robustness Condition of \shortOurEnsemble}
        \label{subsec:a-1-1}
        For \shortOurEnsemble, we have the following robustness condition.
        
        \begin{theorem}[Gradient and Confidence Margin Condition for \shortOurEnsemble Robustness]
            Given input $\vx_0 \in \sR^d$ with ground-truth label $y_0 \in [C]$,
            and $\Mmme$ as an \shortOurEnsemble defined over base models $\{F_1,\,F_2\}$.
            $\Mmme(\vx_0) = y_0$.
            Both $F_1$ and $F_2$ are $\beta$-smooth.
            
            \begin{itemize}
                \item (Sufficient Condition) If for any $y_1, y_2 \in [C]$ such that $y_1 \neq y_0$ and $y_2 \neq y_0$,
                \begin{equation}
                    \|\nabla_\vx f_1^{y_0/y_1}(\vx_0) + \nabla_\vx f_2^{y_0/y_2}(\vx_0)\|_2 \le
                    \frac 1 r (f_1^{y_0/y_1}(\vx_0) + f_2^{y_0/y_2}(\vx_0)) - 2\beta r,
                    \label{eq:gradient-based-sufficient-cond-our-ensemble}
                \end{equation}
                then $\Mmme$ is $r$-robust at point $\vx_0$.
                
                \item (Necessary Condition) Suppose for any $\vx \in \{\vx_0 + \vdelta: ||\vdelta||_2 \le r\}$, for any $i \in \{1,2\}$, either $F_i(\vx) = y_0$ or $F_i^{(2)}(\vx) = y_0$.
                If $\Mmme$ is $r$-robust at point $\vx_0$, then
                for any $y_1, y_2 \in [C]$ such that $y_1 \neq y_0$ and $y_2 \neq y_0$,
                \begin{equation}
                    \|\nabla_\vx f_1^{y_0/y_1}(\vx_0) + \nabla_\vx f_2^{y_0/y_2}(\vx_0)\|_2 \le
                    \frac 1 r (f_1^{y_0/y_1}(\vx_0) + f_2^{y_0/y_2}(\vx_0)) + 2\beta r.
                    \label{eq:gradient-based-necessary-cond-our-ensemble}
                \end{equation}
            \end{itemize}
            \label{thm:gradient-based-sufficient-necessary-cond-our-ensemble}
        \end{theorem}
        Comparing with the robustness conditions of \shortOurEnsemble~(\Cref{thm:gradient-based-sufficient-necessary-cond-weight-ensemble}), the conditions for \shortOurEnsemble have highly similar forms.
        Thus, the discussion for ERI in main text~(\Cref{eq:I}) still applies here, including the positive impact of diversified gradients and large confidence margins towards \shortOurEnsemble ensemble robustness in both sufficient and necessary conditions and the implication of small model-smoothness bound $\beta$.
        A major distinction is that the condition for \shortOurEnsemble is limited to two base models.
        This is because the ``maximum'' operator in \shortOurEnsemble protocol poses difficulties for expressing the robust conditions in succinct continuous functions of base models' confidence. 
        Therefore, Taylor expansion cannot be applied.
        We leave the extension to $N>2$ base models as future work, and we conjecture the tendency would be similar as \Cref{eq:gradient-based-necessary-cond-weight-ensemble}.
        The theorem is proved in \Cref{subsec:a-2}.
        
        \subsubsection{Comparison between Ensemble Robustness and Single-Model Robustness}
        
        To compare the robustness of ensemble models and single models, we have the following corollary that is extended from \Cref{thm:gradient-based-sufficient-necessary-cond-weight-ensemble} and \Cref{thm:gradient-based-sufficient-necessary-cond-our-ensemble}.
        
         \begin{corollary}[Comparison of Ensemble and Single-Model Robustness]
            Given an input $\vx_0 \in \sR^d$ with ground-truth label $y_0 \in [C]$.
            Suppose we have two $\beta$-smooth base models $\{F_1, F_2\}$, which are both $r$-robust at point $\vx_0$.
            For any $\Delta \in [0, 1)$:
            \begin{itemize}[leftmargin=*]
                \item (\weightedEnsemble)
                Define \weightedEnsemble $\Mweight$ with base models $\{F_1, F_2\}$.
                Suppose $\Mweight(\vx_0) = y_0$.
                If for any label $y_i \neq y_0$,
                the base models' smoothness
                $\beta \le \Delta \cdot \min\{f_1^{y_0/y_i}(\vx_0), f_2^{y_0/y_i}(\vx_0)\} / (c^2 r^2)$,
                and the gradient cosine similarity $\cos \langle \nabla_\vx f_1^{y_0/y_i}(\vx_0), \nabla_\vx f_2^{y_0/y_i}(\vx_0) \rangle \le \cos \theta$, 
                then the $\Mweight$ with weights $\{w_1, w_2\}$ is at least $R$-robust at point $\vx_0$ with
                \begin{equation}
                    \begin{small}
                        R = r \cdot \frac{1 - \Delta}{1 + \Delta} \left( 1 - C_{\mathrm{\shortWeightedEnsemble}} (1 - \cos\theta) \right)^{-\nicefrac{1}{2}}, \text{where} 
                    \end{small}
                    \label{eq:weighted-ensemble-robust-radius-lower-bound}
                \end{equation}
                $C_{\mathrm{\shortWeightedEnsemble}} = \underset{y_i: y_i \neq y_0}{\min} \frac{2w_1w_2 f_1^{y_0/y_i}(\vx_0) f_2^{y_0/y_i}(\vx_0)}{(w_1f_1^{y_0/y_i}(\vx_0) + w_2f_2^{y_0/y_i}(\vx_0))^2},
                c = \max\{\frac{1 - \Delta}{1 + \Delta} \left( 1 - C_{\mathrm{\shortWeightedEnsemble}} (1 - \cos\theta) \right)^{-\nicefrac{1}{2}}, 1\}.$
                
                \item (\ourEnsemble)
                Define \ourEnsemble $\Mmme$ with the base models $\{F_1, F_2\}$.
                Suppose $\Mmme(\vx_0) = y_0$.
                If for any label $y_1 \neq y_0$ and $y_2 \neq y_0$,
                the base models' smoothness
                $\beta \le \Delta \cdot \min\{f_1^{y_0/y_1}(\vx_0), f_2^{y_0/y_2}(\vx_0)\} / (c^2 r^2)$,
                and the gradient cosine similarity $\cos \langle \nabla_\vx f_1^{y_0/y_1}(\vx_0), \nabla_\vx f_2^{y_0/y_2}(\vx_0) \rangle \le \cos \theta$, then the $\Mmme$ is at least $R$-robust at point $\vx_0$ with
                \begin{equation}
                \begin{small}
                   R = r \cdot \frac{1 - \Delta}{1 + \Delta} \left( 1 - C_{\mathrm{\shortOurEnsemble}} (1 - \cos\theta) \right)^{-\nicefrac{1}{2}},
                   \text{where}
                    \label{eq:max-margin-ensemble-robust-radius-lower-bound}
                    \end{small}
                \end{equation}
                $C_{\mathrm{\shortOurEnsemble}} = \underset{{\substack{y_1, y_2:\\ y_1, y_2 \neq y_0}}}{\min} \frac{2f_1^{y_0/y_1}(\vx_0) f_2^{y_0/y_2}(\vx_0)}{(f_1^{y_0/y_1}(\vx_0) + f_2^{y_0/y_2}(\vx_0))^2},
                c = \max\{ \frac{1 - \Delta}{1 + \Delta} \left( 1 - C_{\mathrm{\shortOurEnsemble}} (1 - \cos\theta) \right)^{-\nicefrac{1}{2}}, 1\}.$
            \end{itemize}
            \label{cor:robust-radius-lower-bound-for-ensemble-constructed-from-single-models}
        \end{corollary}
        The proof is given in \Cref{subsec:a-3}.
        
        
        \paragraph{Optimizing \weightedEnsemble.}
        As we can observe from \Cref{cor:robust-radius-lower-bound-for-ensemble-constructed-from-single-models}, we can adjust the weights $\{w_1,w_2\}$ for \weightedEnsemble to change $C_{\mathrm{\shortWeightedEnsemble}}$ and the certified robust radius~(\Cref{eq:weighted-ensemble-robust-radius-lower-bound}).
        Then comes the problem of which set of weights can achieve the highest certified robust radius.
        Since larger $C_{\mathrm{\shortWeightedEnsemble}}$ results in higher radius, we need to choose
        $$
            (w_1^{OPT}, w_2^{OPT}) = \arg\max_{w_1,w_2} \min_{y_i: y_i \neq y_0} \dfrac{2w_1 w_2 f_1^{y_0/y_i}(\vx_0) f_2^{y_0/y_i}(\vx_0)}{(w_1f_1^{y_0/y_i}(\vx_0) + w_2f_2^{y_0/y_i}(\vx_0))^2}.
        $$
        Since this quantity is scale-invariant, we can fix $w_1$ and optimize over $w_2$ to get the optimal weights.
        In particular, if there are only two classes, we have a closed-form solution
        $$
        \begin{aligned}
            (w_1^{OPT}, w_2^{OPT}) & = \arg\max_{w_1,w_2} \dfrac{2w_1 w_2 f_1^{y_0/y_1}(\vx_0) f_2^{y_0/y_1}(\vx_0)}{(w_1f_1^{y_0/y_1}(\vx_0) + w_2f_2^{y_0/y_1}(\vx_0))^2} \\
            & = \{ k \cdot f_2^{y_0/y_1}(\vx_0), k \cdot f_1^{y_0/y_1}(\vx_0): k \in \sR_{+} \},
        \end{aligned}
        $$
        and corresponding $C_{\mathrm{\shortWeightedEnsemble}}$ achieves the maximum $1/2$.
        
        For a special case---average weighted ensemble, we get the corresponding certified robust radius by setting $w_1 = w_2$ and plug the yielded 
        $$
            C_{\mathrm{\shortWeightedEnsemble}} = \min_{y_i: y_i \neq y_0} \dfrac{2 f_1^{y_0/y_i}(\vx_0) f_2^{y_0/y_i}(\vx_0)}{(f_1^{y_0/y_i}(\vx_0) + f_2^{y_0/y_i}(\vx_0))^2} \in (0, 1/2].
        $$ into \Cref{eq:weighted-ensemble-robust-radius-lower-bound}.
        
        \paragraph{Comparison between ensemble and single-model robustness.}
        The similar forms of $R$ in the corollary allow us to discuss the \weightedEnsemble and \ourEnsemble together.
        Specifically, we let $C$ be either $C_{\mathrm{\shortWeightedEnsemble}}$ or $C_{\mathrm{\shortOurEnsemble}}$, then
        $$
            R = r \cdot \dfrac{1-\Delta}{1+\Delta} \left(1 - C(1-\cos\theta)\right)^{-\nicefrac{1}{2}}.
        $$
        Since when $R > r$, both ensembles have higher certified robustness than the base models, we solve this condition for $\cos\theta$:
        $$
        \begin{aligned}
             R > r 
            \iff  \left( \dfrac{1-\Delta}{1+\Delta} \right)^2 >
            1 - C(1-\cos\theta) 
            \iff  \cos\theta \le 1 - \dfrac{4\Delta}{C(1+\Delta)^2}.
        \end{aligned}
        $$
        Notice that $C \in (0, 1/2]$.
        From this condition, we can easily observe that when the gradient cosine similarity is smaller, it is more likely that the ensemble has higher certified robustness than the base models.
        When the model is smooth enough, according to the condition on $\beta$, we can notice that $\Delta$ could be close to zero.
        As a result, $1 - \frac{4\Delta}{C(1+\Delta)^2}$ is close to $1$.
        \textit{Thus, unless the gradient of base models is (or close to) colinear, it always holds that the ensemble~(either \shortWeightedEnsemble or \shortOurEnsemble) has higher certified robustness than the base models.}
        
        \paragraph{Larger certified radius with larger number of base models $N$.}
        Following the same methodology, we can further observe that larger number of base models $N$ can lead to larger certified radius as the following proposition shows.
        \begin{adxproposition}[More Base Models Lead to Higher Certified Robustness of \weightedEnsemble]
            At clean input $\vx_0 \in \sR^d$ with ground-truth label $y_0 \in [C]$, suppose all base models $\{f_i\}_{i=1}^{N+M}$ are $\beta$-smooth. 
            Suppose the \weightedEnsemble $\gM_1$ of base models $\{f_i\}_{i=1}^N$ and $\gM_2$ of base models $\{f_i\}_{i=N+1}^{N+M}$ are both $r$-robust according to the sufficient condition in \Cref{thm:gradient-based-sufficient-necessary-cond-weight-ensemble},
            and for any $y_i\neq y_0$ the $\gM_1$ and $\gM_2$'s ensemble gradients~($\sum_{j=1}^N w_j \nabla_\vx f_j^{y_0/y_i}(\vx_0)$ and $\sum_{j=N+1}^{N+M} w_j \nabla_\vx f_j^{y_0/y_i}(\vx_0)$) are non-zero and not colinear, then the \weightedEnsemble $\gM$ of $\{f_i\}_{i=1}^{N+M}$ is $r'$-robust for some $r' > r$.
            \label{prop:larger-n-better-robustness}
        \end{adxproposition}
        
        \begin{proof}[Proof of \Cref{prop:larger-n-better-robustness}]
            For any $y_i \neq y_0$, since both $\gM_1$ and $\gM_2$ are $r$-robust according to the sufficient condition of \Cref{thm:gradient-based-sufficient-necessary-cond-weight-ensemble}, we have
            \begin{equation}
                \Big\| \sum_{j=1}^N w_j \nabla_{\vx} f_j^{y_0/y_i}(\vx_0) \Big\|_2 \le \dfrac{1}{r} \sum_{j=1}^N f_j^{y_0/y_i} (\vx_0) - \beta r \sum_{j=1}^N w_j,
            \end{equation}
            \begin{equation}
                \Big\| \sum_{j=N+1}^{N+M} w_j \nabla_{\vx} f_j^{y_0/y_i}(\vx_0) \Big\|_2 \le \dfrac{1}{r} \sum_{j=N+1}^{N+M} f_j^{y_0/y_i} (\vx_0) - \beta r \sum_{j=N+1}^{N+M} w_j.
            \end{equation}
            Adding above two inequalties we get
            \begin{equation}
                \Big\| \sum_{j=1}^N w_j \nabla_{\vx} f_j^{y_0/y_i}(\vx_0) \Big\|_2 + \Big\| \sum_{j=N+1}^{N+M} w_j \nabla_{\vx} f_j^{y_0/y_i}(\vx_0) \Big\|_2
                \le
                \dfrac{1}{r} \sum_{j=1}^{N+M} f_j^{y_0/y_i} (\vx_0) - \beta r \sum_{j=1}^{N+M} w_j.
            \end{equation}
            Since gradients of ensemble are not colinear and non-zero, from the triangle inequality,
            \begin{equation}
                \Big\| \sum_{j=1}^{N+M} w_j \nabla_{\vx} f_j^{y_0/y_i}(\vx_0) \Big\|_2
                < \Big\| \sum_{j=1}^N w_j \nabla_{\vx} f_j^{y_0/y_i}(\vx_0) \Big\|_2 + \Big\| \sum_{j=N+1}^{N+M} w_j \nabla_{\vx} f_j^{y_0/y_i}(\vx_0) \Big\|_2
            \end{equation}
            and thus
            \begin{equation}
                \Big\| \sum_{j=1}^{N+M} w_j \nabla_{\vx} f_j^{y_0/y_i}(\vx_0) \Big\|_2 < \dfrac{1}{r} \sum_{j=1}^{N+M} f_j^{y_0/y_i} (\vx_0) - \beta r \sum_{j=1}^{N+M} w_j,
            \end{equation}
            which means we can increase $r$ to $r'$ and still keep the inequality hold with ``$\le$'', and in turn certify a larger radius $r'$ according to \Cref{thm:gradient-based-sufficient-necessary-cond-weight-ensemble}.
        \end{proof}
        
        Since \shortApproach imposes diversified gradients via \shortLHSloss, the ``not colinear'' condition easily holds for \shortApproach ensemble, and therefore the proposition implies larger number of base models $N$ lead to higher certified robustness of \shortWeightedEnsemble.
        For \shortOurEnsemble, we empirically observe similar trends.
        
        
        \subsubsection{Robustness Conditions for Smoothed Ensemble Models}
        \label{adxsubsec:robust-cond-smooth-ensemble}
        
        Following the discussion in \Cref{subsec:2-3}, for \textit{smoothed} \shortWeightedEnsemble and \shortOurEnsemble, with the model-smoothness bound~(\Cref{thm:smoothness-bound}), we can concretize the general robustness conditions in this way.
        
        
        
        We define the \textit{soft} smoothed confidence function $\bar g^\varepsilon_f(\vx) := \E_\varepsilon f(\vx + \varepsilon)$.
        This definition is also used in the literature~\citep{salman2019provably,zhai2019macer,aounan2020certifying}.
        As a result, we revise the ensemble protocols of \shortWeightedEnsemble and \shortOurEnsemble by replacing the original confidences $\{f_i(\vx_0)\}_{i=1}^N$ with these soft smoothed confidences $\{\bar g_i^\varepsilon(\vx_0)\}_{i=1}^N$.
        These protocols then choose the predicted class by treating these $\{\bar g_i^\varepsilon(\vx_0)\}_{i=1}^N$ as the base models' confidence scores.
        In the experiments, we did not actually evaluate these revised protocols since their robustness performance are expected to be similar as original ones~\citep{salman2019provably,zhai2019macer,aounan2020certifying}.
        The derived results connect smoothed ensemble robustness with the confidence scores.

        \begin{corollary}[Gradient and Confidence Margin Conditions for Smoothed \shortWeightedEnsemble Robustness]
            \label{cor:gradient-based-sufficient-necessary-cond-smoothed-weight-ensemble}
            Given input $\vx_0 \in \sR^d$ with ground-truth label $y_0 \in [C]$.
            Let $\varepsilon\sim\gN(0, \sigma^2\mI_d)$ be a Gaussian random variable.
            Define soft smoothed confidence $\bar g^\varepsilon_i(\vx) := \E_\varepsilon f_i(\vx + \varepsilon)$ for each base model $F_i$~($1\le i\le N$).
            The $\bar G^\varepsilon_\Mweight$ is a \shortWeightedEnsemble defined over soft smoothed base models $\{\bar g^\varepsilon_i\}_{i=1}^N$ with weights $\{w_i\}_{i=1}^N$.
            $\bar G^\varepsilon_\Mweight(\vx_0) = y_0$.
            \begin{itemize}
                \vspace{-0.5em}
                \item (Sufficient Condition) The $\bar G^\varepsilon_{\Mweight}$ is $r$-robust at point $\vx_0$ if for any $y_i \neq y_0$,
                \begin{equation}
                    \hspace{-1em}
                    \Big\|\sum_{j=1}^N w_j \nabla_\vx (\bar g_j^{\varepsilon})^{y_0/y_i}(\vx_0) \Big\|_2 \le \frac 1 r \sum_{j=1}^N w_j (\bar g_j^\varepsilon)^{y_0/y_i}(\vx_0) - \frac{2r}{\sigma^2} \sum_{j=1}^N w_j,
                    \label{eq:gradient-based-sufficient-cond-smoothed-weight-ensemble}
                \end{equation}
                \item (Necessary Condition) If $\bar G^\varepsilon_{\Mweight}$ is $r$-robust at point $\vx_0$,  for any $y_i \neq y_0$,
                \begin{equation}
                    \hspace{-1em}
                    \Big\|\sum_{j=1}^N w_j \nabla_\vx (\bar g^\varepsilon_j)^{y_0/y_i}(\vx_0) \Big\|_2 \le \frac 1 r \sum_{j=1}^N w_j (\bar g^\varepsilon_j)^{y_0/y_i}(\vx_0) + \dfrac{2r}{\sigma^2} \sum_{j=1}^N w_j.
                    \label{eq:gradient-based-necessary-cond-smoothed-weight-ensemble}
                \end{equation}
            \end{itemize}
        \end{corollary}
        
        \begin{corollary}[Gradient and Confidence Margin Condition for Smoothed \shortOurEnsemble Robustness]
            \label{cor:gradient-based-sufficient-necessary-cond-smoothed-our-ensemble}
            Given input $\vx_0 \in \sR^d$ with ground-truth label $y_0 \in [C]$.
            Let $\varepsilon\sim\gN(0, \sigma^2\mI_d)$ be a Gaussian random variable.
            Define soft smoothed confidence $\bar g^\varepsilon_i(\vx) := \E_\varepsilon f_i(\vx + \varepsilon)$ for either base model $F_1$ or $F_2$.
            The $\bar G^\varepsilon_\Mmme$ is a \shortOurEnsemble defined over soft smoothed base models $\{\bar g^\varepsilon_1,\bar g^\varepsilon_2\}$.
            $\bar G^\varepsilon_\Mmme(\vx_0) = y_0$.
            
            \begin{itemize}
                \item (Sufficient Condition) If for any $y_1, y_2 \in [C]$ such that $y_1 \neq y_0$ and $y_2 \neq y_0$,
                \begin{equation}
                    \|\nabla_\vx (\bar g_1^\varepsilon)^{y_0/y_1}(\vx_0) + \nabla_\vx (\bar g_2^\varepsilon)^{y_0/y_2}(\vx_0)\|_2 \le
                    \frac 1 r ( (\bar g_1^\varepsilon)^{y_0/y_1}(\vx_0) + (\bar g_2^\varepsilon)^{y_0/y_2}(\vx_0)) - \dfrac{4r}{\sigma^2},
                    \label{eq:gradient-based-sufficient-cond-smoothed-our-ensemble}
                \end{equation}
                then $\bar G^\varepsilon_\Mmme$ is $r$-robust at point $\vx_0$.
                
                \item (Necessary Condition) Suppose for any $\vx \in \{\vx_0 + \vdelta: ||\vdelta||_2 \le r\}$, for any $i \in \{1,2\}$, either $G_{F_i}(\vx) = y_0$ or $G_{F_i}^{(2)}(\vx) = y_0$.
                If $\bar G^\varepsilon_\Mmme$ is $r$-robust at point $\vx_0$, then
                for any $y_1, y_2 \in [C]$ such that $y_1 \neq y_0$ and $y_2 \neq y_0$,
                \begin{equation}
                    \|\nabla_\vx (\bar g_1^\varepsilon)^{y_0/y_1}(\vx_0) + \nabla_\vx (\bar g_2^\varepsilon)^{y_0/y_2}(\vx_0)\|_2 \le
                    \frac 1 r ((\bar g_1^\varepsilon)^{y_0/y_1}(\vx_0) + (\bar g_2^\varepsilon)^{y_0/y_2}(\vx_0)) + \dfrac{4r}{\sigma^2}.
                    \label{eq:gradient-based-necessary-cond-smoothed-our-ensemble}
                \end{equation}
            \end{itemize}
        \end{corollary} 
        \begin{remark}
            The above two corollaries are extended from \Cref{thm:gradient-based-sufficient-necessary-cond-weight-ensemble} and \Cref{thm:gradient-based-sufficient-necessary-cond-our-ensemble} respectively, and correspond to our discussion in \Cref{subsec:2-3}.
            We defer the proofs to \Cref{subsec:a-5}.
            From these two corollaries, we can explicit see that the \beforeProtocol~(see \Cref{def:ebs}) provides smoothed classifiers $\bar G^\varepsilon_{\Mweight}$ and $\bar G^\varepsilon_{\Mmme}$ with bounded model-smoothness; and we can see the correlation between the robustness conditions and gradient diversity/confidence margin for smoothed ensembles.
        \end{remark}
    
    \subsection{Proofs of Robustness Conditions for General Ensemble Models}
        \label{subsec:a-2}
        
        This subsection contains the proofs of robustness conditions.
        First, we connect the prediction of ensemble models with the arithmetic relations of confidence scores of base models.
        This connection is straightforward to establish for \weightedEnsemble~(shown in \Cref{prop:sufficient-necessary-condition-weighted-ensemble}), but nontrivial for \ourEnsemble~(shown in \Cref{thm:sufficient-necessary-condition-maximum-margin-ensemble}).
        Then, we prove the desired robustness conditions using Taylor expansion with Lagrange reminder.
        
        \begin{adxproposition}[Robustness Condition for \shortWeightedEnsemble]
            Consider an input $\vx_0 \in \sR^d$ with ground-truth label $y_0 \in [C]$, and
             an ensemble model $\Mweight$  constructed by base models $\{F_i\}_{i=1}^N$ with weights $\{w_i\}_{i=1}^N$.
            Suppose $\Mweight(\vx_0) = y_0$.
            Then, the ensemble $\Mweight$ is $r$-robust at point $\vx_0$ if and only if for any $\vx \in \{\vx_0 + \vdelta: \|\vdelta\|_2 \le r\}$,
            \begin{equation}
                \min_{y_i \in [C]: y_i \neq y_0} \sum_{j=1}^N w_j f_j^{y_0/y_i}(\vx) \ge 0.
                \tag{\ref{eq:prop-1}}
            \end{equation}
            \label{prop:sufficient-necessary-condition-weighted-ensemble}
        \end{adxproposition}
        
        \begin{proof}[Proof of \Cref{prop:sufficient-necessary-condition-weighted-ensemble}]
            According the the definition of $r$-robust,
            we know $\Mweight$ is $r$-robust if and only if for any point $\vx := \vx_0 + \vdelta$ where $\|\vdelta\|_2 \le r$, $\Mweight(\vx_0 + \vdelta) = y_0$, which means that for any other label $y_i \neq y_0$, the confidence score for label $y_0$ is larger or equal than the confidence score for label $y_i$.
            It means that
            $$
                \sum_{j=1}^N w_j f_j(\vx)_{y_0} \ge \sum_{j=1}^N w_j f_j(\vx)_{y_i}
            $$
            for any $\vx \in \{\vx_0 + \vdelta: \|\vdelta\|_2 \le r\}$.
            Since this should hold for any $y_i \neq y_0$, we have the sufficient and necessary condition
            \begin{equation}
                \min_{y_i \in [C]: y_i \neq y_0} \sum_{j=1}^N w_j f_j^{y_0/y_i}(\vx) \ge 0.
                \label{eq:prop-1}
            \end{equation}
        \end{proof}
        
        \begin{theorem}[Robustness Condition for \shortOurEnsemble]
            Consider an input $\vx_0 \in \sR^d$ with ground-truth label $y_0 \in [C]$.
            Let $\Mmme$ be an \shortOurEnsemble defined over base models $\{F_i\}_{i=1}^N$.
            Suppose:
            (1)~$\Mmme(\vx_0) = y_0$;
            (2)~for any $\vx \in \{\vx_0 + \vdelta: \|\vdelta\|_2 \le r\}$, given any base model $i \in [N]$, either $F_i(\vx) = y_0$ or $F_i^{(2)}(\vx) = y_0$.
            Then, the ensemble $\Mmme$ is $r$-robust at point $\vx_0$ \emph{if and only if}
            for any $\vx \in \{\vx_0 + \vdelta: \|\vdelta\|_2 \le r\}$,
            \begin{equation}
                \max_{i \in [N]} \min_{y_i \in [C]: y_i \neq y_0} f_i^{y_0/y_i} (\vx)
                \ge
                \max_{i \in [N]} \min_{y_i' \in [C]: y_i' \neq y_0} f_i^{y_i'/y_0} (\vx).
                \label{eq:thm2}
            \end{equation}
            \label{thm:sufficient-necessary-condition-maximum-margin-ensemble}
        \end{theorem}
        
        The theorem states the sufficient and necessary robustness condition for \shortOurEnsemble.
        We divide the two directions into the following two lemmas and prove them separately.
        We mainly use the alternative form of \Cref{eq:thm2} as such in the following lemmas and their proofs:
        \begin{equation}
            \max_{i \in [N]} \min_{y_i \in [C]: y_i \neq y_0} f_i^{y_0/y_i} (\vx)
            +
            \min_{i \in [N]} \min_{y_i' \in [C]: y_i' \neq y_0} f_i^{y_0/y_i'} (\vx)
            \ge
            0.
            \tag{\ref{eq:thm2}}
        \end{equation}
        
        \begin{lemma}[Sufficient Condition for \shortOurEnsemble]
            \label{adxlem:adx-suf-cond-for-ensemble-rob}
            Let $\Mmme$ be an \shortOurEnsemble defined over base models $\{F_i\}_{i=1}^N$.
            For any input $\vx_0 \in \sR^d$,
            the \ourEnsemble $\Mmme$ predicts $\Mmme(\vx_0) = y_0$ 
            \emph{if}
            \begin{equation}
                \max_{i \in [N]} \min_{y_i \in [C]: y_i \neq y_0} f_i^{y_0/y_i} (\vx_0)
                +
                \min_{i \in [N]} \min_{y_i' \in [C]: y_i' \neq y_0} f_i^{y_0/y_i'} (\vx_0)
                \ge
                0.
                \tag{\ref{eq:thm2}}  
            \end{equation}
        \end{lemma}
        
        \begin{proof}[Proof of \Cref{adxlem:adx-suf-cond-for-ensemble-rob}]
            For brevity, for $i \in [N]$, we denote $y_i := F_i(\vx_0), y_i' := F_i^{(2)}(\vx_0)$ for each base model's top class and runner-up class at point $\vx_0$. 
            
            Suppose $\Mmme(\vx_0) \neq y_0$, then according to ensemble definition~(see \Cref{def:maximum-margin-ensemble}), there exists $c \in [N]$, such that $\Mmme(\vx_0) = F_c(\vx_0) = y_c$, and
            \begin{equation}
                \forall i \in [N],\, i \neq c,\, f_c(\vx_0)^{y_c/y_c'} > f_i(\vx_0)^{y_i/y_i'}.
                \label{adx-eq:lem1-t1}
            \end{equation}
            Because $y_c \neq y_0$, we have $f_c(\vx_0)_{y_0} \le f_c(\vx_0)_{y_c'}$, so that $f_c(\vx_0)^{y_c/y_0} \ge f_c(\vx_0)^{y_c/y_c'}$.
            Now consider any model $F_i$ where $i\in [N]$,
            we would like to show that there exists $y^* \neq y_0$, such that $f_i(\vx_0)^{y_i/y_i'} \ge f_i(\vx_0)^{y_0/y^*}$:
            \begin{itemize}
                \item If $y_i = y_0$, let $y^* := y_i'$, trivially $f_i(\vx_0)^{y_i/y_i'} = f_i(\vx_0)^{y_0/y^*}$;
                
                \item If $y_i \neq y_0$, and $y_i' \neq y_0$, we let $y^* := y_i'$, then $f_i(\vx_0)^{y_i/y_i'} = f_i(\vx_0)^{y_i/y^*} \ge f_i(\vx_0)^{y_0/y^*}$;
                
                \item If $y_i \neq y_0$, but $y_i' = y_0$, we let $y^* := y_i$, then $f_i(\vx_0)^{y_i/y_i'} = f_i(\vx_0)^{y_i/y_0} \ge f_i(\vx_0)^{y_0/y_i} = f_i(\vx_0)^{y_0/y^*}$.
            \end{itemize}
            Combine the above findings with \Eqref{adx-eq:lem1-t1}, we have:
            $$
                \forall i \in [N],\, i \neq c,\, \exists y^*_c \in [C] \,\mathrm{and}\, y^*_c \neq y_0,\, \exists y^*_i \in [C] \,\mathrm{and}\, y^*_i \neq y_0,\,
                f_c(\vx_0)^{y^*_c/y_0} > f_i(\vx_0)^{y_0/y^*_i}.
            $$
            
            Therefore, its negation
            \begin{equation}
                \exists i \in [N],\, i \neq c,\, \forall y^*_c \in [C] \,\mathrm{and}\, y^*_c \neq y_0,\, \forall y^*_i \in [C] \,\mathrm{and}\, y^*_i \neq y_0,\,
                f_c(\vx_0)^{y_0/y^*_c} + f_i(\vx_0)^{y_0/y^*_i} \ge 0
                \label{adx-eq:lem-1-t1}
            \end{equation}
            implies $\gM(\vx_0) = y_0$.
            Since \Cref{adx-eq:lem-1-t1} holds for any $y_c^*$ and $y_i^*$, the equation is equivalent to
            $$
                \exists i \in [N],\, i \neq c,\,
                \min_{y_c\in [C]: y_c \neq y_0} f_c(\vx_0)^{y_0/y_c}(\vx_0) + \min_{y_i'\in [C]: y_i' \neq y_0} f_i(\vx_0)^{y_0/y_i'}(\vx_0) \ge 0.
            $$
            The existence qualifier over $i$ can be replaced by maximum:
            $$
                \min_{y_c\in [C]: y_c \neq y_0} f_c(\vx_0)^{y_0/y_c}(\vx_0) + \max_{i\in [N]} \min_{y_i'\in [C]: y_i' \neq y_0} f_i(\vx_0)^{y_0/y_i'}(\vx_0) \ge 0.
            $$
            It is implied by
            \begin{equation}
                \max_{i \in [N]} \min_{y_i \in [C]: y_i \neq y_0} f_i^{y_0/y_i} (\vx_0)
                +
                \min_{i \in [N]} \min_{y_i' \in [C]: y_i' \neq y_0} f_i^{y_0/y_i'} (\vx_0)
                \ge
                0.
                \tag{\ref{eq:thm2}}
            \end{equation}
            Thus, \Cref{eq:thm2} is a sufficient condition for $\Mmme(\vx_0) = y_0$.
        \end{proof}
        
        \begin{lemma}[Necessary Condition for \shortOurEnsemble]
            \label{adxlem:adx-necs-cond-for-ensemble-rob}
            For any input $\vx_0 \in \sR^d$, if for any base model $i \in [N]$, either $F_i(\vx_0) = y_0$ or $F_i^{(2)}(\vx_0) = y_0$, then
            \ourEnsemble $\Mmme$ predicting $\Mmme(\vx_0) = y_0$ implies
            \begin{equation}
                \max_{i \in [N]} \min_{y_i \in [C]: y_i \neq y_0} f_i^{y_0/y_i} (\vx_0)
                +
                \min_{i \in [N]} \min_{y_i' \in [C]: y_i' \neq y_0} f_i^{y_0/y_i'} (\vx_0)
                \ge
                0.
                \tag{\ref{eq:thm2}}
            \end{equation}
        \end{lemma}
        
        \begin{proof}[Proof of \Cref{adxlem:adx-necs-cond-for-ensemble-rob}]
            Similar as before, for brevity, for $i \in [N]$, we denote $y_i := F_i(\vx_0), y_i' := F_i^{(2)}(\vx_0)$ for each base model's top class and runner-up class at point $\vx_0$.
            
            Suppose \Cref{eq:thm2} is not satisfied, it means that
            $$
                \exists c \in [N],\,
                \exists y_c^* \in [C] \,\mathrm{and}\, y_c^* \neq y_0,\,
                \forall i \in [N],\,
                \exists y_i^* \in [C] \,\mathrm{and}\, y_i^* \neq y_0,\,
                f_c^{y_c^*/y_0}(\vx_0) > f_i^{y_0/y_i^*}(\vx_0).
            $$
            
            \begin{itemize}
                \item
                If $y_c = y_0$, then $f_c^{y_c^*/y_0}(\vx_0) \le 0$, which implies that $f_i^{y_0/y_i^*}(\vx_0) < 0$, and hence $F_i(\vx_0) \neq y_0$.
                Moreover, we know that $f_i^{y_i/y_i'}(\vx_0) = f_i^{y_i/y_0}(\vx_0) \ge f_i^{y_i^*/y_0}(\vx_0) > f_c^{y_0/y_c^*}(\vx_0) \ge f_c^{y_0/y_c'}(\vx_0) = f_c^{y_c/y_c'}(\vx_0)$ so $\gM(\vx_0) \neq F_c(\vx_0) = y_0$.
                
                \item
                If $y_c \neq y_0$, i.e., $y_c' = y_0$, then $f_c^{y_c/y_0}(\vx_0) \ge f_c^{y_c^*/y_0}(\vx_0) > f_i^{y_0/y_1^*}(\vx_0)$.
                If $F_i(\vx_0) = y_0$,
                then
                $f_i^{y_0/y_i^*}(\vx_0) \ge f_i^{y_0/y_i'}(\vx_0) = f_i^{y_i/y_i'}(\vx_0)$.
                Thus,
                $f_c^{y_c/y_c'}(\vx_0) = f_c^{y_c/y_0}(\vx_0) > f_i^{y_i/y_i'}(\vx_0)$.
                As the result, $\gM(\vx_0) = F_c(\vx_0) \neq y_0$.
            \end{itemize}
            For both cases, we show that $\Mmme(\vx_0) \neq y_0$, i.e., \Cref{eq:thm2} is a necessary condition for $\gM(\vx_0) = y_0$.
        \end{proof}
        
        \begin{proof}[Proof of \Cref{thm:sufficient-necessary-condition-maximum-margin-ensemble}]
            \Cref{adxlem:adx-necs-cond-for-ensemble-rob,adxlem:adx-suf-cond-for-ensemble-rob} are exactly the two directions~(necessary and sufficient condition) of $\Mmme$ predicting label $y_0$ at point $\vx$. 
            Therefore, if the condition~(\Cref{eq:thm2}) holds for any $\vx \in \{\vx_0 + \vdelta: \|\vdelta\|_2 \le r\}$, 
            the ensemble $\Mmme$ is $r$-robust at point $\vx_0$; vice versa.
        \end{proof}
        
        For comparison, here we list the trivial robustness condition for single model.
        \begin{fact}[Robustness Condition for Single Model]
            \label{adx-fact:robust-condition-for-single-model}
            Consider an input $\vx_0 \in \sR^d$ with ground-truth label $y_0 \in [C]$.
            Suppose a model $F$ satisfies $F(\vx_0) = y_0$.
            Then, the model $F$ is $r$-robust at point $\vx_0$ if and only if for any $\vx \in \{\vx_0 + \vdelta: \|\vdelta\|_2 \le r\}$,
            $$
                \min_{y_i \in [C]: y_i \neq y_0} f^{y_0/y_i}(\vx) \ge 0.
            $$
        \end{fact}
        The fact is apparent given that the model predicts the class with the highest confidence.
        
        \vspace{1em}
        Now we are ready to apply Taylor expansion to derive the robustness conditions shown in main text.
    
        \begin{customthm}{\ref{thm:gradient-based-sufficient-necessary-cond-weight-ensemble}}[Gradient and Confidence Margin Condition for \shortWeightedEnsemble Robustness]
            Given input $\vx_0 \in \sR^d$ with ground-truth label $y_0 \in [C]$,
            and $\Mweight$ as a \shortWeightedEnsemble defined over base models $\{F_i\}_{i=1}^N$ with weights $\{w_i\}_{i=1}^N$.
            $\Mweight(\vx_0) = y_0$.
            All base model $F_i$'s are $\beta$-smooth.
            \begin{itemize}
                \item (Sufficient Condition) $\Mweight$ is $r$-robust at point $\vx_0$ if for any $y_i \neq y_0$,
                \begin{equation}
                    \Big\|\sum_{j=1}^N w_j \nabla_\vx f_j^{y_0/y_i}(\vx_0) \Big\|_2 \le \frac 1 r \sum_{j=1}^N w_j f_j^{y_0/y_i}(\vx_0) - \beta r \sum_{j=1}^N w_j.
                    \tag{\ref{eq:gradient-based-sufficient-cond-weight-ensemble}}
                \end{equation}
                
                \item (Necessary Condition) If $\Mweight$ is $r$-robust at point $\vx_0$, then for any $y_i \neq y_0$,
                \begin{equation}
                    \Big\|\sum_{j=1}^N w_j \nabla_\vx f_j^{y_0/y_i}(\vx_0) \Big\|_2 \le \frac 1 r \sum_{j=1}^N w_j f_j^{y_0/y_i}(\vx_0) + \beta r \sum_{j=1}^N w_j.
                    \tag{\ref{eq:gradient-based-necessary-cond-weight-ensemble}}
                \end{equation}
            \end{itemize}
        \end{customthm}
        
        \begin{proof}[Proof of \Cref{thm:gradient-based-sufficient-necessary-cond-weight-ensemble}]
            From Taylor expansion with Lagrange remainder and the $\beta$-smoothness assumption on the base models, we have
            \begin{equation}
                \begin{aligned}
                    & \sum_{j=1}^N w_j f_j^{y_0/y_i}(\vx_0) - r \Big\|\sum_{j=1}^N w_j \nabla_\vx f_j^{y_0/y_i}(\vx_0) \Big\|_2 - \dfrac{1}{2} r^2 \sum_{j=1}^N (2\beta w_j)
                    \le
                    \min_{\vx: \|\vx - \vx_0\|_2 \le r}
                    \sum_{j=1}^N w_j f_j^{y_0/y_i}(\vx)
                    \\
                    & \hspace{5em} 
                    \le \sum_{j=1}^N w_j f_j^{y_0/y_i}(\vx_0) - r \Big\|\sum_{j=1}^N w_j \nabla_\vx f_j^{y_0/y_i}(\vx_0) \Big\|_2 + \dfrac{1}{2} r^2 \sum_{j=1}^N (2\beta w_j),
                \end{aligned}
                \label{adx-eq:thm3-1}
            \end{equation}
            where the term $- \dfrac{1}{2} r^2 \sum_{j=1}^N (2\beta w_j)$ and $\dfrac{1}{2} r^2 \sum_{j=1}^N (2\beta w_j)$ are bounded from Lagrange remainder.
            Note that the difference $f_j^{y_0/y_i}$ is $(2\beta)$-smooth instead of $\beta$-smooth since it is the difference of two $\beta$-smooth function, and thus $\sum_{j=1}^N w_j f_j^{y_0/y_i}$ is $\sum_{j=1}^N (2\beta w_j)$-smooth.
            From \Cref{prop:sufficient-necessary-condition-weighted-ensemble}, the sufficient and necessary condition of \shortWeightedEnsemble's $r$-robustness is $\sum_{j=1}^N w_jf_j^{y_0/y_i}(\vx)\ge 0$ for any $y_i \in [C]$ such that $y_i \neq y_0$, and any $\vx = \vx_0 + \vdelta$ where $\|\vdelta\|_2 \le r$.
            Plugging this term into \Cref{adx-eq:thm3-1} we get the theorem.
        \end{proof}
        
        \begin{customthm}{\ref{thm:gradient-based-sufficient-necessary-cond-our-ensemble}}[Gradient and Confidence Margin Condition for \shortOurEnsemble Robustness]
            Given input $\vx_0 \in \sR^d$ with ground-truth label $y_0 \in [C]$,
            and $\Mmme$ as an \shortOurEnsemble defined over base models $\{F_1,\,F_2\}$.
            $\Mmme(\vx_0) = y_0$.
            Both $F_1$ and $F_2$ are $\beta$-smooth.
            
            \begin{itemize}
                \item (Sufficient Condition) If for any $y_1, y_2 \in [C]$ such that $y_1 \neq y_0$ and $y_2 \neq y_0$,
                \begin{equation}
                    \|\nabla_\vx f_1^{y_0/y_1}(\vx_0) + \nabla_\vx f_2^{y_0/y_2}(\vx_0)\|_2 \le
                    \frac 1 r (f_1^{y_0/y_1}(\vx_0) + f_2^{y_0/y_2}(\vx_0)) - 2\beta r,
                    \tag{\ref{eq:gradient-based-sufficient-cond-our-ensemble}}
                \end{equation}
                then $\Mmme$ is $r$-robust at point $\vx_0$.
                
                \item (Necessary Condition) Suppose for any $\vx \in \{\vx_0 + \vdelta: ||\vdelta||_2 \le r\}$, for any $i \in \{1,2\}$, either $F_i(\vx) = y_0$ or $F_i^{(2)}(\vx) = y_0$.
                If $\Mmme$ is $r$-robust at point $\vx_0$, then
                for any $y_1, y_2 \in [C]$ such that $y_1 \neq y_0$ and $y_2 \neq y_0$,
                \begin{equation}
                    \|\nabla_\vx f_1^{y_0/y_1}(\vx_0) + \nabla_\vx f_2^{y_0/y_2}(\vx_0)\|_2 \le
                    \frac 1 r (f_1^{y_0/y_1}(\vx_0) + f_2^{y_0/y_2}(\vx_0)) + 2\beta r.
                    \tag{\ref{eq:gradient-based-necessary-cond-our-ensemble}}
                \end{equation}
            \end{itemize}
        \end{customthm}
        
        \begin{proof}[Proof of \Cref{thm:gradient-based-sufficient-necessary-cond-our-ensemble}]
            We prove the sufficient condition and necessary condition separately.
            \begin{itemize}
                \item (Sufficient Condition)\\
                From \Cref{adxlem:adx-suf-cond-for-ensemble-rob},
                since there are only two base models, we can simplify the sufficient condition for $\Mmme(\vx)=y_0$ as
                $$
                    \min_{y_i\in [C]: y_i \neq y_0} f_1^{y_0/y_i}(\vx)
                    + 
                    \min_{y_i'\in [C]: y_i'\neq y_0}
                    f_2^{y_0/y_i'}(\vx)
                    \ge 0.
                $$
                In other words, for any $y_1 \neq y_0$ and $y_2 \neq y_0$,
                \begin{equation}
                    f_1^{y_0/y_1}(\vx) + f_2^{y_0/y_2}(\vx) \ge 0.
                    \label{adx-eq:thm4-t1}
                \end{equation}
                With Taylor expansion and model-smoothness assumption, we have
                $$
                \begin{aligned}
                    & \min_{\vx: \|\vx - \vx_0\|_2 \le r}
                    f_1^{y_0/y_1}(\vx) + f_2^{y_0/y_2}(\vx) \\
                    \ge &
                    f_1^{y_0/y_1}(\vx_0) + f_2^{y_0/y_2}(\vx_0)
                    - r \|\nabla_\vx f_1^{y_0/y_1}(\vx_0) +
                    \nabla_\vx f_2^{y_0/y_2}(\vx_0)\|_2
                    - \dfrac{1}{2} \cdot 4\beta r^2.
                \end{aligned}
                $$
                Plugging this into \Cref{adx-eq:thm4-t1} yields the sufficient condition.
                {
                   
                    In the above equation, the term $- \frac{1}{2} \cdot 4\beta r^2$ is bounded from Lagrange remainder. 
                    Here, the $4\beta$ term comes from the fact that $f_1^{y_0/y_1}(\vx) + f_2^{y_0/y_2}(\vx)$ is $(4\beta)$-smooth since it is the sum of difference of $\beta$-smooth function.
                }
                
                \item (Necessary Condition)\\
                From \Cref{adxlem:adx-necs-cond-for-ensemble-rob},
                similarly, the necessary condition for $\Mmme(\vx) = y_0$ is simplified to:
                for any $y_1 \neq y_0$ and $y_2 \neq y_0$,
                \begin{equation}
                    f_1^{y_0/y_1}(\vx) + f_2^{y_0/y_2}(\vx) \ge 0.
                    \tag{\ref{adx-eq:thm4-t1}}
                \end{equation}
                Again, from Taylor expansion, we have
                $$
                \begin{aligned}
                    & \min_{\vx: \|\vx - \vx_0\|_2 \le r}
                    f_1^{y_0/y_1}(\vx) + f_2^{y_0/y_2}(\vx) \\
                    \le &
                    f_1^{y_0/y_1}(\vx_0) + f_2^{y_0/y_2}(\vx_0)
                    - r \|\nabla_\vx f_1^{y_0/y_1}(\vx_0) +
                    \nabla_\vx f_2^{y_0/y_2}(\vx_0)\|_2
                    + \dfrac{1}{2} \cdot 4\beta r^2.
                \end{aligned}
                $$
                Plugging this into \Cref{adx-eq:thm4-t1} yields the necessary condition.
                {
                  
                    In the above equation, the term $+ \frac{1}{2} \cdot 4\beta r^2$ is bounded from Lagrange remainder.
                    The $4\beta$ term appears because of the same reason as before.
                }
            \end{itemize}
        \end{proof}
        
        Since we will compare the robustness of ensemble models and the single model, we show the corresponding conditions for single-model robustness.
        \begin{adxproposition}[Gradient and Confidence Margin Conditions for Single-Model Robustness] \label{prop:gradient-based-sufficient-necessary-cond-individual}
            Given input $\vx_0 \in \sR^d$ with ground-truth label $y_0 \in [C]$.
            Model $F(\vx_0) = y_0$, and it is $\beta$-smooth.
            \begin{itemize}
                \item (Sufficient Condition) If for any $y_1 \in [C]$ such that $y_1 \neq y_0$,
                \begin{equation}
                    \| \nabla_\vx f^{y_0/y_1}(\vx_0) \|_2 \le \frac 1 r f^{y_0/y_1}(\vx_0) - \beta r,
                    \label{eq:gradient-based-sufficient-cond-individual}
                \end{equation}
                 $F$ is $r$-robust at point $\vx_0$.
                
                \item (Necessary Condition) If $F$ is $r$-robust at point $\vx_0$,  for any $y_1 \in [C]$ such that $y_1 \neq y_0$,
                \begin{equation}
                    \| \nabla_\vx f^{y_0/y_1}(\vx_0) \|_2 \le \frac 1 r f^{y_0/y_1}(\vx_0) + \beta r.
                    \label{eq:gradient-based-necessary-cond-individual}
                \end{equation}
            \end{itemize}
        \end{adxproposition}
        
        \begin{proof}[Proof of \Cref{prop:gradient-based-sufficient-necessary-cond-individual}]
            This proposition is apparent given the following inequality from Taylor expansion
            $$
                f^{y_0/y_1}(\vx_0) - r\|\nabla_\vx f^{y_0/y_1}(\vx_0)\|_2 - \beta r^2 \le
                \min_{\vx: \|\vx - \vx_0\|_2 \le r} f^{y_0/y_1}(\vx) 
                \le f^{y_0/y_1}(\vx_0) - r\|\nabla_\vx f^{y_0/y_1}(\vx_0)\|_2 + \beta r^2
            $$
            and the sufficient and necessary robust condition in \Cref{adx-fact:robust-condition-for-single-model}.
        \end{proof}
    
    \subsection{Proof of Robustness Comparison Results between Ensemble Models and Single Models}
        \label{subsec:a-3}
        
        \begin{customcor}{\ref{cor:robust-radius-lower-bound-for-ensemble-constructed-from-single-models}}[Comparison of Ensemble and Single-Model Robustness]
            Given an input $\vx_0 \in \sR^d$ with ground-truth label $y_0 \in [C]$.
            Suppose we have two $\beta$-smooth base models $\{F_1, F_2\}$, which are both $r$-robust at point $\vx_0$.
            For any $\Delta \in [0, 1)$:
            \begin{itemize}[leftmargin=*]
                \item (\weightedEnsemble)
                Define \weightedEnsemble $\Mweight$ with base models $\{F_1, F_2\}$.
                Suppose $\Mweight(\vx_0) = y_0$.
                If for any label $y_i \neq y_0$,
                the base models' smoothness
                $\beta \le \Delta \cdot \min\{f_1^{y_0/y_i}(\vx_0), f_2^{y_0/y_i}(\vx_0)\} / (c^2 r^2)$,
                and the gradient cosine similarity $\cos \langle \nabla_\vx f_1^{y_0/y_i}(\vx_0), \nabla_\vx f_2^{y_0/y_i}(\vx_0) \rangle \le \cos \theta$, 
                then the $\Mweight$ with weights $\{w_1, w_2\}$ is at least $R$-robust at point $\vx_0$ with
                \begin{equation}
                    \begin{small}
                        R = r \cdot \frac{1 - \Delta}{1 + \Delta} \left( 1 - C_{\mathrm{\shortWeightedEnsemble}} (1 - \cos\theta) \right)^{-\nicefrac{1}{2}}, \text{where} 
                    \end{small}
                     \tag{\ref{eq:weighted-ensemble-robust-radius-lower-bound}}
                \end{equation}
                $C_{\mathrm{\shortWeightedEnsemble}} = \underset{y_i: y_i \neq y_0}{\min} \frac{2w_1w_2 f_1^{y_0/y_i}(\vx_0) f_2^{y_0/y_i}(\vx_0)}{(w_1f_1^{y_0/y_i}(\vx_0) + w_2f_2^{y_0/y_i}(\vx_0))^2},
                c = \max\{\frac{1 - \Delta}{1 + \Delta} \left( 1 - C_{\mathrm{\shortWeightedEnsemble}} (1 - \cos\theta) \right)^{-\nicefrac{1}{2}}, 1\}.$
                
                \item (\ourEnsemble)
                Define \ourEnsemble $\Mmme$ with the base models $\{F_1, F_2\}$.
                Suppose $\Mmme(\vx_0) = y_0$.
                If for any label $y_1 \neq y_0$ and $y_2 \neq y_0$,
                the base models' smoothness
                $\beta \le \Delta \cdot \min\{f_1^{y_0/y_1}(\vx_0), f_2^{y_0/y_2}(\vx_0)\} / (c^2 r^2)$,
                and the gradient cosine similarity $\cos \langle \nabla_\vx f_1^{y_0/y_1}(\vx_0), \nabla_\vx f_2^{y_0/y_2}(\vx_0) \rangle \le \cos \theta$, then the $\Mmme$ is at least $R$-robust at point $\vx_0$ with
                \begin{equation}
                \begin{small}
                   R = r \cdot \frac{1 - \Delta}{1 + \Delta} \left( 1 - C_{\mathrm{\shortOurEnsemble}} (1 - \cos\theta) \right)^{-\nicefrac{1}{2}},
                   \text{where}
                    \tag{\ref{eq:max-margin-ensemble-robust-radius-lower-bound}}
                    \end{small}
                \end{equation}
                $C_{\mathrm{\shortOurEnsemble}} = \underset{{\substack{y_1, y_2:\\ y_1, y_2 \neq y_0}}}{\min} \frac{2f_1^{y_0/y_1}(\vx_0) f_2^{y_0/y_2}(\vx_0)}{(f_1^{y_0/y_1}(\vx_0) + f_2^{y_0/y_2}(\vx_0))^2},
                c = \max\{ \frac{1 - \Delta}{1 + \Delta} \left( 1 - C_{\mathrm{\shortOurEnsemble}} (1 - \cos\theta) \right)^{-\nicefrac{1}{2}}, 1\}.$
            \end{itemize}
        \end{customcor}

        \begin{proof}[Proof of \Cref{cor:robust-radius-lower-bound-for-ensemble-constructed-from-single-models}]
            We first prove the theorem for \weightedEnsemble.
            For arbitrary $y_i \neq y_0$, we have
            $$
            \small
            \begin{aligned}
                & \| w_1\nabla_\vx f_1^{y_0/y_i}(\vx_0) + w_2\nabla_\vx f_2^{y_0/y_i}(\vx_0) \|_2 \\
                = & \sqrt{
                    w_1^2 \|\nabla_\vx f_1^{y_0/y_i}(\vx_0) \|_2^2 + 
                    w_2^2 \|\nabla_\vx f_2^{y_0/y_i}(\vx_0) \|_2^2 +
                    2 w_1 w_2 \langle \nabla_\vx f_1^{y_0/y_i}(\vx_0),\,f_2^{y_0/y_i}(\vx_0) \rangle
                } \\
                \le & \sqrt{
                    w_1^2 \|\nabla_\vx f_1^{y_0/y_i}(\vx_0) \|_2^2 + 
                    w_2^2 \|\nabla_\vx f_2^{y_0/y_i}(\vx_0) \|_2^2 +
                    2 w_1 w_2 \|\nabla_\vx f_1^{y_0/y_i}(\vx_0) \|_2 \|\nabla_\vx f_2^{y_0/y_i}(\vx_0) \|_2 \cos\theta
                } \\
                \overset{(i.)}{\le} &
                \sqrt{
                    w_1^2 \left(\frac 1 r f_1^{y_0/y_i}(\vx_0) + \beta r\right)^2 +
                    w_2^2 \left(\frac 1 r f_2^{y_0/y_i}(\vx_0) + \beta r\right)^2 +
                    2 w_1 w_2 \left(\frac 1 r f_1^{y_0/y_i}(\vx_0) + \beta r\right) \left(\frac 1 r f_2^{y_0/y_i}(\vx_0) + \beta r\right) \cos\theta
                } \\
                = & \frac 1 r
                \sqrt{
                    w_1^2 \left(f_1^{y_0/y_i}(\vx_0) + \beta r^2\right)^2 +
                    w_2^2 \left(f_2^{y_0/y_i}(\vx_0) + \beta r^2\right)^2 +
                    2 w_1 w_2 \left(f_1^{y_0/y_i}(\vx_0) + \beta r^2\right) \left(f_2^{y_0/y_i}(\vx_0) + \beta r^2\right) \cos\theta
                } \\
                \overset{(ii.)}{\le} &
                \frac 1 r \cdot \left( 1 + \frac{\Delta}{c^2} \right)
                \sqrt{
                    w_1^2 f_1^{y_0/y_i}(\vx_0)^2 +
                    w_2^2 f_2^{y_0/y_i}(\vx_0)^2 +
                    2 w_1 w_2 f_1^{y_0/y_i}(\vx_0) f_2^{y_0/y_i}(\vx_0) \cos\theta
                } \\
                = & 
                \frac 1 r \cdot \left( 1 + \frac{\Delta}{c^2} \right)
                \sqrt{
                    \left( w_1 f_1^{y_0/y_i}(\vx_0) +
                    w_2 f_2^{y_0/y_i}(\vx_0) \right)^2 
                    - 2(1 - \cos\theta) w_1 f_1^{y_0/y_i}(\vx_0) w_2 f_2^{y_0/y_i}(\vx_0)
                } \\
                \overset{(iii.)}{\le} & \frac 1 r \cdot \left( 1 + \frac{\Delta}{c^2} \right)
                \sqrt{
                    1 - (1-\cos\theta)C_{\mathrm{\shortWeightedEnsemble}}
                }
                \left( w_1 f_1^{y_0/y_i}(\vx_0) +
                    w_2 f_2^{y_0/y_i}(\vx_0) \right)
            \end{aligned}
            $$
            where $(i.)$ follows from the necessary condition in \Cref{prop:gradient-based-sufficient-necessary-cond-individual};
            $(ii.)$ uses the condition on $\beta$;
            and $(iii.)$ replaces $2w_1w_2 f_1^{y_0/y_i}(\vx_0)f_2^{y_0/y_i}(\vx_0)$ leveraging $C_{\mathrm{\shortWeightedEnsemble}}$.
            Now, we define 
            $$
                K := \frac{1 - \Delta}{1 + \Delta} \left( 1 - C_{\mathrm{\shortWeightedEnsemble}} (1 - \cos\theta) \right)^{-\nicefrac{1}{2}}.
            $$
            All we need to do is to prove that $\Mweight$ is robust within radius $Kr$.
            To do so, from \Cref{eq:gradient-based-sufficient-cond-weight-ensemble}, we upper bound $\| w_1\nabla_\vx f_1^{y_0/y_i}(\vx_0) + w_2\nabla_\vx f_2^{y_0/y_i}(\vx_0) \|_2 $ by $\frac{1}{Kr} \left( w_1f_1^{y_0/y_i}(\vx_0) + w_2f_2^{y_0/y_i}(\vx_0) \right) - \beta K r (w_1 + w_2)$:
            $$
            \begin{aligned}
                & \| w_1\nabla_\vx f_1^{y_0/y_i}(\vx_0) + w_2\nabla_\vx f_2^{y_0/y_i}(\vx_0) \|_2 \\
                \le & \frac 1 r \cdot \left( 1 + \frac{\Delta}{c^2} \right)
                \sqrt{
                    1 - (1-\cos\theta)C_{\mathrm{\shortWeightedEnsemble}}
                }
                \left( w_1 f_1^{y_0/y_i}(\vx_0) +
                    w_2 f_2^{y_0/y_i}(\vx_0) \right) \\
                \le & \frac 1 r (1 + \Delta)
                \sqrt{
                    1 - (1-\cos\theta)C_{\mathrm{\shortWeightedEnsemble}}
                }
                \left( w_1 f_1^{y_0/y_i}(\vx_0) +
                    w_2 f_2^{y_0/y_i}(\vx_0) \right) \\
                = & \frac 1 r \cdot
                \dfrac{1 - \Delta}{\frac{1 - \Delta}{1 + \Delta} \left(1 - (1-\cos\theta)C_{\mathrm{\shortWeightedEnsemble}}\right)^{-1/2}} 
                \left( w_1 f_1^{y_0/y_i}(\vx_0) +
                    w_2 f_2^{y_0/y_i}(\vx_0) \right) \\
                = & \frac{1}{Kr} (1 - \Delta) \left( w_1 f_1^{y_0/y_i}(\vx_0) +
                    w_2 f_2^{y_0/y_i}(\vx_0) \right) \\
                \le & \frac{1}{Kr} \left(
                    w_1 f_1^{y_0/y_i}(\vx_0) +
                    w_2 f_2^{y_0/y_i}(\vx_0)
                    - \Delta \min\{ f_1^{y_0/y_i}(\vx_0), f_2^{y_0/y_i}(\vx_0) \} (w_1 + w_2)
                \right).
            \end{aligned}
            $$
            Notice that $\Delta \min\{ f_1^{y_0/y_i}(\vx_0), f_2^{y_0/y_i}(\vx_0) \} \ge \beta c^2r^2 $ from $\beta$'s condition,
            so
            $$
            \begin{aligned}
                & \| w_1\nabla_\vx f_1^{y_0/y_i}(\vx_0) + w_2\nabla_\vx f_2^{y_0/y_i}(\vx_0) \|_2 \\
                \le & \frac{1}{Kr} \left(
                    w_1 f_1^{y_0/y_i}(\vx_0) +
                    w_2 f_2^{y_0/y_i}(\vx_0)
                    - \beta c^2r^2 (w_1 + w_2) \right) \\
                = & \frac{1}{Kr} \left(
                    w_1 f_1^{y_0/y_i}(\vx_0) +
                    w_2 f_2^{y_0/y_i}(\vx_0) \right)
                    - \beta Kr (w_1 + w_2) \cdot \dfrac{c^2}{K^2} \\
                \le & \frac{1}{Kr} \left(
                    w_1 f_1^{y_0/y_i}(\vx_0) +
                    w_2 f_2^{y_0/y_i}(\vx_0) \right)
                    - \beta Kr (w_1 + w_2).
            \end{aligned}
            $$
            From \Cref{eq:gradient-based-sufficient-cond-weight-ensemble}, the theorem for \weightedEnsemble is proved.
            \vspace{1em}
            
            Now we prove the theorem for \ourEnsemble.
            Similarly, for any arbitrary $y_1, y_2$ such that $y_1 \neq y_0, y_2 \neq y_0$,
            we have
            $$
            \begin{aligned}
                & \| \nabla_\vx f_1^{y_0/y_1}(\vx_0) + \nabla_\vx f_2^{y_0/y_2}(\vx_0) \|_2 \\
                \le & \frac 1 r \cdot \left( 1 + \frac{\Delta}{c^2} \right)
                \sqrt{
                    1 - (1-\cos\theta)C_{\mathrm{\shortOurEnsemble}}
                }
                \left( f_1^{y_0/y_1}(\vx_0) +
                    f_2^{y_0/y_2}(\vx_0) \right).
            \end{aligned}
            $$
            Now we define
            $$
                K' := \frac{1 - \Delta}{1 + \Delta} \left( 1 - C_{\mathrm{\shortOurEnsemble}} (1 - \cos\theta) \right)^{-\nicefrac{1}{2}}.
            $$
            Again, from $\beta$'s condition we have $\Delta \min\{ f_1^{y_0/y_1}(\vx_0), f_2^{y_0/y_2}(\vx_0) \} \ge \beta c^2r^2 $ and
            $$
            \begin{aligned}
                & \| \nabla_\vx f_1^{y_0/y_1}(\vx_0) + \nabla_\vx f_2^{y_0/y_2}(\vx_0) \|_2 
                \le  \frac{1}{K'r} \left(
                    f_1^{y_0/y_i}(\vx_0) +
                    f_2^{y_0/y_i}(\vx_0) \right)
                    - 2\beta K'r.
            \end{aligned}
            $$
            From \Cref{eq:gradient-based-sufficient-cond-our-ensemble}, the ensemble is $(K'r)$-robust at point $\vx_0$, i.e., the theorem for \ourEnsemble is proved.
        \end{proof}
    
    \subsection{Proofs of Model-Smoothness Bounds for Randomized Smoothing}
        \label{subsec:a-4}
        
        
        \begin{customthm}{\ref{thm:smoothness-bound}}[Model-Smoothness Upper Bound for $\bar g_f^\varepsilon$]
            Let $\varepsilon \sim \gN(0,\sigma^2 \mI_d)$ be a Gaussian random variable, 
            then the soft smoothed confidence function $\bar g^\varepsilon_{f}$ is $(2/\sigma^2)$-smooth.
        \end{customthm}
        
        \begin{proof}[Proof of \Cref{thm:smoothness-bound}]
            Recall that $\bar g^\varepsilon_{f}(\vx)_j = \E_{\varepsilon\sim\gN(0,\sigma^2\mI_d)} f(\vx + \varepsilon)_j$, where $f(\vx + \varepsilon)_j$ is a function from $\sR^d$ to $\{0,\,1\}$.
            Therefore, to prove that $g^\varepsilon_{\gM}$ is $(2/\sigma^2)$-smooth, we only need to show that for any function $f: \sR^d \to [0,\,1]$, the function $\bar f := f * \gN(0, \sigma^2\mI_d)$ is $(2/\sigma^2)$-smooth.
            
            According to \citep[Lemma 1]{salman2019provably}, we have
            \begin{align}
                \bar f(\vx) & =
                \dfrac{1}{(2\pi\sigma^2)^{d/2}}
                \int_{\sR^d}
                f(\vt) \exp\left( - \dfrac{\|\vx - \vt\|_2^2}{2\sigma^2} \right) \dif \vt, \\
                \nabla \bar f(\vx) & =
                \dfrac{1}{(2\pi\sigma^2)^{d/2}\sigma^2}
                \int_{\sR^d}
                f(\vt) (\vx-\vt) \exp\left( - \dfrac{\|\vx - \vt\|_2^2}{2\sigma^2} \right) \dif \vt.
            \end{align}
            To show $\bar f$ is $(2/\sigma^2)$-smooth, we only need to show that $\nabla \bar f$ is $(2/\sigma^2)$-Lipschitz.
            Let $\mH_{\bar f}(\vx)$ be the Hessian matrix of $\bar f$.
            Thus, we only need to show that for any unit vector $\vu$, $|\vu^\T \mH_{\bar f}(\vx) \vu| \le 2/\sigma^2$.
            By the isotropy of $\mH_{\bar f}(\vx)$, it is sufficient to consider $\vu = (1,0,0,\dots,0)^\T$, where $\vu^\T \mH_{\bar f}(\vx) \vu = \mH_{\bar f}(\vx)_{11}$.
            Now we only need to bound the absolute value of $\mH_{\bar f}(\vx)_{11}$:
            \begin{align}
                |\mH_{\bar f}(\vx)_{11}| \nonumber
                & = \Big| \dfrac{1}{(2\pi\sigma^2)^{d/2}\sigma^2} \int_{\sR^d} f(\vt) \cdot
                \dfrac{\partial}{\partial \vx_1} 
                (\vx-\vt) \exp\left( - \dfrac{\|\vx - \vt\|_2^2}{2\sigma^2} \right) \dif \vt \Big| \nonumber \\
                & = \dfrac{1}{(2\pi\sigma^2)^{d/2}\sigma^2} \Big| \int_{\sR^d} f(\vt) \cdot
                \left(1 - \dfrac{(\vx_1 - \vt_1)^2}{\sigma^2}\right) \exp\left( - \dfrac{\|\vx - \vt\|_2^2}{2\sigma^2} \right) \dif \vt \Big| \nonumber \\
                & \le \dfrac{1}{(2\pi\sigma^2)^{d/2}\sigma^2} \Big| \int_{\sR^d} \exp\left( - \dfrac{\|\vx - \vt\|_2^2}{2\sigma^2} \right) \dif \vt \Big| \nonumber \\
                & \hspace{2em}
                + \dfrac{1}{(2\pi\sigma^2)^{d/2}\sigma^2} \Big| 
                \int_{\sR^d}
                \dfrac{(\vx_1 - \vt_1)^2}{\sigma^2}
                \exp\left( - \dfrac{\|\vx - \vt\|_2^2}{2\sigma^2} \right) \dif \vt \Big| \nonumber \\
                & = \dfrac{1}{\sigma^2} + \dfrac{1}{\sqrt{2\pi\sigma^2}\sigma^2} \cdot 2 \int_{0}^\infty \dfrac{x^2}{\sigma^2} \exp\left( - \dfrac{x^2}{2\sigma^2} \right) \dif x \nonumber \\
                & = \dfrac{1}{\sigma^2} + \sqrt{\dfrac{2}{\pi}} \cdot \dfrac{1}{\sigma^2} \int_0^\infty t^2 \exp(-\nicefrac{t^2}{2}) \dif t. \label{eq:thm-2-pf-1}
            \end{align}
            Let $\Gamma(\cdot)$ be the Gamma function, we note that
            $$
                \int_0^\infty t^2 \exp(-\nicefrac{t^2}{2}) \dif t = \int_0^\infty t \exp(-\nicefrac{t^2}{2}) \dif (-\nicefrac{t^2}{2}) = \sqrt{2} \int_0^\infty \sqrt{t} \exp(-t) \dif t = \sqrt{2}\Gamma(\nicefrac{3}{2}) = \sqrt{\nicefrac{\pi}{2}},
            $$
            and thus
            \begin{equation}
                |\mH_{\bar f}(\vx)_{11}| \le \dfrac{1}{\sigma^2} + \sqrt{\dfrac{2}{\pi}} \cdot \dfrac{1}{\sigma^2} \cdot \sqrt{\dfrac{\pi}{2}} = \dfrac{2}{\sigma^2},
            \end{equation}
            which concludes the proof.
        \end{proof}
        
        \begin{remark}
            The model-smoothness upper bound \Cref{thm:smoothness-bound} is not limited to the ensemble model with \beforeProtocol strategy.
            Indeed, for arbitrary classification models, since the confidence score is in range $[0,\,1]$, the theorem still holds.
            If the confidence score is bounded in $[a,\,b]$, simple scaling yields the model-smoothness upper bound $\beta = \frac{2(b-a)}{\sigma^2}$.
        \end{remark}

        \begin{adxproposition}[Model-Smoothness Lower Bound for $\bar g_f^\varepsilon$]
            There exists a smoothed confidence function $\bar g^\varepsilon_f$ that is $\beta$-smooth if and only if $\beta \ge \left(\dfrac{1}{\sqrt{2\pi e}\sigma^2}\right)$.
            \label{prop:smoothness-lower-bound}
        \end{adxproposition}

        \begin{proof}[Proof of \Cref{prop:smoothness-lower-bound}]
            We prove by construction.
            Consider the single dimensional input space $\sR$, and
            a model $f$ that has confidence $1$ if and only if input $x \ge 0$.
            As a result, 
            $$
                g^\varepsilon_{f} (x)_{y_0} = \dfrac{1}{\sqrt{2\pi}\sigma} \int_{0}^{+\infty} \exp\left(-\dfrac{(t-x)^2}{2\sigma^2}\right) \dif t
                = \dfrac{1}{\sqrt{2\pi}\sigma} \int_{-x}^{+\infty} \exp\left(-\dfrac{t^2}{2\sigma^2}\right) \dif t.
            $$
            Thus,
            $$
                \dfrac{\dif g^\varepsilon_{f} (x)_{y_0}}{\dif x} = \dfrac{1}{\sqrt{2\pi}\sigma} \exp\left(-\dfrac{x^2}{2\sigma^2}\right)
                \quad \text{and} \quad
                \Big|\dfrac{\dif g^\varepsilon_{f} (x)_{y_0}^2}{\dif^2 x} \Big| = \dfrac{1}{\sqrt{2\pi}\sigma^2}
                \cdot \Big| \dfrac{x}{\sigma} \Big| \exp\left( -\dfrac{x^2}{2\sigma^2} \right).
            $$
            By symmetry, we study the function $h(x) = x\exp(-x^2/2)$ for $x \ge 0$.
            We have $h'(x) = (1 - x) \exp(- x^2/2)$.
            Thus, $h(x)$ obtains its maximum at $x_0 = 1$: $h(x_0) = \exp(-\nicefrac{1}{2})$, which implies that
            $$
                \max \Big|\dfrac{\dif g^\varepsilon_{f} (x)_{y_0}^2}{\dif^2 x} \Big| = \dfrac{\exp(-\nicefrac{1}{2})}{\sqrt{2\pi}{\sigma^2}} = \dfrac{1}{\sqrt{2\pi e}\sigma^2}
            $$
            which implies $\beta \ge \dfrac{1}{\sqrt{2\pi e}\sigma^2}$ for this $\bar g_f^\varepsilon$ per smoothness definition~(\Cref{def:beta-smooth}).
        \end{proof}
    
    \subsection{Proofs of Robustness Conditions for Smoothed Ensemble Models}
        \label{subsec:a-5}

        \begin{customcor}{\ref{cor:gradient-based-sufficient-necessary-cond-smoothed-weight-ensemble}}[Gradient and Confidence Margin Conditions for Smoothed \shortWeightedEnsemble Robustness]
            Given input $\vx_0 \in \sR^d$ with ground-truth label $y_0 \in [C]$.
            Let $\varepsilon\sim\gN(0, \sigma^2\mI_d)$ be a Gaussian random variable.
            Define soft smoothed confidence $\bar g^\varepsilon_i(\vx) := \E_\varepsilon f_i(\vx + \varepsilon)$ for each base model $F_i$~($1\le i\le N$).
            The $\bar G^\varepsilon_\Mweight$ is a \shortWeightedEnsemble defined over soft smoothed base models $\{\bar g^\varepsilon_i\}_{i=1}^N$ with weights $\{w_i\}_{i=1}^N$.
            $\bar G^\varepsilon_\Mweight(\vx_0) = y_0$.
            \begin{itemize}
                \vspace{-0.5em}
                \item (Sufficient Condition) The $\bar G^\varepsilon_{\Mweight}$ is $r$-robust at point $\vx_0$ if for any $y_i \neq y_0$,
                \begin{equation}
                    \hspace{-1em}
                    \Big\|\sum_{j=1}^N w_j \nabla_\vx (\bar g_j^{\varepsilon})^{y_0/y_i}(\vx_0) \Big\|_2 \le \frac 1 r \sum_{j=1}^N w_j (\bar g_j^\varepsilon)^{y_0/y_i}(\vx_0) - \frac{2r}{\sigma^2} \sum_{j=1}^N w_j,
                    \tag{\ref{eq:gradient-based-sufficient-cond-smoothed-weight-ensemble}}
                \end{equation}
                \item (Necessary Condition) If $\bar G^\varepsilon_{\Mweight}$ is $r$-robust at point $\vx_0$,  for any $y_i \neq y_0$,
                \begin{equation}
                    \hspace{-1em}
                    \Big\|\sum_{j=1}^N w_j \nabla_\vx (\bar g^\varepsilon_j)^{y_0/y_i}(\vx_0) \Big\|_2 \le \frac 1 r \sum_{j=1}^N w_j (\bar g^\varepsilon_j)^{y_0/y_i}(\vx_0) + \dfrac{2r}{\sigma^2} \sum_{j=1}^N w_j.
                    \tag{\ref{eq:gradient-based-necessary-cond-smoothed-weight-ensemble}}
                \end{equation}
            \end{itemize}
        \end{customcor}
        
        \begin{proof}[Proof of \Cref{cor:gradient-based-sufficient-necessary-cond-smoothed-weight-ensemble}]
            Since $\bar G^\varepsilon_{\Mweight}$ is a \shortWeightedEnsemble defined over $\{\bar g_i^\varepsilon\}_{i=1}^N$, we apply \Cref{thm:gradient-based-sufficient-necessary-cond-weight-ensemble} directly for $\bar G^\varepsilon_{\Mweight}$.
            Notice that each $\bar g_i^\varepsilon$ has model-smoothness bound $\beta = 2/\sigma^2$ from \Cref{thm:smoothness-bound} and the corollary statement follows.
        \end{proof}
        
        \begin{customcor}{\ref{cor:gradient-based-sufficient-necessary-cond-smoothed-our-ensemble}}[Gradient and Confidence Margin Condition for Smoothed \shortOurEnsemble Robustness]
            Given input $\vx_0 \in \sR^d$ with ground-truth label $y_0 \in [C]$.
            Let $\varepsilon\sim\gN(0, \sigma^2\mI_d)$ be a Gaussian random variable.
            Define soft smoothed confidence $\bar g^\varepsilon_i(\vx) := \E_\varepsilon f_i(\vx + \varepsilon)$ for either base model $F_1$ or $F_2$.
            The $\bar G^\varepsilon_\Mmme$ is a \shortOurEnsemble defined over soft smoothed base models $\{\bar g^\varepsilon_1,\bar g^\varepsilon_2\}$.
            $\bar G^\varepsilon_\Mmme(\vx_0) = y_0$.
            
            \begin{itemize}
                \item (Sufficient Condition) If for any $y_1, y_2 \in [C]$ such that $y_1 \neq y_0$ and $y_2 \neq y_0$,
                \begin{equation}
                    \|\nabla_\vx (\bar g_1^\varepsilon)^{y_0/y_1}(\vx_0) + \nabla_\vx (\bar g_2^\varepsilon)^{y_0/y_2}(\vx_0)\|_2 \le
                    \frac 1 r ( (\bar g_1^\varepsilon)^{y_0/y_1}(\vx_0) + (\bar g_2^\varepsilon)^{y_0/y_2}(\vx_0)) - \dfrac{4r}{\sigma^2},
                    \tag{\ref{eq:gradient-based-sufficient-cond-smoothed-our-ensemble}}
                \end{equation}
                then $\bar G^\varepsilon_\Mmme$ is $r$-robust at point $\vx_0$.
                
                \item (Necessary Condition) Suppose for any $\vx \in \{\vx_0 + \vdelta: ||\vdelta||_2 \le r\}$, for any $i \in \{1,2\}$, either $G_{F_i}(\vx) = y_0$ or $G_{F_i}^{(2)}(\vx) = y_0$.
                If $\bar G^\varepsilon_\Mmme$ is $r$-robust at point $\vx_0$, then
                for any $y_1, y_2 \in [C]$ such that $y_1 \neq y_0$ and $y_2 \neq y_0$,
                \begin{equation}
                    \|\nabla_\vx (\bar g_1^\varepsilon)^{y_0/y_1}(\vx_0) + \nabla_\vx (\bar g_2^\varepsilon)^{y_0/y_2}(\vx_0)\|_2 \le
                    \frac 1 r ((\bar g_1^\varepsilon)^{y_0/y_1}(\vx_0) + (\bar g_2^\varepsilon)^{y_0/y_2}(\vx_0)) + \dfrac{4r}{\sigma^2}.
                    \tag{\ref{eq:gradient-based-necessary-cond-smoothed-our-ensemble}}
                \end{equation}
            \end{itemize}
        \end{customcor} 
        
        \begin{proof}[Proof of \Cref{cor:gradient-based-sufficient-necessary-cond-smoothed-our-ensemble}]
            Since $\bar G^\varepsilon_{\Mmme}$ is constructed over confidences $\bar g^\varepsilon_1$ and $\bar g^\varepsilon_2$, we can directly apply \Cref{thm:gradient-based-sufficient-necessary-cond-weight-ensemble}.
            Again, with the model-smoothness bound $\beta = 2/\sigma^2$ we can easily derive the corollary statement.
        \end{proof}

\section{Analysis of Ensemble Smoothing Strategies}

    \label{adx-sec:2}

    In main text we mainly use the adapted randomized model smoothing strategy which is named \beforeProtocol~(\shortBeforeProtocol).
    We also consider \afterProtocol~(\afterProtocol).
    Through the following analysis, we will show \beforeProtocol generally provides higher certified robust radius than \afterProtocol which justifies our choice of the strategy.
    
    The \beforeProtocol strategy is defined in \Cref{def:ebs}.
    The \afterProtocol strategy is defined as such.
    %
    \begin{definition}[\newterm{\afterProtocol~(\shortAfterProtocol)}]
        \label{def:eas}
        Let $\gM$ be an ensemble model over base models $\{F_i\}_{i=1}^N$.
        Let $\rvepsilon$ be a random variable.
        The \shortAfterProtocol ensemble $H^\rvepsilon_{\gM}: \sR^d \mapsto [C]$ at input $\vx_0 \in \sR^d$ is defined as:
        \begin{equation}
            H^\rvepsilon_{\gM}(\vx_0) := G^\rvepsilon_{F_c}(\vx_0)
            \quad
            \mathrm{where}
            \quad
             c = \argmax_{i \in [N]} g^\rvepsilon_{F_i}(\vx_0)_{G^\rvepsilon_{F_i}(\vx_0)}.
            \label{adx-eq:def-b-1}
        \end{equation}
        Here, $c$ is the index of the smoothed base model selected.
    \end{definition}
    
    \begin{remark}
        In \shortBeforeProtocol, we first construct a model ensemble $\gM$ based on base models using \shortWeightedEnsemble or \shortOurEnsemble protocol, then apply randomized smoothing on top of the ensemble.
        The resulting smoothed ensemble predicts the most frequent class of $\gM$ when the input follows distribution $\vx_0 + \rvepsilon$.
        
        In \shortAfterProtocol, we use $\rvepsilon$ to construct smoothed classifiers for base models respectively.
        Then, for given input $\vx_0$, the ensemble agrees on the base model which has the highest probability for its predicted class. 
    \end{remark}
    
    \subsection{Certified Robustness}
    
        In this subsection, we characterize the certified robustness when using both strategies.
    
        \subsubsection{\beforeProtocol}
        
            The following theorem gives an explicit method~(first compute $g^\rvepsilon_{\gM}(\vx_0)_{G^\rvepsilon_{\gM}(\vx_0)}$ via sampling then compute $r$) to compute the certified robust radius $r$ for \shortBeforeProtocol protocol.
            This method is used for computing the certified robust radius in our paper.
            All other baselines appeared in our paper also use this method.
        
            \begin{adxproposition}[Certified Robustness for \beforeProtocol]
                \label{adx-prop:rob-radii-before}
                Let $G^\rvepsilon_{\gM}$ be an ensemble constructed by \shortBeforeProtocol strategy.
                The random variable $\rvepsilon \sim \gN(0, \sigma^2\mI_d)$.
                Then the ensemble $G^\rvepsilon_{\gM}$ is $r$-robust at point $\vx_0$ where
                \begin{equation}
                    \label{adx-eq:before-protocol-r}
                    r := \sigma
                    \Phi^{-1}\left(g^\rvepsilon_{\gM}(\vx_0)_{G^\rvepsilon_{\gM}(\vx_0)}\right).
                \end{equation}
                Here, $g^\rvepsilon_{\gM}(\vx_0)_j = \Pr_{\epsilon} (\gM(\vx_0+\rvepsilon)=j)$.
            \end{adxproposition}
        
            The proposition is a direct application of \Cref{lem:rand-smooth-vanilla-certify}.
        
        \subsubsection{\afterProtocol}
        
            The following theorem gives an explicit method to compute the certified robust radius $r$ for \shortAfterProtocol protocol.
        
            \begin{theorem}[Certified robustness for \afterProtocol]
                \label{thm:rob-radii-after}
                Let $H^\rvepsilon_{\gM}$ be an ensemble constructed by \shortAfterProtocol strategy over base models $\{F_i\}_{i=1}^N$.
                The random variable $\epsilon\sim\gN(0,\sigma^2\mI_d)$.
                Let $y_0 = H^\rvepsilon_{\gM}(\vx_0)$. For each $i \in [N]$, define
                $$
                    r_i := \left\{
                    \begin{aligned}
                        \sigma \Phi^{-1}\left( g^\rvepsilon_{F_i}(\vx_0)_{G^\rvepsilon_{F_i}(\vx_0)} \right), & \quad \text{if } G^\rvepsilon_{F_i}(\vx_0) = y_0 \\
                        -\sigma \Phi^{-1}\left( g^\rvepsilon_{F_i}(\vx_0)_{G^\rvepsilon_{F_i}(\vx_0)} \right). & \quad \text{if } G^\rvepsilon_{F_i}(\vx_0) \neq y_0
                    \end{aligned}
                    \right.
                $$
                Then the ensemble $H^\rvepsilon_{\gM}$ is $r$-robust at point $\vx_0$ where
                \begin{equation}
                    \label{adx-eq:after-protocol-r}
                    r := \dfrac{\max_{i\in [N]} r_i + \min_{i\in [N]} r_i}{2}.
                \end{equation}
            \end{theorem}
            
            \begin{remark}
                The theorem appears to be a bit counter-intuitive --- picking the best smoothed model in terms of certified robustness cannot give strong certified robustness for the ensemble.
                As long as the base models have different certified robust radius~(i.e., $r_i$'s are different), the $r$, certified robust radius for the ensemble, is strictly inferior to that of the best base model~(i.e., $\max r_i$).
                Furthermore, if there exists a base model with wrong prediction~(i.e., $r_i \le 0$), the certified robust radius $r$ is strictly smaller than \emph{half} of the best base model.
            \end{remark}
            
            \begin{proof}[Proof of \Cref{thm:rob-radii-after}]
                Without loss of generality, we assume $r_1 > r_2 > \dots > r_N$.
                Let the perturbation added to $\vx_0$ has $L_2$ length $\delta$.
                
                When $\delta \le r_N$, since picking any model always gives the right prediction, the ensemble is robust.
                
                When $r_N < \delta \le \frac{r_1 + r_N}{2}$, the highest robust radius with wrong prediction is $\delta - r_N$, and we can still guarantee that model $F_1$ has robust radius at least $r_1 - \delta$ from the smoothness of function $\vx \mapsto g^\rvepsilon_{F_1}(\vx)_{G^\rvepsilon_{F_1}(\vx_0)}$~\citep{salman2019provably}.
                Since $r_1 - \delta \ge \frac{r_1 - r_N}{2} \ge \delta - r_N$, the ensemble will agree on $F_1$ or other base model with correct prediction and still gives the right prediction.
                
                When $\delta > \frac{r_1 + r_N}{2}$, suppose $f_N$ is a linear model and only predicts two labels~(which achieves the tight robust radius bound according to \cite{cohen2019certified}), then $f_N$ can have robust radius $\delta - r_N$ for the wrong prediction.
                At the same time, for any other model $F_i$ which is linear and predicts correctly, the robust radius is at most $r_i - \delta$.
                Since $r_i - \delta < r_1 - \delta < \frac{r_1 - r_N}{2} < \delta - r_N$, the ensemble can probably give wrong prediction.
                
                In summary, as we have shown, the certified robust radius can be at most $r$.
                For any radius $\delta > r$, there exist base models which lead the ensemble $H^\rvepsilon_\gM(\vx_0 + \delta\ve)$ to predict the label other than $y_0$.
            \end{proof}
    
    \subsection{Comparison of Two Strategies}
    
        In this subsection, we compare the two ensemble strategies when the ensembles are constructed from two base models.
        \begin{corollary}[Smoothing Strategy Comparison]
            \label{cor:smoothing-strategy-comparison}
            Given $\Mmme$, a \ourEnsemble constructed from base models $\{f_a, f_b\}$.
            Let $\rvepsilon\sim\gN(0,\sigma^2\mI_d)$.
            Let $G^\rvepsilon_{\Mmme}$ be the \shortBeforeProtocol ensemble, and $H^\rvepsilon_{\Mmme}$ be the \shortAfterProtocol ensemble.
            Suppose at point $\vx_0$ with ground-truth label $y_0$, 
            $G^\rvepsilon_{F_a}(\vx_0) = G^\rvepsilon_{F_b}(\vx_0) = y_0$, $g^\rvepsilon_{F_a}(\vx_0) > 0.5$, $g^\rvepsilon_{F_b}(\vx_0) > 0.5$.
            
            Let $\delta$ be their probability difference for class $y_0$, i.e, $\delta := |g^\rvepsilon_{F_a}(\vx_0)_{y_0} - g^\rvepsilon_{F_b}(\vx_0)_{y_0}|$,.
            Let $p_{\min}$ be the smaller probability for class $y_0$ between them, i.e., $p_{\min} := \min\{g^\rvepsilon_{F_a}(\vx_0)_{y_0},\, g^\rvepsilon_{F_b}(\vx_0)_{y_0}\}$.
            We denote $p$ to the probability of choosing the correct class when the base models disagree with each other;
            denote $p_{ab}$ to the probability of both base models agreeing on the correct class:
            $$
                \begin{aligned}
                    & p := \Pr_\rvepsilon \left( \Mmme(\vx_0 + \rvepsilon) = y_0\, |\, F_a(\vx_0 + \rvepsilon) \neq F_b(\vx_0 + \rvepsilon) \,\mathrm{and}\, (F_a(\vx_0 + \rvepsilon)=y_0 \,\mathrm{or}\, F_b(\vx_0 + \rvepsilon)=y_0) \right), \\
                    & p_{ab} := \Pr_\rvepsilon \left( F_a(\vx_0 + \rvepsilon) = F_b(\vx_0 + \rvepsilon) = y_0 \right).
                \end{aligned}
            $$
            We have:
            \begin{enumerate}[leftmargin=*]
                \item 
                If
                $
                    p >  1/2 + (2+4(p_{\min}-p_{ab})/\delta)^{-1},
                $
                $r_G > r_H$.
                
                \item
                If 
                $
                    \displaystyle
                    p \le 1/2,
                $
                $r_H \ge r_G$.
            \end{enumerate}
            Here, $r_G$ is the certified robust radius of $G^\rvepsilon_{\gM_{\mathrm{\shortOurEnsemble}}}$ computed from \Cref{adx-eq:before-protocol-r}; and $r_H$ is the certified robust radius of $H^\rvepsilon_{\gM_{\mathrm{\shortOurEnsemble}}}$ computed from \Cref{adx-eq:after-protocol-r}.
        \end{corollary}
        
        \begin{remark}
            Since $p$ is the probability where the ensemble chooses the correct prediction between two base model predictions,
            with \ourEnsemble, we think $p > 1/2$ with non-trivial margin.
            
            The quantity $p_{\min} - p_{ab}$ and $\delta$ both measure the base model's diversity in terms of predicted label distribution, and generally they should be close.
            As a result, $1/2 + (2+4(p_{\min}-p_{ab})/\delta)^{-1} \approx 1/2 + 1/6 = 2/3$, and
            case~(1) should be much more likely to happen than case~(2). 
            Therefore, \emph{\shortBeforeProtocol usually yields higher robustness guarantee}.
            We remark that the similar tendency also holds with multiple base models.
        \end{remark}
        
        \begin{proof}[Proof of \Cref{cor:smoothing-strategy-comparison}]
            For convenience, define $p_a := g^\rvepsilon_{F_a}(\vx_0)_{y_0}, p_b := g^\rvepsilon_{F_b}(\vx_0)_{y_0}$, where $p_a = p_b + \delta$ and $p_{\min} = p_b$.
            
            From \Cref{adx-prop:rob-radii-before} and \Cref{thm:rob-radii-after}, we have
            $$
                r_G := \dfrac{\sigma}{2} \cdot 2\Phi^{-1}\left( \Pr_\epsilon(\gM_{\mathrm{\shortOurEnsemble}}(\vx_0 + \epsilon) = y_0) \right),
                \quad
                r_H := \dfrac{\sigma}{2}\left( \Phi^{-1}(p_a) + \Phi^{-1}(p_b) \right).
            $$
            Notice that
            $
                \Pr_\epsilon (\gM_{\mathrm{\shortOurEnsemble}}(\vx_0 + \epsilon) = y_0) = p_{ab} + p(p_a + p_b - 2p_{ab}),
            $
            we can rewrite $r_G$ as 
            $$
                r_G = \dfrac{\sigma}{2} \cdot 2\Phi^{-1}(p_{ab} + p(p_a + p_b - 2p_{ab})).
            $$
            \begin{enumerate}[leftmargin=*]
                \item
                When $p > 1/2 + (2 + 4(p_{\min}-p_{ab})/\delta)^{-1}$, \\
                since
                $$
                    p > 
                    \frac 1 2 + \dfrac{1}{2 + \frac{4(p_{\min} - p_{ab})}{\delta}}
                    =
                    \frac 1 2 + \frac{\delta}{2\delta + 4(p_b - p_{ab})}
                    =
                    \frac{p_a + p_b + \delta - 2p_{ab}}{2(p_a + p_b - 2p_{ab})}
                    =
                    \frac{p_a - p_{ab}}{p_a + p_b - 2p_{ab}},
                $$
                we have
                $
                    p_{ab} + p(p_a + p_b - 2p_{ab}) > p_a.
                $
                Therefore, $r_G > \sigma \Phi^{-1}(p_a)$.
                Whereas, $r_H \le \sigma/2 \cdot 2\Phi^{-1}(p_a) = \sigma \Phi^{-1}(p_a)$. So $r_G > r_H$.
                
                \item
                When $p \le 1/2$,
                $$
                    p_{ab} + p(p_a + p_b - 2p_{ab}) \le p_{ab} + 1/2 \cdot (p_a + p_b -2p_{ab}) = (p_a + p_b)/2.
                $$
                Therefore, $r_G \le \sigma \Phi^{-1}((p_a + p_b)/2)$.
                Notice that $\Phi^{-1}$ is convex in $[1/2,+\infty)$, so $\Phi^{-1}(p_a) + \Phi^{-1}(p_b) \ge 2\Phi^{-1}((p_a + p_b)/2)$, i.e., $r_H \ge r_G$.
            \end{enumerate}
        \end{proof}
    
\section{Robustness for Smoothed ML Ensemble: Statistical Robustness Perspective}

    \label{adx-sec:3}

    In this appendix, we study the robustness of ensemble models from the statistical robustness perspective.
    This perspective is motivated from \Cref{lem:rand-smooth-vanilla-certify}, where the certified robust radius of a model smoothed with Gaussian distribution $\varepsilon\sim\gN(0,\sigma^2\mI_d)$ is directly proportional to the probability of the original~(unsmoothed) model predicting the correct class under such noise.
    
    We first define the notation of statistical robustness in \Cref{adx-subsec:3-1}; 
    then we show and prove the certified robustness guarantees of \shortWeightedEnsemble, \shortOurEnsemble, and single models respectively in \Cref{adx-subsec:3-1-b};
    next we use these results to compare these ensembles under both general assumptions~(\Cref{adx-subsec:3-2}) and more specific uniform distribution assumptions~(\Cref{adx-subsec:3-3}) where several findings are also discussed;
    finally, we conduct extensive numerical experiments to verify all these findings in \Cref{adx-subsec:3-4}.
    

    \subsection{Definitions of Statistical Robustness}
    
        \label{adx-subsec:3-1}

        \begin{definition}[$(\rvepsilon,\,p)$-Statistical Robust]
            \label{def:statistical-robustness}
            Given a random variable $\rvepsilon$ and model $F: \sR^d \mapsto [C]$, at point $\vx_0$ with ground truth label $y_0$, we call $F$ is $(\rvepsilon,\,p)$-statistical robust if $\Pr_\rvepsilon(F(\vx_0 + \rvepsilon) = y_0) \ge p$.
        \end{definition}
    
    \begin{remark}
        Note that based on \Cref{lem:rand-smooth-vanilla-certify}, when $\rvepsilon \sim \gN(0, \sigma^2\mI_d)$,
        if $F$ is $(\rvepsilon,\,p)$-statistical robust at point $\vx_0$, the smoothed model $G^\rvepsilon_F$ over $F$ is $(\sigma \Phi^{-1}(p))$-robust at point $\vx_0$.
    \end{remark}
        
        
        The following three definitions are used in the theorem statements in the following subsections.
        They can be viewed as the ``confidence margins'' under noised inputs $\vx_0 + \varepsilon$ for single model and ensemble respectively.
        
        \begin{definition}[$(\rvepsilon,\lambda,p)$-Single Confident]
            \label{def:strong-runner-up-single-model}
            Given a classification model $F$.
            If at point $\vx_0$ with ground-truth label $y_0$ and the random variable $\rvepsilon$, we have
            $$
                \Pr_{\rvepsilon} \left( \max_{y_j \in [C]: y_j \neq y_0} f(\vx_0 + \rvepsilon)_{y_j} \le \lambda (1 - f(\vx_0 + \rvepsilon)_{y_0}) \right) = 1 - p,
            $$
           we call $F$  $(\rvepsilon,\lambda,p)$-single confident at point $\vx_0$.
        \end{definition}
        
        \begin{definition}[$(\rvepsilon,\lambda,p)$-\shortWeightedEnsemble Confident]
            \label{def:strong-runner-up-weighted}
            Let $\Mweight$ be a weighted ensemble defined over base models $\{F_i\}_{i=1}^N$ with weights $\{w_i\}_{i=1}^N$.
            If at point $\vx_0$ with ground-truth  $y_0$ and random variable $\rvepsilon$, we have
            \begin{equation}
                \begin{aligned}
                & \Pr_{\rvepsilon} \left(  \max_{y_j \in [C]: y_j \neq y_0} \left(\sum_{i=1}^N w_if_i(\vx_0 + \vepsilon)_{y_j}\right) \le 
                 \lambda \sum_{i=1}^N w_i  \left( 1 - f_i(\vx_0 + \vepsilon)_{y_0} \right) \right) = 1 - p,
                \end{aligned}
                \label{eq:def-7}
            \end{equation}
            we call weighted ensemble $\Mweight$  $(\rvepsilon,\lambda,p)$-\shortWeightedEnsemble confident at point $\vx_0$.
        \end{definition}
        
        \begin{definition}[$(\rvepsilon,\lambda,p)$-\shortOurEnsemble Confident]
            \label{def:strong-runner-up-maximum-margin}
            Let $\Mmme$ be a max-margin ensemble over $\{F_i\}_{i=1}^N$.
            If at point $\vx_0$ with ground-truth  $y_0$ and  random variable $\rvepsilon$, we have
            \begin{equation}
                \begin{aligned}
                & \Pr_{\rvepsilon} \left( \bigwedge_{i\in [N]} \left(  \max_{y_j \in [C]: y_j \neq y_0} f_i(\vx_0 + \rvepsilon)_{y_j} \le 
                \lambda (1 - f_i(\vx_0 + \rvepsilon)_{y_0}) \right) \right) 
                 = 1 - p,
                \label{eq:def-8}
                \end{aligned}
            \end{equation}
            we call max-margin ensemble $\Mmme$ $(\rvepsilon,\lambda,p)$-\shortOurEnsemble confident at point $\vx_0$.
        \end{definition}
        
            Note that the confidence of every single model lies in the probability simplex, and $\lambda$  reflects 
            the confidence portion that a wrong prediction class can take beyond the true class ($1-f_i(\vx_0+\rvepsilon)$).
        
        To reduce ambiguity, we usualy use $\lambda_1$ in \shortWeightedEnsemble Confident, $\lambda_2$ in \shortOurEnsemble Confident, and $\lambda_3$ in Single Confident.
        Note that given $\lambda_1$ is the weighted average and $\lambda_2$  the maximum over $\lambda$'s of all base models, under the same $p$, $\lambda_1/\lambda_2 \le 1$.
        Furthermore, \textit{under the same $p$, $\lambda_1 / \lambda_2$ reflects the adversarial transferability~\citep{papernot2016transferability} among base models}: If the transferability is high, the confidence scores of  base models are similar ($\lambda$'s are similar), and thus $\lambda_1$ is large resulting in large $\lambda_1 / \lambda_2$.
            On the other hand, when the transferability is low, the confidence scores are diverse ($\lambda$'s are diverse), and thus $\lambda_1$ is small resulting in small $\lambda_1 / \lambda_2$.
        
        The following lemma is frequently used in our following proofs:
        \begin{lemma}
            \label{fact:general-bound-fact}
            Suppose the random variable $X$ satisfies $\E X > 0$, $\Var(X) < \infty$ and for any $x \in \sR_+$, $\Pr(X \ge \E X + x) = \Pr(X \le \E X - x)$, then
            $$
                \Pr(X \le 0) \le \frac{\Var(X)}{2(\E X)^2}.
            $$
        \end{lemma}
    
        \begin{proof}[Proof of \Cref{fact:general-bound-fact}]
            Apply Chebyshev's inequality on random variable $X$ and notice that $X$ is symmetric, then we can easily observe this lemma.
        \end{proof}
        
        Now we are ready to present the certified robustness for different ensemble models.
    
    \subsection{Statistical Certified Robustness Guarantees}
        
        \label{adx-subsec:3-1-b}
        
        The main results in this subsection are \Cref{thm:lower-bound-weighted-ensemble} and \Cref{thm:lower-bound-maximum-margin}.
        
        \subsubsection{Certified Robustness for Single Model}
        
            As the start point, we first show a direct proposition stating the certified robustness guarantee of the single model.
        
            \begin{adxproposition}[Certified Robustness for Single Model]
                \label{prop:lower-bound-single}
                Let $\rvepsilon$ be a random variable.
                Let $F$ be a classification model, which is $(\rvepsilon, \lambda_3, p)$-single confident.
                Let $\vx_0 \in \sR^d$ be the input with ground-truth  $y_0 \in [C]$.
                Suppose $f(\vx_0+\rvepsilon)_{y_0}$ follows symmetric distribution with mean $\mu$ and variance $s^2$, where $\mu > (1 + \lambda_3^{-1})^{-1}$.
                We have
                $$
                    \Pr_\rvepsilon(F(\vx_0 + \rvepsilon) = y_0) \ge
                    1 - p - \dfrac{s^2}{2(\mu - (1 + \lambda_3^{-1})^{-1})^2}.
                $$
            \end{adxproposition}
            
            \begin{proof}[Proof of \Cref{prop:lower-bound-single}]
                We consider the distribution of quantity $Y := f(\vx_0 + \rvepsilon)_{y_0} - \lambda_3 (1 - f(\vx_0 + \rvepsilon)_{y_0})$.
                Since the model $F$ is $(\rvepsilon, \lambda_3, p)$-single confident, with probability $1-p$,
                $Y \le f(\vx_0 + \rvepsilon)_{y_0} - \max_{y_j \in [C]: y_j \neq y_0} f(\vx_0 + \rvepsilon)_{y_j}$.
                We note that since
                $$
                    \E Y = (1+\lambda_3)\mu - \lambda_3,\, 
                    \Var(Y) = (1+\lambda_3)^2 s^2,
                $$
                from \Cref{fact:general-bound-fact},
                $$
                    \Pr(Y \le 0) \le \dfrac{s^2}{2(\mu - (1 + \lambda_3^{-1})^{-1})^2}.
                $$
                Thus,
                $$
                \begin{aligned}
                    \Pr(F(\vx_0+\rvepsilon)=y_0) & = 1 - \Pr(F(\vx_0+ \rvepsilon) \neq y_0) \\
                    &= 1 - \Pr\left(f(\vx_0 + \rvepsilon)_{y_0} - \max_{y_j \in [C]: y_j \neq y_0} f(\vx_0 + \rvepsilon)_{y_j} < 0\right) \\
                    & \ge 1 - p - \Pr(Y \le 0) \\
                    & \ge 1 - p - \dfrac{s^2}{2(\mu - (1 + \lambda_3^{-1})^{-1})^2}.
                \end{aligned}
                $$
            \end{proof}
    
        \subsubsection{Certified Robustness for Ensembles}
        
            Now we are ready to prove the certified robustness of the \weightedEnsemble and \ourEnsemble~(\Cref{thm:lower-bound-weighted-ensemble,thm:lower-bound-maximum-margin}).
        
            In the following text, we first define statistical margins for both \shortWeightedEnsemble and \shortOurEnsemble, and point out their connections to the notion of $(\rvepsilon,p)$-Statistical Robust.
            Then, we reason about the expectation, variance, and tail bounds of the statistical margins.
            Finally, we derive the certified robustness from the statistical margins.
            
            \begin{adxdefinition}[$\hat X_1$; Statistical Margin for \shortWeightedEnsemble $\Mweight$]
                \label{adx-def:X1}
                Let $\Mweight$ be \weightedEnsemble defined over base models $\{F_i\}_{i=1}^N$ with weights $\{w_i\}_{i=1}^N$.
                Suppose $\Mweight$ is $(\rvepsilon,\lambda_1,p)$-\shortWeightedEnsemble-confident.
                We define random variable $\hat X_1$ which is depended by random variable $\rvepsilon$:
                \begin{equation}
                    \label{adx-eq:def-A}
                    \hat X_1(\vepsilon) := (1 + \lambda_1) \sum_{j=1}^N w_j f_j (\vx_0 + \vepsilon)_{y_0} - \lambda_1 \|\vw\|_1.
                \end{equation}
            \end{adxdefinition}
            
            \begin{adxdefinition}[$\hat X_2$; Statistical Margin for \shortOurEnsemble $\Mmme$]
                \label{adx-def:X2}
                Let $\Mmme$ be \ourEnsemble defined over base models $\{F_i\}_{i=1}^N$.
                Suppose $\Mmme$ is $(\rvepsilon,\lambda_2,p)$-\shortOurEnsemble-confident.
                We define random variable $\hat X_2$ which is depended by random variable $\rvepsilon$:
                \begin{equation}
                    \label{adx-eq:def-B}
                    \hat X_2(\vepsilon) := (1 + \lambda_2) \left( \max_{i\in [N]} f_i(\vx_0 + \vepsilon)_{y_0} + \min_{i \in [N]} f_i(\vx_0 + \vepsilon)_{y_0} \right) - 2\lambda_2.
                \end{equation}
            \end{adxdefinition}
    
            We have the following observation:
            \begin{lemma}
                \label{fact:connection-margin-robustness}
                For \weightedEnsemble,
                $$
                    \Pr_{\rvepsilon}~(\Mweight(\vx_0 + \rvepsilon) = y_0) \ge 1 - p - \Pr_{\rvepsilon}~(\hat X_1(\rvepsilon) < 0).
                $$
                For \ourEnsemble,
                $$
                    \Pr_{\rvepsilon}~(\Mmme(\vx_0 + \rvepsilon) = y_0) \ge 1 - p -  \Pr_{\rvepsilon}~(\hat X_2(\rvepsilon) < 0).
                $$
            \end{lemma}
            \begin{proof}[Proof of \Cref{fact:connection-margin-robustness}]
                (1) For \weightedEnsemble, we define the random variable $X_1$:
                $$
                    X_1(\vepsilon) := \min_{y_i\in [C]: y_i \neq y_0} \sum_{j=1}^N w_j f_j^{y_0/y_i}(\vx_0 + \vepsilon).
                $$
                Since $\Mweight$ is $(\rvepsilon,\lambda_1,p)$-\shortWeightedEnsemble-confident, from \Cref{def:strong-runner-up-weighted}, with probability $1 - p$, we have
                $$
                \begin{aligned}
                    X_1(\rvepsilon) & \ge \sum_{j=1}^N w_j \left( f_j(\vx_0+\rvepsilon)_{y_0} - \lambda_2(1 - f_j(\vx_0+\rvepsilon)_{y_0}) \right) \\
                    & = (1 + \lambda_2) \sum_{j=1}^N w_j f_j(\vx_0+\rvepsilon)_{y_0} - \lambda_1 \|\vw\|_1 = \hat X_1(\rvepsilon).
                \end{aligned}
                $$
                Therefore,
                $$
                    \Pr_{\rvepsilon} (\Mweight(\vx_0 + \rvepsilon) = y_0) = \Pr_\rvepsilon (X_1(\rvepsilon) \ge 0) \ge 1 - p - \Pr_\rvepsilon(\hat X_2(\rvepsilon) < 0).
                $$
                
                (2)~For \ourEnsemble, we define the random variable $X_2$:
                $$
                    X_2(\vepsilon) := \max_{i \in [N]} \min_{y_i \in [C]: y_i \neq y_0} f_i^{y_0/y_i} (\vx_0 + \vepsilon)
                    +
                    \min_{i \in [N]} \min_{y_i \in [C]: y_i \neq y_0} f_i^{y_0/y_i} (\vx_0 + \vepsilon).
                $$
                Similarly, since $\Mmme$ is $(\rvepsilon,\lambda_2,p)$-\shortOurEnsemble-confident, from \Cref{def:strong-runner-up-maximum-margin},
                with probability $1 - p$, we have
                $$
                \begin{aligned}
                    X_2(\vepsilon) & \ge \max_{i\in [N]} \left( f_i(\vx_0+\rvepsilon)_{y_0} - \lambda_2(1 - f_i(\vx_0+\rvepsilon)_{y_0}) \right) + \min_{i\in [N]} \left( f_i(\vx_0+\rvepsilon)_{y_0} - \lambda_2(1 - f_i(\vx_0+\rvepsilon)_{y_0}) \right) \\
                    & = (1 + \lambda_2) \left( \max_{i\in [N]} f_i(\vx_0 + \vepsilon)_{y_0} + \min_{i \in [N]} f_i(\vx_0 + \vepsilon)_{y_0} \right) - 2\lambda_2 = \hat X_2(\rvepsilon).
                \end{aligned}
                $$
                Moreover, from \Cref{adxlem:adx-suf-cond-for-ensemble-rob}, we know
                $$
                    \Pr_{\rvepsilon} (\gM(\vx_0 + \rvepsilon) = y_0) \ge \Pr_\rvepsilon(X_2(\rvepsilon) \ge 0) \ge 1 - p - \Pr_\rvepsilon(\hat X_2(\rvepsilon) < 0).
                $$
            \end{proof}
            
            As the result, to quantify the statistical robustness of two types of ensembles, we can analyze the distribution of statistical margins $\hat X_1$ and $\hat X_2$.
            
            \begin{lemma}[Expectation and variance of $\hat X_1$ and $\hat X_2$]
                \label{lemma:expectation-variance-compare}
                Let $\hat X_1$ and $\hat X_2$ be defined by \Cref{adx-def:X1} and \Cref{adx-def:X2} respectively.
                Assume $\{f_i(\vx_0+\rvepsilon)_{y_0}\}_{i=1}^N$ are i.i.d. and follow symmetric distribution with mean $\mu$ and variance $s^2$.
                Define $s_f^2 = \Var(\min_{i\in [N]} f_i(\vx_0+\rvepsilon)_{y_0})$.
                We have
                $$
                    \begin{aligned}
                        \E\, \hat X_1(\rvepsilon) = & (1+\lambda_1)\|\vw\|_1 \mu - \lambda_1 \|\vw\|_1, & &
                        \Var\, \hat X_1(\rvepsilon) = (1 + \lambda_1)^2 s^2 \|\vw\|_2^2, \\
                        \E\, \hat X_2(\rvepsilon) = & 2(1+\lambda_2)\mu - 2\lambda_2, & &
                        \Var\, \hat X_2(\rvepsilon) \le 4(1+\lambda_2)^2 s_f^2.
                    \end{aligned}
                $$
            \end{lemma}
            
            \begin{proof}[Proof of \Cref{lemma:expectation-variance-compare}]
                $$
                \begin{aligned}
                    \E \hat X_1(\rvepsilon) & = (1 + \lambda_1) \sum_{j=1}^N \E w_j f_j (\vx_0 + \vepsilon)_{y_0} - \lambda_1 \|\vw\|_1 = (1+\lambda_1) \|\vw\|_1 \mu - \lambda_1 \|\vw\|_1; \\
                    \Var \hat X_1(\rvepsilon) & = (1 + \lambda_1)^2 \sum_{j=1}^N w_j^2 \Var(f_j (\vx_0 + \vepsilon)_{y_0}) = (1 + \lambda_1)^2 s^2 \|\vw\|_2^2.
                \end{aligned}
                $$
                According to the symmetric distribution property of $\{f_i(\vx_0+\rvepsilon)_{y_0}\}_{i=1}^N$, we have
                $$
                    \begin{aligned}
                        \E\, \hat X_2(\rvepsilon) & = \E (1 + \lambda_2) \left( \max_{i\in [N]} f_i(\vx_0 + \vepsilon)_{y_0} + \min_{i \in [N]} f_i(\vx_0 + \vepsilon)_{y_0} \right) - 2\lambda_2 \\
                        & = 2(1+\lambda_2)\mu - 2\lambda_2.
                    \end{aligned}
                $$
                Also, due the symmetry, we have
                $$
                    \Var\left( \min_{i\in [N]} f_i(\vx_0+\rvepsilon)_{y_0} \right)
                    =
                    \Var\left( \max_{i\in [N]} f_i(\vx_0+\rvepsilon)_{y_0} \right)
                    = s_f^2.
                $$
                As a result,
                $$
                    \Var\, \hat X_2(\rvepsilon) \le (1 + \lambda_2)^2 \cdot 4 s_f^2.
                $$
            \end{proof}
        
            From \Cref{lemma:expectation-variance-compare}, now with \Cref{fact:general-bound-fact}, we are ready to derive the statistical robustness lower bound for \shortWeightedEnsemble and \shortOurEnsemble.
            
            \begin{theorem}[Certified Robustness for \shortWeightedEnsemble]
                \label{thm:lower-bound-weighted-ensemble}
                {Let $\rvepsilon$ be a random variable supported on $\sR^d$}.
                Let $\Mweight$ be a \weightedEnsemble defined over $\{F_i\}_{i=1}^N$ with weights $\{w_i\}_{i=1}^N$.
                The $\Mweight$ is $(\rvepsilon,\lambda_1,p)$-\shortWeightedEnsemble confident.
                Let $\vx_0 \in \sR^d$ be the input with ground-truth label $y_0 \in [C]$.
                Assume $\{f_i(\vx_0+\rvepsilon)_{y_0}\}_{i=1}^N$, the confidence scores across base models for label $y_0$, are i.i.d. and follow symmetric distribution with mean $\mu$ and variance $s^2$, where $\mu > (1+\lambda_1^{-1})^{-1}$.
                We have
                \begin{equation}
                    \label{eq:thm5}
                    \Pr_\rvepsilon(\Mweight(\vx_0 + \rvepsilon) = y_0) \ge
                    1 - p - \dfrac{\|\vw\|_2^2}{\|\vw\|_1^2}\cdot
                    \dfrac{s^2}{2\left(\mu - \left(1+\lambda_1^{-1}\right)^{-1}\right)^2}.
                \end{equation}
            \end{theorem}
            
            \begin{theorem}[Certified Robustness for \shortOurEnsemble]
                \label{thm:lower-bound-maximum-margin}
                Let $\rvepsilon$ be a random variable.
                Let $\Mmme$ be a \ourEnsemble defined over $\{F_i\}_{i=1}^N$.
                The $\Mmme$ is $(\rvepsilon,\lambda_2,p)$-\shortOurEnsemble confident.
                Let $\vx_0 \in \sR^d$ be the input with ground-truth label $y_0 \in [C]$.
                Assume $\{f_i(\vx_0+\rvepsilon)_{y_0}\}_{i=1}^N$, the confidence scores across base models for label $y_0$, are i.i.d. and follow symmetric distribution with mean $\mu$ where $\mu > (1 + \lambda_2^{-1})^{-1}$.
                Define $s_f^2 = \Var(\min_{i\in [N]} f_i(\vx_0+\rvepsilon)_{y_0})$.
                We have
                \begin{equation}
                    \label{eq:thm6}
                    \Pr_\rvepsilon(\Mmme(\vx_0+\rvepsilon) = y_0)
                    \ge
                    1 - p - \dfrac{s_f^2}{2\left(\mu-\left(1 + \lambda_2^{-1}\right)^{-1}\right)^2}.
                \end{equation}
            \end{theorem}
        
        \begin{proof}[Proof of \Cref{thm:lower-bound-weighted-ensemble,thm:lower-bound-maximum-margin}]
            Combining \Cref{lemma:expectation-variance-compare,fact:general-bound-fact,fact:connection-margin-robustness}, we get the theorem.
        \end{proof}
        
        \begin{remark}
            \Cref{thm:lower-bound-weighted-ensemble,thm:lower-bound-maximum-margin} provide two statistical robustness lower bounds for both types of ensembles, which is shown to be able to translate to certified robustness.
            
            For the \weightedEnsemble, noticing that $\hat X_1$ is the weighted sum of several independent variables, we can further apply McDiarmid's Inequality to get another bound
            $$
                \Pr_\rvepsilon(\Mweight(\vx_0 + \rvepsilon) = y_0) \ge 1 - p - \exp\left(-2\dfrac{\|\vw\|_1^2}{\|\vw\|_2^2} \left(\mu - \left(1 + \lambda_1^{-1}\right)^{-1}\right)^2 \right),
            $$
            which is tighter than \Cref{eq:thm5} when $\|\vw\|_1^2 / \|\vw\|_2^2$ is large.
            For average weighted ensemble, $\|\vw\|_1^2 / \|\vw\|_2^2 = N$.
            Thus, when $N$ is large, this bound is tighter.
        
            Both theorems are applicable under the i.i.d. assumption of confidence scores.
            The another assumption $\mu > \max\{(1 + \lambda_1^{-1})^{-1}, (1 + \lambda_2^{-1})^{-1}\}$ insures that both ensembles have higher probability of predicting the true class rather than other classes, i.e., the ensembles have non-trivial clean accuracy.
        \end{remark}

    \subsection{Comparison of Certified Robustness}
        \label{adx-subsec:3-2}
        We first show and prove an important lemma.
        Then, based on the lemma and \Cref{thm:lower-bound-weighted-ensemble,thm:lower-bound-maximum-margin}, we derive the comparison corollary.
        
        \begin{lemma}
            \label{lemma:comparison-foundation}
            For $\mu,\lambda_1,\lambda_2,C > 0$,
            when $ \max\{\lambda_1 / (1 + \lambda_1), \lambda_2 / (1 + \lambda_2)\} < \mu \le 1$,
            and $C < 1$, 
            we have
            \begin{equation}
                \dfrac{\mu - (\lambda_2^{-1} + 1)^{-1}}{\mu - (\lambda_1^{-1} + 1)^{-1}} < C
                \iff
                \dfrac{\lambda_1}{\lambda_2} < \lambda_2^{-1}
                \left( \left( C^{-1} \left( \mu - \dfrac{\lambda_2}{1 + \lambda_2} \right) + 1 - \mu \right)^{-1} - 1 \right).
                \label{adx-eq:comparison-foundation}
            \end{equation}
        \end{lemma}
        
        \begin{proof}[Proof of \Cref{lemma:comparison-foundation}]
            $$
            \begin{aligned}
                & \dfrac{\mu - (\lambda_2^{-1} + 1)^{-1}}{\mu - (\lambda_1^{-1} + 1)^{-1}} < C \\
                \iff & \dfrac{1}{\lambda_2^{-1} + 1} - \dfrac{C}{\lambda_1^{-1}+1} > \mu (1-C) \\
                \iff & \dfrac{\lambda_1/\lambda_2}{\lambda_2^{-1} + \lambda_1/\lambda_2} <
                \dfrac{C^{-1}}{\lambda_2^{-1}+1} - \mu(C^{-1}-1) \\
                \iff & \dfrac{\lambda_1}{\lambda_2} 
                \left( 1 - \mu + C^{-1} \left( \mu - \frac{1}{\lambda_2^{-1}+1} \right) \right) < \lambda_2^{-1} \left( C^{-1} \left( \frac{1}{\lambda_2^{-1}+1} - \mu \right) + \mu \right) \\
                \iff & \dfrac{\lambda_1}{\lambda_2} < \lambda_2^{-1} 
                \dfrac{C^{-1} \left( \frac{1}{\lambda_2^{-1}+1} - \mu \right) + \mu}
                {C^{-1} \left( \mu - \frac{1}{\lambda_2^{-1}+1} \right) + 1 - \mu} \\
                \iff & \dfrac{\lambda_1}{\lambda_2} < \lambda_2^{-1}
                \left(
                    \left( C^{-1} \left( \mu - \frac{\lambda_2}{1 + \lambda_2} \right) + 1 - \mu \right)^{-1} - 1
                \right).
            \end{aligned}
            $$
        \end{proof}
        
        Now we can show and prove the comparison corollary.
        \begin{corollary}[Comparison of Certified Robustness]
            {Let $\rvepsilon$ be a random variable supported on $\sR^d$}.
            Over base models $\{F_i\}_{i=1}^N$, let $\Mmme$ be \ourEnsemble, and $\Mweight$ the \weightedEnsemble with weights $\{w_i\}_{i=1}^N$.
            Let $\vx_0 \in \sR^d$ be the input with ground-truth label $y_0 \in [C]$.
            Assume $\{f_i(\vx_0+\rvepsilon)_{y_0}\}_{i=1}^N$, the confidence scores across base models for label $y_0$, are i.i.d, and follow symmetric distribution with mean $\mu$ and variance $s^2$, where $\mu > \max\{(1+\lambda_1^{-1})^{-1}, (1+\lambda_2^{-1})^{-1}\}$.
            Define $s_f^2 = \Var(\min_{i\in [N]} f_i(\vx_0+\rvepsilon)_{y_0})$ and assume $s_f < s$.
            \begin{itemize}
                \item When
                \begin{equation}
                    \dfrac{\lambda_1}{\lambda_2} < \lambda_2^{-1} \left(
                        \left(
                            \frac{s}{s_f} \left( \mu - \left(1 + \lambda_2^{-1} \right)^{-1} \right) + 1 - \mu
                        \right)^{-1} 
                        -1
                    \right),
                    \label{eq:cor2-1}
                \end{equation}
                for any weights $\{w_i\}_{i=1}^N$, $\Mweight$ has higher certified robustness than $\Mmme$.
                \item When
                \begin{equation}
                    \dfrac{\lambda_1}{\lambda_2} > \lambda_2^{-1} \left(
                        \left(
                            \frac{s}{\sqrt N s_f} \left( \mu - \left(1 + \lambda_2^{-1} \right)^{-1} \right) + 1 - \mu
                        \right)^{-1} 
                        -1
                    \right),
                    \label{eq:cor2-2}
                \end{equation}
                for any weights $\{w_i\}_{i=1}^N$, $\Mmme$ has higher certified robustness than $\Mweight$.
            \end{itemize}
            Here, the certified robustness is given by \Cref{thm:lower-bound-weighted-ensemble,thm:lower-bound-maximum-margin}.
            \label{cor:comparision-statistical-lower-bound}
        \end{corollary}
        
        \begin{proof}[Proof of \Cref{cor:comparision-statistical-lower-bound}]
            (1) According to \Cref{lemma:comparison-foundation}, we have
            $$
            \begin{aligned}
                & \dfrac{\lambda_1}{\lambda_2} < \lambda_2^{-1} \left(
                    \left(
                        \frac{s}{s_f} \left( \mu - \left(1 + \lambda_2^{-1} \right)^{-1} \right) + 1 - \mu
                    \right)^{-1} 
                    -1
                \right) \\
                \Longrightarrow & \dfrac{\mu - (\lambda_2^{-1}+1)^{-1}}{\mu - (\lambda_1^{-1}+1)^{-1}}
                <
                \dfrac{s_f}{s} \\
                \Longrightarrow & 
                \sqrt{\frac{\|\vw\|_2^2}{\|\vw\|_1^2}}
                \dfrac{\mu - (\lambda_2^{-1}+1)^{-1}}{\mu - (\lambda_1^{-1}+1)^{-1}}
                <
                \dfrac{s_f}{s} \\
                \Longrightarrow & 
                \dfrac{\|\vw\|_2^2}{\|\vw\|_1^2}\cdot
                \dfrac{s^2}{2\left(\mu - \left(1+\lambda_1^{-1}\right)^{-1}\right)^2} 
                < \dfrac{s_f^2}{2\left(\mu-\left(1 + \lambda_2^{-1}\right)^{-1}\right)^2}.
            \end{aligned}
            $$
            According to \Cref{thm:lower-bound-weighted-ensemble,thm:lower-bound-maximum-margin}, we know the RHS in \Cref{eq:thm5} is larger than the RHS in \Cref{eq:thm6}, i.e., $\gM_{\mathrm{\shortWeightedEnsemble}}$ has higher certified robustnesss than $\gM_{\mathrm{\shortOurEnsemble}}$.
            
            (2) According to \Cref{lemma:comparison-foundation}, we have
            $$
            \begin{aligned}
                & \dfrac{\lambda_1}{\lambda_2} > \lambda_2^{-1} \left(
                    \left(
                        \frac{s}{\sqrt N s_f} \left( \mu - \left(1 + \lambda_2^{-1} \right)^{-1} \right) + 1 - \mu
                    \right)^{-1} 
                    -1
                \right) \\
                \Longrightarrow & \dfrac{\mu - (\lambda_2^{-1}+1)^{-1}}{\mu - (\lambda_1^{-1}+1)^{-1}}
                >
                \dfrac{\sqrt N s_f}{s} \\
                \Longrightarrow & 
                \sqrt{\frac{\|\vw\|_2^2}{\|\vw\|_1^2}}
                \dfrac{\mu - (\lambda_2^{-1}+1)^{-1}}{\mu - (\lambda_1^{-1}+1)^{-1}}
                >
                \dfrac{s_f}{s} \\
                \Longrightarrow & 
                \dfrac{\|\vw\|_2^2}{\|\vw\|_1^2}\cdot
                \dfrac{s^2}{2\left(\mu - \left(1+\lambda_1^{-1}\right)^{-1}\right)^2} 
                > \dfrac{s_f^2}{2\left(\mu-\left(1 + \lambda_2^{-1}\right)^{-1}\right)^2}.
            \end{aligned}
            $$
            According to \Cref{thm:lower-bound-weighted-ensemble,thm:lower-bound-maximum-margin}, we know the RHS in \Cref{eq:thm6} is larger than the RHS in \Cref{eq:thm5}, i.e., $\gM_{\mathrm{\shortOurEnsemble}}$ has higher certified robustnesss than $\gM_{\mathrm{\shortWeightedEnsemble}}$.
        \end{proof}
        
        \begin{remark}
          
            (1)~
            Given that $\lambda_1/\lambda_2$ reflects the adversarial transferability among base models~\Cref{adx-subsec:3-1},
            the corollary implies that, \shortOurEnsemble is more robust when the transferability is high;
            \shortWeightedEnsemble is more robust when the transferability is low.

            (2)~As we can observe in the proof, there is a gap between \Cref{eq:cor2-1} and \Cref{eq:cor2-2} --- when $\lambda_1/\lambda_2$ lies in between RHS of \Cref{eq:cor2-1} and RHS of \Cref{eq:cor2-2}, it is undetermined which ensemble protocol has higher robustness.
            Indeed, this uncertainty is caused by the adjustable weights $\{w_i\}_{i=1}^N$ of the \weightedEnsemble.
            If we only consider the average ensemble, then this gap is closed:
            $$
                \dfrac{\lambda_1}{\lambda_2} 
                \underset{\text{$\Mweight$ more robust}}{\overset{\text{$\Mmme$ more robust}}{\gtrless}}  
                \lambda_2^{-1} \left(
                    \left(
                        \frac{s}{\sqrt N s_f} \left( \mu - \left(1 + \lambda_2^{-1} \right)^{-1} \right) + 1 - \mu
                    \right)^{-1} 
                    -1
                \right).
            $$
            
            (3)~Note that we assume that $s_f < s$, where
            $s^2$ is the variance of single variable and $s_f^2$ is the variance of minimum of $N$ i.i.d. variables.
            For common symmetry distributions, along with the increase of $N$, $s_f$ shrinks in the order of $O(1/N^B)$ where $B \in (0,2]$.
            Thus, as long as $N$ is large, the assumption $s_f < s$ will always hold.
            An exception is that when these random variables follow the exponential distribution, where $s_f$ does not shrink along with the increase of $N$.
            However, since these random variables are confidence scores which are in $[0,1]$, they cannot obey exponential distribution.
        \end{remark}
    
    \subsection{A Concrete Case: Uniform Distribution}
    
        \label{adx-subsec:3-3}
        
        As shown by \cite{saremi2020provable} (Remark 2.1), when the input dimension $d$ is large, the Gaussian noise $\rvepsilon\sim\gN(0,\sigma^2\mI_d) \approx \mathrm{Unif}(\sigma \sqrt d S_{d-1})$, i.e., $\vx_0 + \rvepsilon$ is highly \emph{uniformly distributed} on the $(d-1)$-sphere centered at $\vx_0$.
        Motivated by this, we study the case where the confidence scores $\{f_i(\vx_0 + \rvepsilon)_{y_0}\}_{i=1}^N$ are also uniformly distributed.
        
        Under this additional assumption, we can further make the certified robustness for the single model and both ensembles more concrete.
        
        \subsubsection{Certified Robustness for Single Model}
        
            \begin{adxproposition}[Certified Robustness for Single Model under Uniform Distribution]
                \label{prop:lower-bound-single-unif}
                {Let $\rvepsilon$ be a random variable supported on $\sR^d$}.
                Let $F$ be a classification model, which is $(\rvepsilon, \lambda_3, p)$-single confident.
                Let $\vx_0 \in \sR^d$ be the input with ground-truth $y_0 \in [C]$.
                Suppose $f(\vx_0+\rvepsilon)_{y_0}$ is uniformly distributed in  $[a,\,b]$.
                We have
                $$
                \begin{aligned}
                    & \Pr_\rvepsilon(F(\rx_0+\rvepsilon) = y_0)
                    \ge
                    1 - p - \mathrm{clip}\left( \dfrac{1 / (1 + \lambda_3^{-1}) - a}{b - a} \right), \\
                    \text{where}\quad &
                    \mathrm{clip}(x) = \max(\min(x, 1), 0).
                \end{aligned}
                $$
            \end{adxproposition}
        
            \begin{proof}[Proof of \Cref{prop:lower-bound-single-unif}]
                We consider the distribution of quantity $Y := f(\vx_0 + \rvepsilon)_{y_0} - \lambda_3 (1 - f(\vx_0 + \rvepsilon)_{y_0})$.
                Since the model $F$ is $(\rvepsilon, \lambda_3, p)$-single confident, with probability $1-p$,
                $Y \le f(\vx_0 + \rvepsilon)_{y_0} - \max_{y_j \in [C]: y_j \neq y_0} f(\vx_0 + \rvepsilon)_{y_j}$.
                At the same time, because $f(\vx_0 + \epsilon)_{y_0}$ follows the distribution $\gU([a,\,b])$,
                $$
                    Y = (1 + \lambda_3) f(\vx_0+\rvepsilon)_{y_0} - \lambda_3
                $$
                follows the distribution $\gU([(1 + \lambda_3)a - \lambda_3,\, (1 + \lambda_3)b - \lambda_3])$.
                Therefore, 
                $$
                    \Pr(Y \le 0) = \mathrm{clip}\left(\dfrac{\lambda_3 - (1 + \lambda_3)a}{(1+\lambda_3)(b-a)} \right).
                $$
                As the result,
                $$
                    \Pr\left( f(\vx_0 + \rvepsilon)_{y_0} - \max_{y_j \in [C]: y_j \neq y_0} f(\vx_0 + \rvepsilon)_{y_j} \le 0 \right) \le
                    p + \mathrm{clip}\left(\dfrac{\lambda_3 - (1 + \lambda_3)a}{(1+\lambda_3)(b-a)} \right),
                $$
                which is exactly
                $$
                    \Pr\left( F(\vx_0 + \rvepsilon) = y_0 \right) \ge 1 - p - \mathrm{clip}\left(\dfrac{\lambda_3 - (1 + \lambda_3)a}{(1+\lambda_3)(b-a)} \right) = 
                    1 - p - \mathrm{clip}\left( \dfrac{1 / (1 + \lambda_3^{-1}) - a}{b - a} \right).
                $$
            \end{proof}
        
        \subsubsection{Certified Robustness for Ensembles}
        
            Still, we define $\hat X_1(\rvepsilon)$ and $\hat X_2(\rvepsilon)$ according to \Cref{adx-def:X1,adx-def:X2}.
            Under the uniform distribution assumption, we have the following lemma.
            \begin{lemma}[Expectation and Variance of $\hat X_1$ and $\hat X_2$ under Uniform Distribution]
                \label{lemma:expectation-variance-compare-uniform}
                Let $\hat X_1$ and $\hat X_2$ be defined by \Cref{adx-def:X1} and \Cref{adx-def:X2} respectively.
                Assume that under the distribution of $\rvepsilon$, the base models' confidence scores for true class $\{f_i(\vx_0+\rvepsilon)_{y_0}\}_{i=1}^N$ are pairwise i.i.d and uniformly distributed in range $[a,\,b]$.
                We have
                $$
                    \begin{aligned}
                        &\E\, \hat X_1(\rvepsilon) =& & \hspace{-0.8em} \dfrac{1}{2} (1+\lambda_1) \|\vw\|_1 (a+b) - \lambda_1 \|\vw\|_1,
                        & & \Var\,\hat X_1(\rvepsilon) = \dfrac{1}{12}(1 + \lambda_1)^2 \|\vw\|_2^2 (b-a)^2, \\
                        &\E\, \hat X_2(\rvepsilon) =& & \hspace{-0.8em} (1 + \lambda_2)(a+b) - 2\lambda_2,\, 
                        & & \Var\,\hat X_2(\rvepsilon) \le (1 + \lambda_2)^2  \dfrac{4}{N+1} \left(\dfrac{2}{N+2} - \dfrac{1}{N+1} \right) (b-a)^2.
                    \end{aligned}
                $$
            \end{lemma}
        
            \begin{proof}[Proof of \Cref{lemma:expectation-variance-compare-uniform}]
                We start from analyzing $\hat X_1$.
                From the definition 
                \begin{equation}
                    \hat X_1(\vepsilon) := (1 + \lambda_1) \sum_{j=1}^N w_j f_j (\vx_0 + \vepsilon)_{y_0} - \lambda_1 \|\vw\|_1
                    \tag{\ref{adx-eq:def-A}}
                \end{equation}
                where $\{f_i(\vx_0 + \epsilon)_{y_0}\}_{i=1}^N$ are i.i.d. variables obeying uniform distribution $\gU([a,\,b])$,
                $$
                \begin{aligned}
                    & \E\, \hat X_1(\vepsilon) = (1 + \lambda_1) \|\vw\|_1 \dfrac{a+b}{2} - \lambda_1 \|\vw\|_1 = \dfrac{1}{2} (1+\lambda_1) \|\vw\|_1 (a+b) - \lambda_1 \|\vw\|_1 , \\
                    & \Var\, \hat X_1(\vepsilon) = (1 + \lambda_1)^2 \sum_{j=1}^N w_j^2 \dfrac{1}{12}(b-a)^2 = \dfrac{1}{12}(1 + \lambda_1)^2 \|\vw\|_2^2 (b-a)^2.
                \end{aligned}
                $$
                
                Now analyze the expectation of $\hat X_2$.
                By the symmetry of uniform distribution, we know
                $$
                    \E\, \hat X_2(\vepsilon) = (1 + \lambda_2) \cdot 2 \E\, f_i(\vx_0 + \vepsilon)_{y_0} - 2\lambda_2 = (1 + \lambda_2)(a+b) - 2\lambda_2.
                $$
                To reason about the variance, we need the following fact:
                \begin{fact}
                    \label{fact:min-uniform-rv}
                    Let $\rx_1,\,\rx_2,\,\dots,\,\rx_n$ be uniformly distributed and independent random variables. Specifically, for each $1 \le i \le n$, $\vx_i \sim \gU([a,\,b])$.
                    Then we have
                    $$
                        \Var \left(\min_{1 \le i \le n} \rx_i\right) = \Var \left( \max_{1 \le i \le n}  \rx_i\right) = \dfrac{1}{n+1} \left(\dfrac{2}{n+2} - \dfrac{1}{n+1}\right) (b-a)^2 .
                    $$
                \end{fact}
                Observing that each i.i.d. $f_i(\vx_0 + \rvepsilon)_{y_0}$ is exactly identical to $\rx_i$ in \Cref{fact:min-uniform-rv}, we have
                $$
                    \Var \left( \max_{i\in [N]} f_i(\vx_0 + \vepsilon)_{y_0} + \min_{i \in [N]} f_i(\vx_0 + \vepsilon)_{y_0} \right) \le \dfrac{4}{N+1} \left(\dfrac{2}{N+2} - \dfrac{1}{N+1} \right) (b-a)^2.
                $$
                Therefore, 
                $$
                    \Var \, \hat X_2(\rvepsilon) \le (1 + \lambda_2)^2  \dfrac{4}{N+1} \left(\dfrac{2}{N+2} - \dfrac{1}{N+1} \right) (b-a)^2.
                $$
            \end{proof}
            
            \begin{proof}[Proof of \Cref{fact:min-uniform-rv}]
                From symmetry of uniform distribution, we know $ \Var \left(\min_{1 \le i \le n} \rx_i\right) = \Var \left( \max_{1 \le i \le n}  \rx_i\right)$. So here we only consider $Y:= \min_{1 \le i \le n} \rx_i$.
                Its CDF $F$ and PDF $f$ can be easily computed:
                $$
                    F(y) = 1 - \Pr\left(\min_i \rx_i \ge y\right) = 1 - \left(\dfrac{b-y}{b-a}\right)^n, \,
                    f(y) = F'(y) = n\dfrac{(b-y)^{n-1}}{(b-a)^n}, \, \mathrm{where}\, y \in [a,\,b].
                $$
                Hence,
                $$
                    \E\, Y = \int_a^b yf(y)\dif y = \dfrac{y(b-y)^n+(n+1)^{-1}(b-y)^{n+1}}{(b-a)^n}\Big|_b^a = a + \dfrac{b-a}{n+1},
                $$
                $$
                \begin{aligned}
                    \E\, Y^2 & = \int_a^b y^2f(y)\dif y = \int_a^b ny^2\dfrac{(b-y)^{n-1}}{(b-a)^n} \dif y \\
                    & =  - \left( \dfrac{b-y}{b-a} \right)^n  y^2 \Big|_a^b + 2 \int_a^b \left(\dfrac{b-y}{b-a}\right)^n y \dif y \\
                    & = - \left( \dfrac{b-y}{b-a} \right)^n  y^2 \Big|_a^b + \dfrac{2}{n+1} \left( -\dfrac{(b-y)^{n+1}}{(b-a)^n}y + \int \dfrac{(b-y)^{n+1}}{(b-a)^n} \dif y \right)\Big|_a^b \\
                    & = - \left( \dfrac{b-y}{b-a} \right)^n  y^2 \Big|_a^b + \dfrac{2}{n+1} \left( -\dfrac{(b-y)^{n+1}}{(b-a)^n}y - \dfrac{1}{n+2} \dfrac{(b-y)^{n+2}}{(b-a)^n} \right)\Big|_a^b \\
                    & = a^2 + \dfrac{2}{n+1} (b-a) a + \dfrac{2}{(n+1)(n+2)}(b-a)^2.
                \end{aligned}
                $$
                As the result, $\Var\,Y = \E Y^2 - (\E Y)^2 = \frac{1}{n+1}\left(\frac{2}{n+2} - \frac{1}{n+1}\right) (b-a)^2$.
            \end{proof}
            
            Now, similarly, we use \Cref{fact:general-bound-fact} to derive the statistical robustness lower bound for \shortWeightedEnsemble and \shortOurEnsemble.
            We omit the proofs since they are direct applications of \Cref{lemma:expectation-variance-compare-uniform}, \Cref{fact:general-bound-fact}, and \Cref{fact:connection-margin-robustness}.
            
             \begin{theorem}[Certified Robustness for \shortWeightedEnsemble under Uniform Distribution]
                Let $\Mweight$ be a \weightedEnsemble defined over $\{F_i\}_{i=1}^N$ with weights $\{w_i\}_{i=1}^N$.
                Let $\vx_0 \in \sR^d$ be the input with ground-truth label $y_0 \in [C]$.
                {Let $\rvepsilon$ be a random variable supported on $\sR^d$}.
                Under the distribution of $\rvepsilon$, suppose $\{f_i(\vx_0+\rvepsilon)_{y_0}\}_{i=1}^N$ are i.i.d. and uniformly distributed in $[a,\,b]$.
                The $\Mweight$ is $(\rvepsilon,\lambda_1,p)$-\shortWeightedEnsemble confident.
                Assume $\frac{a+b}{2} > \frac{1}{1 + \lambda_1^{-1}}$.
                We have
                \begin{equation}
                    \begin{aligned}
                        & \Pr_\rvepsilon(\Mweight(\vx_0+\rvepsilon) = y_0)
                        \ge
                        1 - p -
                        \dfrac{d_\vw K_1^2}{12}, \\
                        \text{where}\quad &
                        d_\vw = \frac{\|\vw\|_2^2}{\|\vw\|_1^2},
                        \ 
                        K_1 = \frac{b-a}{\frac{a+b}{2} - \frac{1}{1 + \lambda_1^{-1}}}.
                    \end{aligned}
                \end{equation}
                \label{thm:lower-bound-weighted-uniform}
            \end{theorem}
        
            \begin{theorem}[Certified Robustness for \shortOurEnsemble under Uniform Distribution]
                Let $\Mmme$ be a \ourEnsemble over $\{F_i\}_{i=1}^N$.
                Let $\vx_0 \in \sR^d$ be the input with ground-truth label $y_0 \in [C]$.
                {Let $\rvepsilon$ be a random variable supported on $\sR^d$}.
                Under the distribution of $\rvepsilon$, suppose $\{f_i(\vx_0+\rvepsilon)_{y_0}\}_{i=1}^N$ are i.i.d. and uniformly distributed in $[a,\,b]$.
                $\Mmme$ is $(\rvepsilon,\lambda_2,p)$-\shortOurEnsemble confident.
                Assume $\frac{a + b}{2} > \frac{1}{1 + \lambda_2^{-1}}$.
                We have
                \begin{equation}
                    \begin{aligned}
                        & \Pr_\rvepsilon(\Mmme(\vx_0+\rvepsilon) = y_0)
                        \ge
                        1 - p - \dfrac{c_N K_2^2}{4},\\
                        \text{where}\quad & 
                        c_N = \frac{2}{N+1} \left( \frac{2}{N+2} - \frac{1}{N+1} \right),
                        \ 
                        K_2 = \frac{b-a}{\frac{a+b}{2} - \frac{1}{1 + \lambda_2^{-1}}}.
                    \end{aligned}
                \end{equation}
                \label{thm:lower-bound-our-uniform}
            \end{theorem}
        
        \subsubsection{Comparison of Certified Robustness for Ensembles}
        
            Now under the uniform distribution, we can also have the certified robustness comparison.
        
            \begin{corollary}[Comparison of Certified Robustness under Uniform Distribution]
            
                Over base models $\{F_i\}_{i=1}^N$, let $\Mmme$ be \ourEnsemble, and $\Mweight$ the \weightedEnsemble  with weights $\{w_i\}_{i=1}^N$.
                Let $\vx_0 \in \sR^d$ be the input with ground-truth label $y_0 \in [C]$.
                {Let $\rvepsilon$ be a random variable supported on $\sR^d$}.
                Under the distribution of $\rvepsilon$, suppose $\{f_i(\vx_0 + \rvepsilon)_{y_0}\}_{i=1}^N$ are i.i.d. and uniformly distributed with mean $\mu$.
                Suppose $\Mweight$ is $(\rvepsilon,\lambda_1,p)$-\shortWeightedEnsemble confident, and $\Mmme$ is $(\rvepsilon,\lambda_2,p)$-\shortOurEnsemble confident.
                Assume $\mu > \max\left\{ \frac{1}{1 + \lambda_1^{-1}},\, \frac{1}{1 + \lambda_2^{-1}} \right\}$.
                \begin{itemize}[leftmargin=*]
                    \item When
                    \begin{equation}
                        \dfrac{\lambda_1}{\lambda_2} < \lambda_2^{-1}\left( \left(
                            (N+1) \sqrt{\frac{N+2}{6N}}
                            \left(\mu - \frac{1}{1+\lambda_2^{-1}}\right) + 1 - \mu\right)^{-1} - 1
                        \right),
                        \label{eq:cor2-1-uniform}
                    \end{equation}
                    $\Mweight$ has higher certified robustness than $\Mmme$.
                
                    \item When 
                    \begin{equation}
                        \dfrac{\lambda_1}{\lambda_2} > \lambda_2^{-1}\left( \left(
                        \frac{N+1}{N}
                        \sqrt{\frac{N+2}{6}}
                        \left(\mu - \frac{1}{1 + \lambda_2^{-1}}\right) + 1 - \mu \right)^{-1} - 1 \right),
                        \label{eq:cor2-2-uniform}
                    \end{equation}
                    $\Mmme$ has higher certified robustness than $\Mweight$.
                    
                    \item When
                    \begin{equation}
                         N > 
                         6\left( 1 - 
                         \dfrac{1}{\mu(1 + \lambda_2^{-1})}
                         \right)^{-2}
                         - 2,
                        \label{eq:cor2-3-uniform}
                    \end{equation}
                    for any $\lambda_1$, 
                   $\Mmme$ has higher or equal certified robustness than $\Mweight$.
                \end{itemize}
                Here, the certified robustness is given by \Cref{thm:lower-bound-weighted-uniform,thm:lower-bound-our-uniform}.
                \label{cor:comparision-statistical-lower-bound-uniform}
            \end{corollary}
            
            \begin{proof}[Proof of \Cref{cor:comparision-statistical-lower-bound-uniform}]
                First, we notice that a uniform distribution with mean $\mu$ can be any distribution $\gU([a,\,b])$ where $(a+b)/2 = \mu$.
                We replace $\mu$ by $(a+b)/2$.
                
                Then (1) and (2) follow from \Cref{lemma:comparison-foundation} similar to the proof of \Cref{cor:comparision-statistical-lower-bound}.

                (3)~Since
                $$
                \begin{aligned}
                     N > 
                     6\left( 1 - 
                     \dfrac{1}{\mu(1 + \lambda_2^{-1})}
                     \right)^{-2}
                     - 2
                     & \Longrightarrow
                     \left(
                     \sqrt{\dfrac{N+2}{6}}
                     \left(\mu - \frac{1}{1 + \lambda_2^{-1}}\right) + 1 - \mu \right)^{-1} < 1 \\
                     & \Longrightarrow 
                     \left(
                     \frac{N+1}{N} \sqrt{\dfrac{N+2}{6}}
                     \left(\mu - \frac{1}{1 + \lambda_2^{-1}}\right) + 1 - \mu \right)^{-1} < 1,
                \end{aligned}
                $$
                the RHS of \Cref{eq:cor2-2-uniform} is smaller than $0$.
                Thus, for any $\lambda_1$, since $\lambda_1/\lambda_2 > 0$, the \Cref{eq:cor2-2} is satisfied.
                According to (2), $\gM_{\mathrm{\shortOurEnsemble}}$ has higher certified robustnesss than $\gM_{\mathrm{\shortWeightedEnsemble}}$.
            \end{proof}

            \begin{remark}
                Comparing to the general corollary~(\Cref{cor:comparision-statistical-lower-bound}), under the uniform distribution, we have an additional finding that when $N$ is sufficiently large, we will always have higher certified robustness for \ourEnsemble than \weightedEnsemble.
                This is due to the more efficient variance reduction of \ourEnsemble than \weightedEnsemble.
                As shown in \Cref{lemma:expectation-variance-compare-uniform}, the quantity $\Var \hat X(\rvepsilon)/(\E \hat X(\rvepsilon))^2$ for \weightedEnsemble is $\Omega(1/N)$, while for \ourEnsemble is $O(1/N^2)$.
                As the result, when $N$ becomes larger, \ourEnsemble has higher certified robustness.
                
                {
                    
                    We use uniform assumption here to give an illustration in a specific regime.
                    We think it would be an interesting future direction to generalize the analysis to other distributions such as the Gaussian distribution that corresponds to locally linear classifiers. 
                    The result from these distribution may be derived from their specific concentration bound for maximum/minimum i.i.d. random variables as discussed at the end of \Cref{adx-subsec:3-2}. 
                } 
            \end{remark}

    \subsection{Numerical Experiments}
    
        \label{adx-subsec:3-4}
        
        To validate and give more intuitive explanations for our theorems, we present some numerical experiments.
        
        \subsubsection{Ensemble Comparison from Numerical Sampling}
        
            As discussed in \Cref{adx-subsec:3-1}, $\lambda_1/\lambda_2$ reflects the transferability across base models.
            It is challenging to get enough amount of different ensembles of various transferability levels while keeping all other variables controlled.
            Therefore, we simulate the transferability of ensembles numerically by varying $\lambda_1/\lambda_2$~(see the definitions of $\lambda_1$ and $\lambda_2$ in \Cref{def:strong-runner-up-weighted,def:strong-runner-up-maximum-margin}), and sampling the confidence scores $\{f_i(\vx_0+\rvepsilon)_{y_0}\}$ and $\{\max_{j\in [C]: j\neq y_0} f_i(\vx_0+\rvepsilon)_j\}$ under determined $\lambda_1$ and $\lambda_2$.
            For each level of $\lambda_1/\lambda_2$, with the samples, we compute the certified robust radius $r$ using randomized smoothing~(\Cref{lem:rand-smooth-vanilla-certify}) and compare the radius difference of \weightedEnsemble and \ourEnsemble.
            According to \Cref{cor:comparision-statistical-lower-bound}, we should observe the tendency that along with the increase of transferability $\lambda_1/\lambda_2$, \ourEnsemble would gradually become more certifiably robust than \weightedEnsemble.
            
\begin{figure}[!t]
    \begin{subfigure}{.33\textwidth}
    \includegraphics[width=\textwidth]{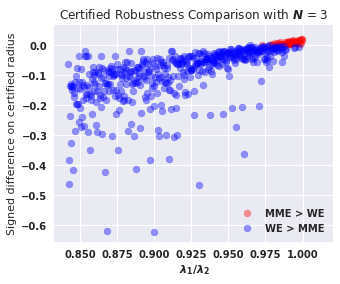}
    \caption{\# of base models $N=3$}
    \end{subfigure}
   \begin{subfigure}{.33\textwidth}
    \includegraphics[width=\textwidth]{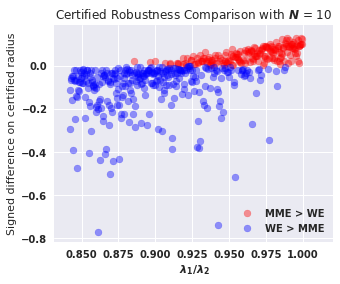}
    \caption{\# of base models $N =10$}
    \end{subfigure}
    \begin{subfigure}{.33\textwidth}
    \includegraphics[width=\textwidth]{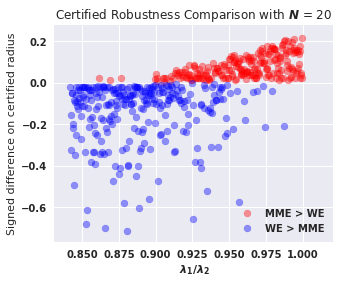}
    \caption{\# of base models $N=20$}
    \end{subfigure}
    \caption{Signed \emph{certified robust radius} difference between MME and WE by $\lambda_1/\lambda_2$ under different numbers of base models $N$. Here we fix $\lambda_2$ to be $0.95$ and uniformly sample $\lambda_1 \in [0.8, 0.95)$. The confidence score for the true class on each base model is uniformly sampled from $[a, b]$, where $a$ is sampled from $[0.3, 1.0)$ and $b$ is sampled from $[a, 1.0)$ uniformly for each instance. \textcolor{blue}{\textbf{Blue}} points correspond to the negative radius difference (i.e., WE has larger radius than MME) and \textcolor{red}{\textbf{Red}} points correspond to the positive radius difference (i.e., MME has larger radius than WE). }
    \label{fig:transferability}
\end{figure}

            \Cref{fig:transferability} verifies the trends: with the increase of $\lambda_1/\lambda_2$, MME model tends to achieve higher certified radius than WE model.
            Moreover, we notice that under the same $\lambda_1/\lambda_2$, with the larger number of base models $N$, the \shortOurEnsemble tends to be relatively more certifiably robust compared with \shortWeightedEnsemble.
            This is because we sample the confidence score uniformly and under the uniform distribution, \shortOurEnsemble tends to be more certifiably robust than \shortWeightedEnsemble when the number of base models $N$ becomes large, according to \Cref{cor:comparision-statistical-lower-bound-uniform}.
            
            The concrete number settings of $\lambda_1, \lambda_2$, and the sampling interval of confidence scores are entailed in the caption of \Cref{fig:transferability}.
            
        \subsubsection{Ensemble Comparison from Certified Robustness Plotting}
        
\begin{figure}
                \centering
                
                \begin{subfigure}[b]{0.45\textwidth}
                    \includegraphics[width=\textwidth]{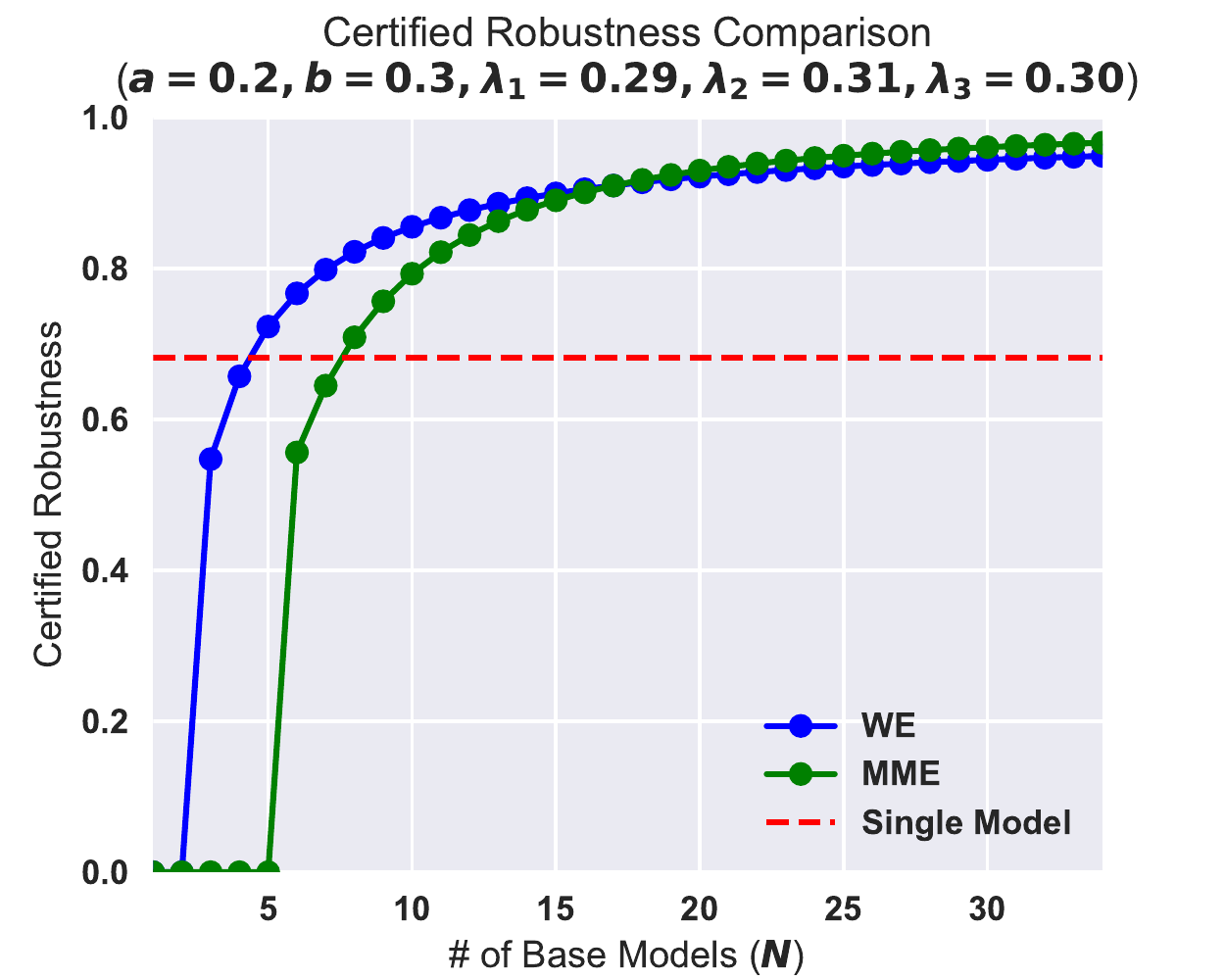}
                    \caption{$[a,\,b] = [0.2,\,0.3], \lambda_1 = 0.29, \lambda_2 = 0.31, \lambda_3 = 0.30$.}
                \end{subfigure}
                \hfill
                \begin{subfigure}[b]{0.45\textwidth}
                    \includegraphics[width=\textwidth]{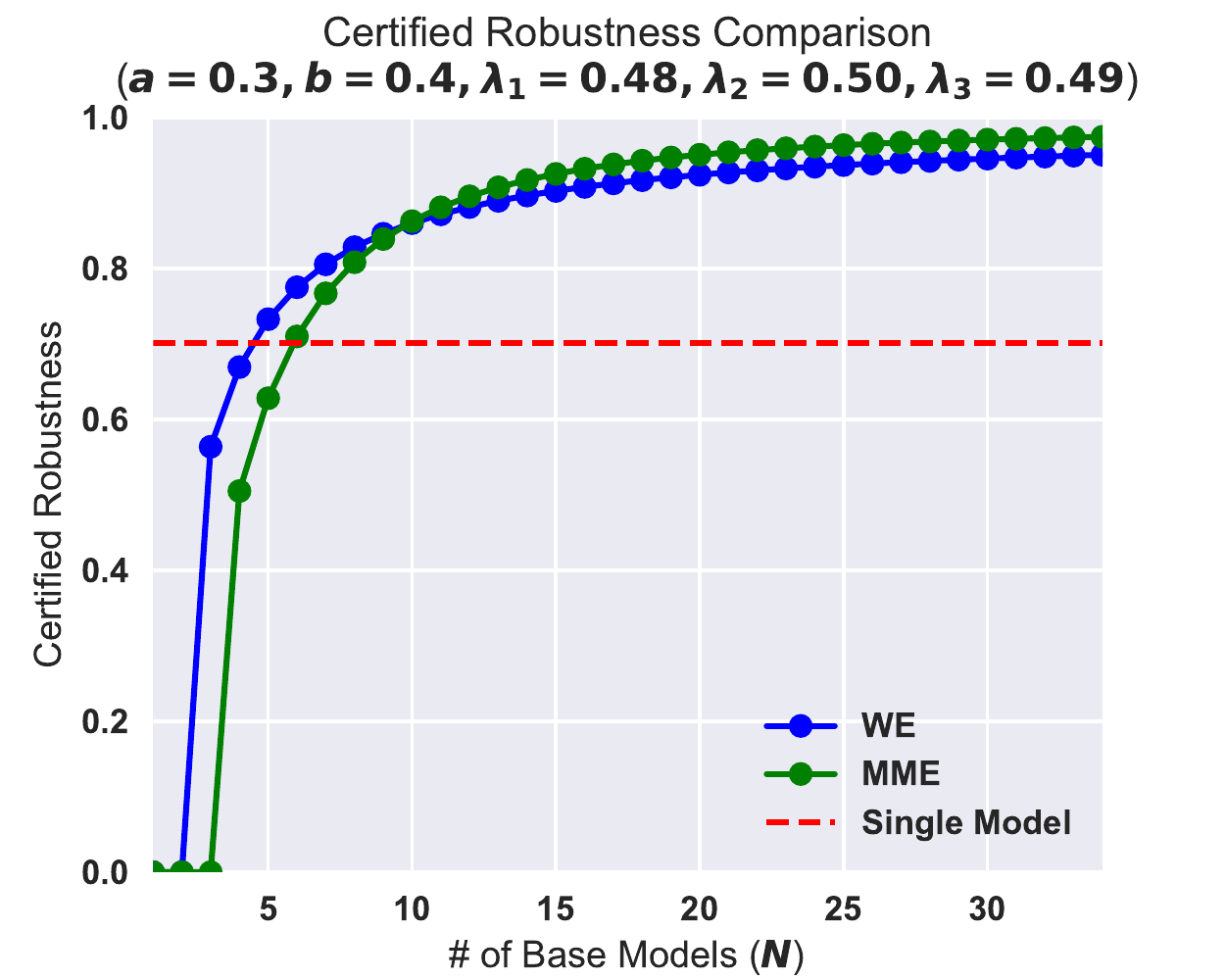}
                    \caption{$[a,\,b] = [0.3,\,0.4], \lambda_1 = 0.48, \lambda_2 = 0.50, \lambda_3 = 0.49$.}
                \end{subfigure}
                
                \caption{Comparison of certified robustness (in terms of certified robust radius) of \ourEnsemble, \weightedEnsemble, and single model under concrete numerical settings.
                The $y$-axis is the certified robustness and the $x$-axis is the number of base models.
                The confidence score for the true class is uniformly distributed in $[a,\,b]$.
                The \weightedEnsemble~(shown by \textcolor{blue}{\textbf{blue line}}) is $(\rvepsilon,\lambda_1,0.01)$-\shortWeightedEnsemble confident;
                the \ourEnsemble~(shown by \textcolor{darkgreen}{\textbf{green line}}) is $(\rvepsilon,\lambda_2,0.01)$-\shortOurEnsemble confident;
                and the single model~(shown by \textcolor{red}{\textbf{red line}}) is $(\rvepsilon,\lambda_3,0.01)$-\shortOurEnsemble confident.
                }
                \label{fig:relation-ensemble-robustness-params}
            \end{figure}

            In \Cref{cor:comparision-statistical-lower-bound-uniform}, we derive the concrete certified robustness for both ensembles and the single model under i.i.d. and uniform distribution assumption.
            In fact, from the corollary, we can directly compute the certified robust radius without sampling, as long as we assume the added noise $\rvepsilon$ is Gaussian.
            In \Cref{fig:relation-ensemble-robustness-params}, we plot out such certified robust radius for the single model, the \shortWeightedEnsemble, and the \shortOurEnsemble.
        
            
            Concretely, in the figure, we assume that the true class confidence score for each base model is i.i.d. and \emph{uniformly distributed} in $[a,\,b]$.
            The \weightedEnsemble is $(\rvepsilon,\lambda_1,0.01)$-\shortWeightedEnsemble confident;
            the \ourEnsemble is $(\rvepsilon,\lambda_2,0.01)$-\shortOurEnsemble confident;
            and the single model is $(\rvepsilon,\lambda_3,0.01)$-\shortOurEnsemble confident.
            We guarantee that $\lambda_1 \le \lambda_3 \le \lambda_2$ to simulate the scenario that ensembles are based on the same set of base models to make a fair comparison.
            We directly apply the results from our analysis~(\Cref{thm:lower-bound-weighted-uniform}, \Cref{thm:lower-bound-our-uniform}, \Cref{prop:lower-bound-single-unif}) to get the statistical robustness for single model and both ensembles.
            Then, we leverage \Cref{lem:rand-smooth-vanilla-certify} to get the certified robust radius~(with $\sigma = 1.0, N=100000$ and failing probability $\alpha=0.001$ which are aligned with realistic setting).
            The $x$-axis is the number of base models $N$ and the $y$-axis is the certified robustness.
            We note that $N$ is not applicable to the single model, so we plot the single model's curve by a horizontal red dashed line.
            
            From the figure, we observe that when the number of base models $N$ becomes larger, both ensembles perform much better than the single model.
            We remark that when $N$ is small, the ensembles have $0$ certified robustness mainly because our theoretical bounds for ensembles are not tight enough with the small $N$.
            Furthermore, we observe that the \ourEnsemble gradually surpasses \weightedEnsemble when $N$ is large, which conforms to our \Cref{cor:comparision-statistical-lower-bound-uniform}.
            Note that the left sub-figure has smaller transferability $\lambda_1/\lambda_2$ and the right subfigure has larger transferability $\lambda_1/\lambda_2$, it again conforms to our \Cref{cor:comparision-statistical-lower-bound} and its following remarks in \Cref{adx-subsec:3-2} that in the left subfigure the \weightedEnsemble is relatively more robust than the \ourEnsemble.
            
    \subsubsection{Ensemble Comparison from Realistic Data}
    
        
        
        We study the correlation between transferability $\lambda_1/\lambda_2$ and whether \weightedEnsemble or \ourEnsemble is more certifiably robust using realistic data.
        
        By varying the hyper-parameters of \shortApproach, we find out a setting where over the same set of base models, \weightedEnsemble and \ourEnsemble have similar certified robustness, i.e., for about half of the test set samples, \shortWeightedEnsemble is more robust; for another half, \shortOurEnsemble is more robust.
        We collect $1,000$ test set samples in total.
        Then, for each test set sample, we compute the transferability $\lambda_1/\lambda_2$ and whether \shortWeightedEnsemble or \shortOurEnsemble has the higher certified robust radius.
        We remark that $\lambda_1$ and $\lambda_2$ are difficult to be practically estimated so we use the average confidence portion as the proxy:
        \begin{itemize}
            \item For \shortWeightedEnsemble, 
            $$
                \lambda_1 = \E_\rvepsilon \dfrac{\max_{y_j\in [C]: y_j \neq y_0} \sum_{i=1}^N w_i f_i(\vx_0+\rvepsilon)_{y_j}}{\sum_{i=1}^N w_i (1 - f_i(\vx_0 + \rvepsilon)_{y_0})}.
            $$
            
            \item For \shortOurEnsemble, 
            $$
                \lambda_2 = \E_\rvepsilon \max_{i\in [N]} \dfrac{\max_{y_j\in [C]: y_j\neq y_0} f_i(\vx_0+\rvepsilon)_{y_j}}{(1 - f_i(\vx_0 + \rvepsilon)_{y_0})}.
            $$
        \end{itemize}
        
\begin{figure}[!t]
        \centering
        \includegraphics[width=0.5\textwidth]{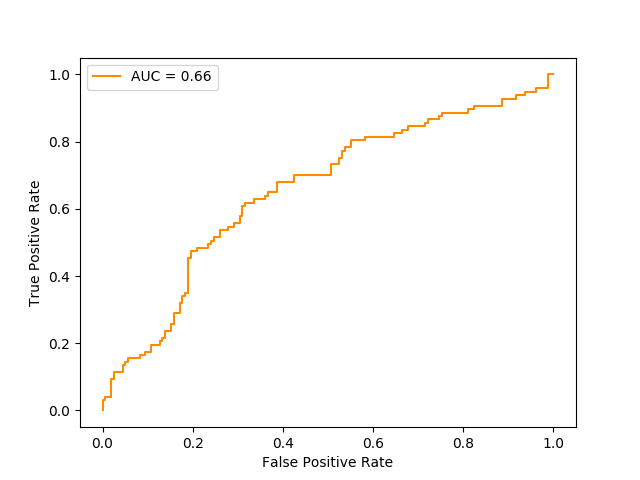}
        \caption{ROC curve of the $\1[\text{\shortOurEnsemble has higher certified robustness}]$ classification task with the threshold variable $X$.}
        \label{fig:roc}
        \end{figure}

        Now we study the correlation between 
        $$
        X := \lambda_1/\lambda_2 - \mathrm{RHS} \text{ of \Cref{eq:cor2-2}}  \text{ and } 
        Y := \1[\text{\shortOurEnsemble has higher certified robustness}].
        $$
        To do so, we draw the ROC curve where the threshold on $X$ does binary classification on $Y$.
        The curve and the AUC score is shown in \Cref{fig:roc}.
        From the ROC curve, we find that $X$ and $Y$ are apparently positively correlated since $\mathrm{AUC} = 0.66 > 0.5$, which again verifies \Cref{cor:comparision-statistical-lower-bound}.
        We remark that besides $X$, other factors such as non-symmetric or non-i.i.d. confidence score distribution may also play a role.
    
    \emph{Closing Remarks.} \quad
    The analysis in this appendix mainly shows two major findings theoretically and empirically:
    (1)~\shortOurEnsemble is more robust when the adversarial transferability is high; while \shortWeightedEnsemble is more robust when the adversarial transferability is low;
    (2)~If each $f_i(x_0 + \varepsilon)_{y_0}$ follows uniform distribution, when number of base models $N$ is sufficiently large, the \shortOurEnsemble is always more certifiably robust.
    Our analysis does have limitations:
    we assume the symmetric and i.i.d. distribution of $f_i(x_0 + \varepsilon)_{y_0}$ or even more strict uniform distribution to derive these findings.
    Though they model the real-world scenario in some extent as our realistic data results show, they are not perfect considering the transferability among base models and boundedness of confidence scores.
    We hope current analysis can open an angle of theoretical analysis of ensembles and leave a more general analysis as the future work.

\section{Analysis of Alternative Design of \shortApproach}
    \label{adxsec:alternative-design}
    
    In the main text, we design our \shortApproach based on \shortLHSloss
    \begin{equation}
    \begin{small}
        \Lgd(\vx_0)_{ij} = \big\|\nabla_\vx f_i^{y_0/y_i^{(2)}}(\vx_0) + \nabla_\vx f_j^{y_0/y_j^{(2)}}(\vx_0)\big\|_2
        \tag{\ref{eq:loss1}}
    \end{small}
    \end{equation}
    and \shortRHSloss
    \begin{equation}
    \begin{small}
        \Lcm(\vx_0)_{ij} = f_i^{y_i^{(2)}/y_0}(\vx_0) + f_j^{y_j^{(2)}/y_0} (\vx_0).
        \tag{\ref{eq:loss2}}
    \end{small}
    \end{equation}
    Following the convention, we apply Gaussian augmentation to train the models, i.e., replacing $\vx_0$ by $\vx_0 + \varepsilon$ where $\varepsilon\sim\gN(0,\sigma^2\mI_d)$ in \Cref{eq:loss1} and \Cref{eq:loss2}.
    We apply these two regularizers to every valid base model pair $(F_i,F_j)$, where the valid pair means both base models predict the ground truth label $y_0$: $F_i(\vx_0 + \varepsilon) = F_j(\vx_0 + \varepsilon)=y_0$.
    
    One may concern that in the worst case, there could be $O(N^2)$ valid pairs, i.e., $O(N^2)$ regularization terms in the training loss.
    However, we should notice that each base model $F_i$ only appears in $O(N)$ valid pairs.
    Therefore, when $N$ is large, we can optimize DRT by training iteratively, i.e., by regularizing each base model one by one to save the computational cost.
    
    An alternative design inspired from the theorems~(e.g., \Cref{thm:gradient-based-sufficient-necessary-cond-weight-ensemble}) is to use overall summation instead of pairwise summation, which directly correlates with $I_{y_i}$~(\Cref{eq:I}):
    \begin{equation}
    \begin{small}
        \Lgd'(\vx_0) = \Big\| \sum_{i=1}^N \nabla_\vx f_i^{y_0/y_i^{(2)}}(\vx_0) \Big\|_2,
        \label{eq:loss1-new}
    \end{small}
    \end{equation}
    \begin{equation}
    \begin{small}
        \Lcm'(\vx_0) = \sum_{i=1}^N f_i^{y_i^{(2)}/y_0}(\vx_0).
        \label{eq:loss2-new}
    \end{small}
    \end{equation}
    Although this design appears to be more aligned with the theorem and more efficient with $O(N)$ regularization terms, 
    it also requires all base models $F_i$ to have the same runner-up prediction $y_i^{(2)}$ as observed from both theorem and intuition~(otherwise diversified gradients and confidence margins are for different and independent labels that are meaningless to jointly optimize).
    It is less likely to have all base models having the same runner-up prediction than a pair of base models having the same runner-up prediction especially in the initial training phase.
    Therefore, this alternative design will cause fewer chances of meaningful optimization than the previous design and we use the previous design for our \shortApproach in practice.
    

\section{Experiment Details}


\label{adx-subsec:exp}

\textbf{Baselines.}
We consider the following state-of-the-art baselines for certified robustness: (1)~\textbf{Gaussian smoothing}~\citep{cohen2019certified} trains a smoothed classifier by applying  Gaussian augmentation.
2. MACER~\citep{zhai2019macer}: Adding the regularization term to maximize the certified radius $R = \frac{\sigma}{2}(p_A - p_B)$ on training instances. 
(2)~\textbf{SmoothAdv}~\citep{salman2019provably} combines adversarial training with Gaussian augmentation.
(3)~\textbf{MACER}~\citep{zhai2019macer} improves a single model's certified robustness by adding regularization terms to minimize the Negative Log Likelihood (NLL) between smoothed classifier's output $g_{F}(\vx)$ and label $y$, and maximize the certified radius 
$R = \frac{\sigma}{2}(\Phi^{-1}(g_{F}^\varepsilon(\vx)_y) - \Phi^{-1}(\max_{y' \neq y} g_{F}^\varepsilon(\vx)_{y'}))$, where
$\varepsilon\sim \gN(0,\sigma^2\mI_d)$ and
$g_F^\varepsilon$ is as defined in \Cref{def:smoothed-conf-g}.
(4)~\textbf{Stability}~\citep{li2019certified} maintains the stability of the smoothed classifier $g_F$ by minimizing the Rényi Divergence between $g_F(\vx)$ and $g_F(\vx+\rvepsilon)$ where $\rvepsilon \sim \gN(0,\sigma^2 \mI_d)$. (5)~\textbf{SWEEN}~\citep{liu2020enhancing} builds smoothed Weighted Ensemble~(WE), which is the only prior work computing certified robustness for ensemble to our knowledge. 

\textbf{Evaluation Metric.} We report the standard \emph{certified accuracy} under different $L_2$  radii $r$'s as our evaluation metric following \citet{cohen2019certified}, which is defined as the fraction of the test set samples that the smoothed classifier can certify the robustness within the $L_2$ ball of radius $r$.
Since the computation of the accurate value of this metric is intractable, we report the \emph{approximate certified test accuracy}~\citep{cohen2019certified} sampled through the Monte Carlo procedure. For each sample, the robustness certification holds  with probability at least $1 - \alpha$. Following the literature, we choose $\alpha=0.001$, $n_0 = 100$ for Monte Carlo sampling during prediction phase, and $n = 10^5$ for Monte Carlo sampling during certification phase. 
On MNIST and CIFAR-10 we evaluated every $10$-th image in the test set, for $1,000$ images total.
On ImageNet we evaluated every $100$-th image in the validation set, for $500$ images total.
This evaluation protocol is the same as prior work~\citep{cohen2019certified, salman2019provably}.


  \subsection{MNIST}
  \label{adx-subsec:mnist}
  \textbf{Baseline Configuration.} 
  Following the literature~\citep{salman2019provably,jeong2020consistency,zhai2019macer}, 
  in each batch, each training sample is Gaussian augmented twice~(augmenting more times yields negligible difference as \citet{salman2019provably} show).
  We choose Gaussian smoothing variance $\sigma \in \{0.25,0.5,1.0\}$ for training and evaluation for all methods.
  For SmoothAdv, we consider the attack to be $10$-step $L_2$ PGD attack with perturbation scale $\delta = 1.0$ without pretraining and unlabelled data augmentation. 
We reproduced results similar to their paper by using their open-sourced code\footnote{\textcolor{blue}{\url{https://github.com/Hadisalman/smoothing-adversarial/}}}.

\textbf{Training Details.}
First, we use LeNet architecture and train each base model for $90$ epochs. For the training optimizer, we use the SGD-momentum with the initial learning rate $\alpha = 0.01$. The learning rate is decayed for every $30$ epochs with decay ratio $\gamma = 0.1$ and the batch size equals to $256$. Then, we apply DRT to finetune our model with small learning rate $\alpha$  for another $90$ epochs. 
We explore different DRT hyper-parameters $\rho_1, \rho_2$ together with the initial learning rate $\alpha$, and report the best certified accuracy on each radius $r$ among all the trained ensemble models. 

\begin{table}[!htbp]
\centering
\caption{Certified accuracy of DRT-$(\rho_1, \rho_2)$ under different radii $r$ on MNIST dataset. Smoothing parameter $\sigma=0.25$. The grey rows present the performance of the proposed \shortApproach approach. The brackets show the base models we use.}
\begin{tabular}{c|c|c|c|c|c|c}
\toprule
Radius $r$                               & $\rho_1$ & $\rho_2$ & 0.00                      & 0.25                      & 0.50                      & 0.75                     \\ \hline
Gaussian~\citep{cohen2019certified}                              & -      & -      & 99.1                      & 97.9                      & 96.6                      & 93.0                     \\
SmoothAdv~\citep{salman2019provably}                              & -      & -      & 99.1                      & 98.4                      & 97.0                      & 96.3                     \\
\hline
MME (Gaussian)                         & -      & -      & 99.2                      & 98.4                      & 96.8                      & 93.6                     \\ \hline
  \rowcolor{tabgray}& 0.1    & 0.2    & 99.4                      & 98.3                      & 97.5                      & 95.1                     \\
     \rowcolor{tabgray}                                    & 0.1    & 0.5    & 99.5                      & \textbf{98.6}                      & 97.1                      & 94.8                     \\
     \rowcolor{tabgray}                \multirow{-3}{*}{DRT + MME (Gaussian)}                    & 0.2    & 0.5    & \textbf{99.5}                      & 98.5                      & 97.4                      & 95.1                     \\ \hline
MME (SmoothAdv)                        & -      & -      & 99.2                      & 98.2                      & 97.3                      & 96.4                     \\ \hline
  \rowcolor{tabgray} & 0.1    & 0.2    & \multicolumn{1}{l|}{99.1} & \multicolumn{1}{l|}{98.4} & \multicolumn{1}{l|}{97.5} & \multicolumn{1}{l}{96.4} \\
    \rowcolor{tabgray}                                     & 0.1    & 0.5    & \multicolumn{1}{l|}{99.1} & \multicolumn{1}{l|}{98.3} & \multicolumn{1}{l|}{\textbf{97.6}} & \multicolumn{1}{l}{\textbf{96.7}} \\
     \rowcolor{tabgray}                   \multirow{-3}{*}{DRT + MME (SmoothAdv)}                 & 0.2    & 0.5    & \multicolumn{1}{l|}{99.1} & \multicolumn{1}{l|}{98.4} & \multicolumn{1}{l|}{97.5} & \multicolumn{1}{l}{96.6} \\ \hline
WE (Gaussian)                          & -      & -      & 99.2                      & 98.4                      & 96.9                      & 93.7                     \\ \hline
   \rowcolor{tabgray}  & 0.1    & 0.2    & 99.5                      & 98.4                      & 97.3                      & 95.1                     \\
      \rowcolor{tabgray}                                   & 0.1    & 0.5    & \textbf{99.5}                      & \textbf{98.6}                      & 97.1                      & 94.9                     \\
   \rowcolor{tabgray}                    \multirow{-3}{*}{DRT + WE (Gaussian)}                  & 0.2    & 0.5    & \textbf{99.5}                      & 98.5                      & 97.3                      & 95.3                     \\ \hline
WE (SmoothAdv)                         & -      & -      & 99.2                      & 98.2                      & 97.4                      & 96.4                     \\ \hline
   \rowcolor{tabgray}& 0.1    & 0.2    & 99.1                      & 98.4                      & 97.5                      & 96.5                     \\
   \rowcolor{tabgray}                                      & 0.1    & 0.5    & 99.1                      & 98.2                      & \textbf{97.6}                      & 96.6                     \\
      \rowcolor{tabgray}                     \multirow{-3}{*}{DRT + WE (SmoothAdv)}               & 0.2    & 0.5    & 99.0                      & 98.4                      & 97.5                      & \textbf{96.7}                     \\ \bottomrule
\end{tabular}
\label{tab:bigtablemnist0.25}
\end{table}


\begin{table}[!t]
\centering
\caption{Certified accuracy of DRT-$(\rho_1, \rho_2)$ under different radii $r$ on MNIST dataset. Smoothing parameter $\sigma=0.50$. The grey rows present the performance of the proposed \shortApproach approach. The brackets show the base models we use.}
\scalebox{0.65}{
\begin{tabular}{c|c|c|c|c|c|c|c|c|c|c}
\toprule
Radius $r$                           & $\rho_1$              & $\rho_2$ & 0.00          & 0.25          & 0.50          & 0.75          & 1.00          & 1.25          & 1.50          & 1.75          \\ \hline
Gaussian~\citep{cohen2019certified}                             & -                     & -        & 99.0          & 97.7          & 96.4          & 94.7          & 90.0          & 83.0          & 68.2          & 43.5          \\
SmoothAdv~\citep{salman2019provably}                            & -                     & -        & 98.6          & 98.0          & 97.0          & 95.4          & 93.0          & 87.7          & 80.2          & 66.3          \\ \hline
MME (Gaussian)                        & -                     & -        & 99.0          & 97.7          & 96.8          & 94.9          & 90.5          & 84.3          & 69.8          & 48.5          \\ \hline
 \rowcolor{tabgray}  &   & 2.0      & 99.1          & 98.4          & 97.2          & 95.2          & 92.6          & 86.5          & 74.3          & 54.1          \\
    \rowcolor{tabgray}                                  &     \multirow{-2}{*}{0.2}                  & 5.0      & 99.1          & \textbf{98.6} & 97.1          & 95.3          & 92.6          & 86.2          & 74.0          & 54.3          \\ \cline{2-11} 
     \rowcolor{tabgray}                                 & & 2.0      & \textbf{99.2} & 98.3          & \textbf{97.4} & \textbf{95.5} & 92.1          & 86.4          & 74.7          & 55.6          \\
       \rowcolor{tabgray}                               &                       & 5.0      & 99.0          & 98.2          & 97.3          & 95.1          & 91.6          & 84.8          & 73.7          & 52.4          \\
        \rowcolor{tabgray}                              &    \multirow{-3}{*}{0.5}                     & 10.0     & 99.1          & 98.1          & 97.1          & 95.0          & 91.8          & 85.7          & 73.3          & 51.4          \\ \cline{2-11} 
       \rowcolor{tabgray}                               & 1.0                   & 5.0      & 99.1          & 98.2          & 97.2          & 95.2          & 92.2          & 85.8          & 74.4          & 54.4          \\ \cline{2-11} 
        \rowcolor{tabgray}                              &  & 0.1      & 98.8          & 98.0          & 96.8          & 94.7          & 91.5          & 86.5          & 75.5          & 59.1          \\
        \rowcolor{tabgray}                              &                       & 0.2      & 98.9          & 98.1          & 96.9          & 95.1          & 92.1          & 85.8          & 76.1          & 56.4          \\
         \rowcolor{tabgray}                             &                       & 0.5      & 98.7          & 98.1          & 96.8          & 95.2          & 92.1          & 85.8          & 76.0          & 56.9          \\
          \rowcolor{tabgray}                            &                       & 2.5      & 99.0          & 98.3          & 97.0          & 95.1          & 92.4          & 85.8          & 75.7          & 57.0          \\
           \rowcolor{tabgray}                           &                       & 5.0      & 99.0          & 98.1          & 96.8          & 95.0          & 91.9          & 85.5          & 74.4          & 54.6          \\
           \rowcolor{tabgray}                           &   \multirow{-6}{*}{10.0}                    & 10.0     & 99.0          & 98.2          & 96.9          & 95.1          & 91.9          & 85.5          & 74.6          & 54.5          \\ \cline{2-11} 
          \rowcolor{tabgray}                            &  & 2.5      & 98.7          & 98.0          & 96.7          & 95.1          & 91.7          & 86.4          & 75.6          & 59.8          \\
          \rowcolor{tabgray}                            &                       & 5.0      & 98.5          & 97.7          & 96.5          & 94.9          & 91.9          & 85.9          & 76.1          & 59.3          \\
       \rowcolor{tabgray}         \multirow{-15}{*}{DRT + MME (Gaussian)}                      &   \multirow{-3}{*}{80.0}                    & 25.0     & 98.9          & 98.0          & 96.9          & 94.9          & 92.2          & 85.7          & 76.5          & 58.3          \\ \hline
MME (SmoothAdv)                       & -                     & -        & 98.6          & 98.0          & 97.0          & \textbf{95.5} & 93.2          & 88.1          & 80.6          & 67.8          \\ \hline
 \rowcolor{tabgray}  &   & 0.5      & 98.4          & 97.8          & 97.0          & \textbf{95.5} & 92.7          & 87.7          & 80.9          & 67.9          \\
 \rowcolor{tabgray}                                      &                       & 1.0      & 98.4          & 97.9          & 97.0          & \textbf{95.5} & 92.9          & 88.1          & 80.8          & 67.2          \\
     \rowcolor{tabgray}                                  &     \multirow{-3}{*}{0.1}                  & 5.0      & 98.5          & 98.2          & 97.0          & 95.4          & 93.1          & 88.4          & \textbf{81.2} & 68.3          \\ \cline{2-11} 
     \rowcolor{tabgray}                                  &   & 0.5      & 98.4          & 97.7          & 97.2          & 95.3          & 92.3          & 87.7          & 79.3          & 68.4          \\
     \rowcolor{tabgray}                                  &                       & 2.0      & 98.4          & 97.6          & 97.1          & 95.3          & 92.3          & 87.8          & 80.2          & 67.7          \\
     \rowcolor{tabgray}                                  &                       & 5.0      & 98.4          & 97.8          & 97.1          & 95.2          & 93.0          & 87.9          & 80.3          & 68.3          \\
     \rowcolor{tabgray}                                  &    \multirow{-4}{*}{0.2}                   & 10.0     & 98.4          & 97.8          & 97.1          & 95.3          & 92.9          & \textbf{88.5} & 81.0          & 67.6          \\ \cline{2-11} 
     \rowcolor{tabgray}                                  &   & 5.0      & 98.4          & 97.5          & 97.1          & 95.0          & 92.4          & 87.7          & 79.7          & 68.3          \\
     \rowcolor{tabgray}                                  &   \multirow{-2}{*}{0.3}                    & 10.0     & 98.5          & 97.7          & 97.0          & 95.2          & 92.6          & \textbf{88.5} & 81.1          & 68.1          \\ \cline{2-11} 
     \rowcolor{tabgray}                                  &   & 2.0      & 98.5          & 97.3          & 96.6          & 94.3          & 91.6          & 86.7          & 79.5          & \textbf{68.6} \\
     \rowcolor{tabgray}                                  &   \multirow{-2}{*}{0.5}                    & 5.0      & 98.4          & 97.5          & 96.9          & 94.6          & 92.0          & 87.5          & 80.1          & 67.8          \\ \cline{2-11} 
     \rowcolor{tabgray}                                  &   & 0.5      & 97.7          & 96.8          & 95.5          & 92.3          & 89.6          & 84.1          & 76.7          & 66.3          \\
     \rowcolor{tabgray}                                  &   \multirow{-2}{*}{1.0}                    & 1.0      & 97.9          & 96.6          & 95.7          & 92.6          & 89.7          & 84.6          & 77.5          & 66.2          \\ \cline{2-11} 
      \rowcolor{tabgray}                                 &  & 0.1      & 95.4          & 93.3          & 91.2          & 88.1          & 83.8          & 76.8          & 68.3          & 59.9          \\
        \rowcolor{tabgray}                               &                       & 0.2      & 95.5          & 93.7          & 90.9          & 87.7          & 82.0          & 75.7          & 68.7          & 59.6          \\
        \rowcolor{tabgray}                               &                       & 0.5      & 95.0          & 93.3          & 91.1          & 87.8          & 82.6          & 76.3          & 68.2          & 59.7          \\
        \rowcolor{tabgray}                               &                       & 2.5      & 94.6          & 92.9          & 90.1          & 86.3          & 81.6          & 76.0          & 69.6          & 62.5          \\
         \rowcolor{tabgray}                              &                       & 5.0      & 94.3          & 93.1          & 90.0          & 86.1          & 81.9          & 76.3          & 70.0          & 63.6          \\
         \rowcolor{tabgray}                              &    \multirow{-6}{*}{10.0}                   & 10.0     & 94.9          & 93.4          & 91.3          & 87.3          & 83.2          & 78.2          & 71.8          & 65.9          \\ \cline{2-11} 
         \rowcolor{tabgray}                              &  & 2.5      & 87.7          & 84.4          & 79.9          & 75.0          & 70.5          & 65.5          & 58.9          & 50.5          \\
         \rowcolor{tabgray}             \multirow{-21}{*}{DRT + MME (SmoothAdv)}                 &  \multirow{-2}{*}{80.0}                     & 5.0      & 88.5          & 85.1          & 81.0          & 76.8          & 71.4          & 67.4          & 60.6          & 52.1          \\ \hline
WE (Gaussian)                         & -                     & -        & 99.0          & 97.8          & 96.8          & 94.9          & 90.6          & 84.5          & 70.4          & 48.2          \\ \hline
 \rowcolor{tabgray}   &   & 2.0      & \textbf{99.2} & 98.4          & 97.2          & 95.2          & 92.5          & 86.2          & 74.3          & 53.5          \\
     \rowcolor{tabgray}                                 &    \multirow{-2}{*}{0.2}                   & 5.0      & 99.1          & \textbf{98.6} & 97.1          & 95.3          & 92.6          & 86.4          & 74.2          & 54.4          \\ \cline{2-11} 
       \rowcolor{tabgray}                               &   & 2.0      & \textbf{99.2} & 98.3          & \textbf{97.4} & 95.6          & 92.1          & 86.5          & 74.7          & 55.3          \\
        \rowcolor{tabgray}                              &                       & 5.0      & 99.0          & 98.1          & \textbf{97.4} & 95.1          & 91.4          & 84.8          & 73.7          & 52.5          \\
         \rowcolor{tabgray}                             &      \multirow{-3}{*}{0.5}                 & 10.0     & 99.1          & 98.2          & 97.1          & 95.1          & 91.7          & 85.4          & 73.5          & 51.0          \\ \cline{2-11} 
         \rowcolor{tabgray}                             & 1.0                   & 5.0      & 99.1          & 98.2          & 97.2          & 95.2          & 92.2          & 85.9          & 75.1          & 55.3          \\ \cline{2-11} 
           \rowcolor{tabgray}                           &  & 0.1      & 98.8          & 98.0          & 96.8          & 94.8          & 91.6          & 86.7          & 76.3          & 59.0          \\
           \rowcolor{tabgray}                           &                       & 0.2      & 98.8          & 98.1          & 97.0          & 95.0          & 92.1          & 86.0          & 75.7          & 56.8          \\
          \rowcolor{tabgray}                            &                       & 0.5      & 98.8          & 98.1          & 96.9          & 95.2          & 92.2          & 86.0          & 76.2          & 57.0          \\
            \rowcolor{tabgray}                          &                       & 2.5      & 98.9          & 98.3          & 97.0          & 95.1          & 92.4          & 85.9          & 76.2          & 56.3          \\
             \rowcolor{tabgray}                         &                       & 5.0      & 99.0          & 98.1          & 96.9          & 95.0          & 91.8          & 85.5          & 74.5          & 55.0          \\
             \rowcolor{tabgray}                         &      \multirow{-6}{*}{10.0}                 & 10.0     & 99.0          & 98.1          & 96.9          & 95.1          & 91.9          & 85.7          & 74.3          & 54.4          \\ \cline{2-11} 
              \rowcolor{tabgray}                        &  & 2.5      & 98.7          & 97.9          & 96.7          & 95.1          & 91.8          & 86.2          & 75.5          & 60.1          \\
               \rowcolor{tabgray}                       &                       & 5.0      & 98.4          & 97.8          & 96.8          & 95.0          & 91.9          & 86.2          & 75.6          & 60.2          \\
              \rowcolor{tabgray}    \multirow{-15}{*}{DRT + WE (Gaussian)}                    &       \multirow{-3}{*}{80.0}                & 25.0     & 99.0          & 98.1          & 96.9          & 94.9          & 92.1          & 85.9          & 76.7          & 58.4          \\ \hline
WE (SmoothAdv)                        & -                     & -        & 98.7          & 98.0          & 97.0          & \textbf{95.5} & \textbf{93.4} & 88.2          & 81.1          & 67.9          \\ \hline
 \rowcolor{tabgray} &   & 0.5      & 98.4          & 97.8          & 97.0          & \textbf{95.5} & 92.7          & 87.8          & 80.6          & 68.1          \\
    \rowcolor{tabgray}                                 &                       & 1.0      & 98.5          & 97.9          & 97.0          & \textbf{95.5} & 93.1          & 88.0          & \textbf{81.2} & 67.7          \\
      \rowcolor{tabgray}                               &     \multirow{-3}{*}{0.1}                  & 5.0      & 98.5          & 98.2          & 97.0          & 95.4          & 93.3          & \textbf{88.5} & 81.4          & \textbf{68.6} \\ \cline{2-11} 
         \rowcolor{tabgray}                            &   & 0.5      & 98.4          & 97.7          & 97.2          & 95.4          & 92.3          & 87.6          & 79.7          & 68.0          \\
         \rowcolor{tabgray}                            &                       & 2.0      & 98.4          & 97.6          & 97.1          & 95.3          & 92.3          & 87.8          & 80.6          & 68.1          \\
         \rowcolor{tabgray}                            &                       & 5.0      & 98.4          & 97.9          & 97.1          & 95.1          & 93.0          & 88.2          & 80.4          & 69.1          \\
          \rowcolor{tabgray}                           &    \multirow{-4}{*}{0.2}                   & 10.0     & 98.3          & 97.8          & 97.1          & 95.3          & 92.9          & 88.4          & 80.7          & 68.1          \\ \cline{2-11} 
         \rowcolor{tabgray}                            &   & 5.0      & 98.4          & 97.5          & 97.1          & 95.0          & 92.4          & 87.9          & 79.9          & 69.3          \\
          \rowcolor{tabgray}                           &   \multirow{-2}{*}{0.3}                    & 10.0     & 98.4          & 97.7          & 97.0          & 95.2          & 92.6          & 88.4          & 81.1          & 68.2          \\ \cline{2-11} 
         \rowcolor{tabgray}                            &  & 2.0      & 98.4          & 97.3          & 96.6          & 94.3          & 91.8          & 86.7          & 79.6          & 68.1          \\
         \rowcolor{tabgray}                            &     \multirow{-2}{*}{0.5}                   & 5.0      & 98.4          & 97.5          & 96.9          & 94.7          & 92.0          & 87.7          & 79.7          & 67.7          \\ \cline{2-11} 
           \rowcolor{tabgray}                          &   & 0.5      & 97.8          & 96.8          & 95.4          & 92.3          & 89.7          & 84.1          & 77.0          & 65.9          \\
           \rowcolor{tabgray}                          &     \multirow{-2}{*}{1.0}                  & 1.0      & 97.9          & 96.6          & 95.6          & 92.7          & 89.8          & 84.4          & 77.4          & 66.2          \\ \cline{2-11} 
           \rowcolor{tabgray}                          &  & 0.1      & 95.3          & 93.5          & 91.2          & 88.7          & 83.8          & 76.8          & 68.9          & 60.1          \\
             \rowcolor{tabgray}                        &                       & 0.2      & 95.4          & 93.8          & 90.9          & 88.1          & 83.2          & 76.6          & 69.1          & 59.9          \\
           \rowcolor{tabgray}                          &                       & 0.5      & 95.1          & 93.5          & 90.9          & 87.7          & 83.6          & 76.6          & 69.1          & 59.8          \\
           \rowcolor{tabgray}                          &                       & 2.5      & 94.8          & 93.0          & 90.5          & 86.8          & 82.1          & 75.1          & 69.1          & 62.0          \\
           \rowcolor{tabgray}                          &                       & 5.0      & 94.4          & 93.3          & 90.1          & 86.6          & 82.0          & 75.8          & 70.0          & 63.2          \\
            \rowcolor{tabgray}                         &     \multirow{-6}{*}{10.0}                  & 10.0     & 94.7          & 93.3          & 90.5          & 86.8          & 82.5          & 77.2          & 71.8          & 65.6          \\ \cline{2-11} 
             \rowcolor{tabgray}                        &  & 2.5      & 87.8          & 83.1          & 78.5          & 74.0          & 67.7          & 62.3          & 54.9          & 47.0          \\
           \rowcolor{tabgray}   \multirow{-21}{*}{DRT + WE (SmoothAdv)}                       &   \multirow{-2}{*}{80.0}                    & 5.0      & 88.4          & 84.2          & 79.9          & 75.3          & 69.3          & 63.7          & 56.5          & 48.7          \\ \bottomrule
\end{tabular}}
\label{tab:bigtablemnist0.50}
\end{table}

\begin{table}[!t]
\centering
\caption{Certified accuracy of DRT-$(\rho_1, \rho_2)$ under different radii $r$ on MNIST dataset. Smoothing parameter $\sigma=1.00$. The grey rows present the performance of the proposed \shortApproach approach. The brackets show the base models we use.}
\scalebox{0.73}{
\begin{tabular}{c|c|c|c|c|c|c|c|c|c|c|c|c|c}
\toprule
Radius $r$                             & $\rho_1$             & $\rho_2$ & 0.00 & 0.25 & 0.50 & 0.75 & 1.00 & 1.25 & 1.50 & 1.75 & 2.00          & 2.25          & 2.50          \\ \hline
Gaussian~\citep{cohen2019certified}                               & -                    & -        & \bf 96.5 & 94.3 & 91.1 & 87.0 & 80.2 & 71.8 & 60.1 & 46.6 & 33.0          & 20.5          & 11.5          \\
SmoothAdv~\citep{salman2019provably}                              & -                    & -        & 95.3 & 93.5 & 89.3 & 85.6 & 80.4 & 72.8 & 63.9 & 54.6 & 43.2          & 34.3          & 24.0          \\ \hline
MME (Gaussian)                         & -                    & -        & 96.4 & 94.8 & \bf 91.3 & \bf 87.7 & \bf 80.8 & \bf 73.5 & 61.0 & 48.8 & 34.7          & 23.4          & 12.7          \\ \hline
 \rowcolor{tabgray}  &  & 2.0      & 96.0 & 93.9 & 90.1 & 86.3 & 80.7 & 73.2 & 63.0 & 52.0 & 38.9          & 26.9          & 15.6          \\
     \rowcolor{tabgray}                                    &    \multirow{-2}{*}{0.5}                  & 5.0      & 95.8 & 94.1 & 90.0 & 86.6 & 80.4 & 72.9 & 62.4 & 51.3 & 40.0          & 27.8          & 16.5          \\ \cline{2-14} 
      \rowcolor{tabgray}                                   & 1.0                  & 5.0      & 95.3 & 93.1 & 89.7 & 85.8 & 80.0 & 72.7 & 62.9 & 52.0 & 39.8          & 28.5          & 17.6          \\ \cline{2-14} 
       \rowcolor{tabgray}                                  &  & 0.5      & 91.3 & 89.7 & 85.6 & 78.8 & 73.3 & 65.8 & 59.1 & 52.2 & 43.9          & 36.0          & 29.1          \\
       \rowcolor{tabgray}                                  &                      & 2.5      & 92.5 & 90.2 & 87.7 & 82.0 & 76.3 & 69.6 & 60.7 & 52.8 & 43.4          & 35.4          & 26.0          \\
      \rowcolor{tabgray}                 \multirow{-6}{*}{DRT + MME (Gaussian)}                   &    \multirow{-3}{*}{5.0}                  & 5.0      & 93.2 & 90.6 & 88.1 & 82.9 & 78.1 & 70.6 & 62.3 & 52.5 & 43.3          & 34.4          & 23.8          \\ \hline
MME (SmoothAdv)                        & -                    & -        & 95.4 & 93.4 & 89.3 & 86.1 & 80.7 & 73.1 & \bf 65.0 & 55.0 & 44.8          & 35.0          & 25.2          \\ \hline
 \rowcolor{tabgray}  &  & 2.0      & 94.1 & 91.9 & 88.6 & 84.5 & 79.4 & 72.4 & 63.4 & 54.0 & 45.0          & 36.6          & 27.3          \\
  \rowcolor{tabgray}                                       &      \multirow{-2}{*}{0.2}                & 5.0      & 94.1 & 91.6 & 88.9 & 84.4 & 79.3 & 72.3 & 63.2 & 54.2 & 46.1          & 36.9          & 28.5          \\ \cline{2-14} 
     \rowcolor{tabgray}                                    &  & 2.0      & 92.8 & 91.3 & 87.7 & 83.2 & 77.3 & 71.2 & 62.2 & 53.3 & 45.5          & 37.0          & 29.7          \\
     \rowcolor{tabgray}                                    &   \multirow{-2}{*}{0.5}                   & 5.0      & 92.5 & 91.2 & 88.0 & 83.5 & 78.5 & 71.2 & 62.2 & 53.8 & 45.2          & 37.7          & 29.2          \\ \cline{2-14} 
      \rowcolor{tabgray}                                   & 1.0                  & 5.0      & 92.1 & 90.0 & 86.3 & 81.3 & 76.2 & 69.4 & 61.1 & 54.0 & 46.4          & 38.6          & 31.1          \\ \cline{2-14} 
      \rowcolor{tabgray}                                   & & 1.0      & 89.3 & 86.5 & 82.2 & 76.5 & 70.5 & 62.8 & 54.6 & 48.5 & 41.4          & 35.2          & 29.2          \\
       \rowcolor{tabgray}                                  &    \multirow{-2}{*}{5.0}                   & 5.0      & 87.6 & 83.3 & 78.8 & 73.1 & 67.4 & 61.8 & 56.2 & 50.5 & 44.9          & 38.4          & 32.8          \\ \cline{2-14} 
       \rowcolor{tabgray}                     \multirow{-8}{*}{DRT + MME (SmoothAdv)}             & 10.0                 & 20.0     & 82.7 & 79.6 & 75.3 & 72.0 & 67.9 & 63.3 & 58.6 & 51.1 & 46.6          & \textbf{40.3} & \textbf{34.7} \\ \hline
WE (Gaussian)                          & -                    & -        & 96.3 & \bf 94.9 & \bf 91.3 & \bf 87.7 & 80.7 & \bf 73.5 & 61.1 & 49.0 & 35.2          & 23.7          & 12.9          \\ \hline
 \rowcolor{tabgray} &  & 2.0      & 95.9 & 93.9 & 90.2 & 86.3 & 80.7 & 73.2 & 63.2 & 51.9 & 38.6          & 27.0          & 15.5          \\
  \rowcolor{tabgray}                                     &    \multirow{-2}{*}{0.5}                  & 5.0      & 95.9 & 94.1 & 90.0 & 86.4 & 80.4 & 73.1 & 62.3 & 51.7 & 39.8          & 27.5          & 16.4          \\ \cline{2-14} 
    \rowcolor{tabgray}                                   & 1.0                  & 5.0      & 95.4 & 93.1 & 89.7 & 85.8 & 80.0 & 72.7 & 62.9 & 52.1 & 39.9          & 28.5          & 17.8          \\ \cline{2-14} 
  \rowcolor{tabgray}                                     &  & 0.5      & 91.3 & 89.8 & 85.9 & 79.0 & 73.4 & 65.5 & 59.2 & 52.2 & 43.9          & 35.4          & 28.8          \\
   \rowcolor{tabgray}                                    &                      & 2.5      & 92.4 & 90.2 & 87.8 & 81.7 & 76.2 & 69.5 & 60.5 & 52.5 & 43.5          & 35.8          & 26.8          \\
    \rowcolor{tabgray}               \multirow{-6}{*}{DRT + WE (Gaussian)}                     &    \multirow{-3}{*}{5.0}                  & 5.0      & 92.9 & 90.7 & 88.0 & 82.7 & 78.1 & 70.5 & 62.3 & 52.6 & 43.1          & 34.5          & 24.4          \\ \hline
WE (SmoothAdv)                         & -                    & -        & 95.2 & 93.4 & 89.4 & 86.2 & \bf 80.8 & 73.3 & 64.8 & \bf 55.1 & 44.7          & 35.2          & 24.9          \\ \hline
\rowcolor{tabgray}  &  & 2.0      & 94.2 & 91.9 & 88.6 & 84.5 & 79.6 & 72.5 & 63.7 & 53.9 & 44.9          & 36.4          & 27.3          \\
   \rowcolor{tabgray}                                    &     \multirow{-2}{*}{0.2}                 & 5.0      & 94.2 & 91.6 & 88.9 & 84.4 & 79.3 & 72.5 & 63.3 & 54.3 & 45.9          & 36.9          & 28.7          \\ \cline{2-14} 
   \rowcolor{tabgray}                                    & & 2.0      & 92.6 & 91.3 & 87.7 & 83.1 & 77.5 & 71.1 & 62.4 & 53.3 & 45.3          & 36.7          & 29.3          \\
    \rowcolor{tabgray}                                   &  \multirow{-2}{*}{0.5}                     & 5.0      & 92.5 & 91.2 & 88.0 & 83.4 & 78.5 & 71.1 & 62.3 & 53.7 & 45.3          & 37.8          & 29.5          \\ \cline{2-14} 
   \rowcolor{tabgray}                                    & 1.0                  & 5.0      & 92.1 & 90.0 & 86.4 & 81.4 & 76.3 & 69.7 & 61.1 & 54.0 & 46.4          & 38.4          & 31.0          \\ \cline{2-14} 
    \rowcolor{tabgray}                                   &  & 1.0      & 89.1 & 86.5 & 82.5 & 76.7 & 70.5 & 63.0 & 54.8 & 48.4 & 41.5          & 35.3          & 29.1          \\
   \rowcolor{tabgray}                                    &  \multirow{-2}{*}{5.0}                    & 5.0      & 87.9 & 83.4 & 78.8 & 73.0 & 67.5 & 61.6 & 56.2 & 50.4 & 44.8          & 38.5          & 32.7          \\ \cline{2-14} 
   \rowcolor{tabgray}                     \multirow{-8}{*}{DRT + WE (SmoothAdv)}               & 10.0                 & 20.0     & 82.0 & 79.1 & 75.2 & 71.8 & 67.6 & 63.4 & 58.6 & 51.2 & \textbf{46.7} & 40.2          & \textbf{34.7} \\ \bottomrule
\end{tabular}}
\label{tab:bigtablemnist1.00}
\end{table}

\begin{figure}[!ht]
    \begin{subfigure}{.33\textwidth}
    \includegraphics[width=\textwidth]{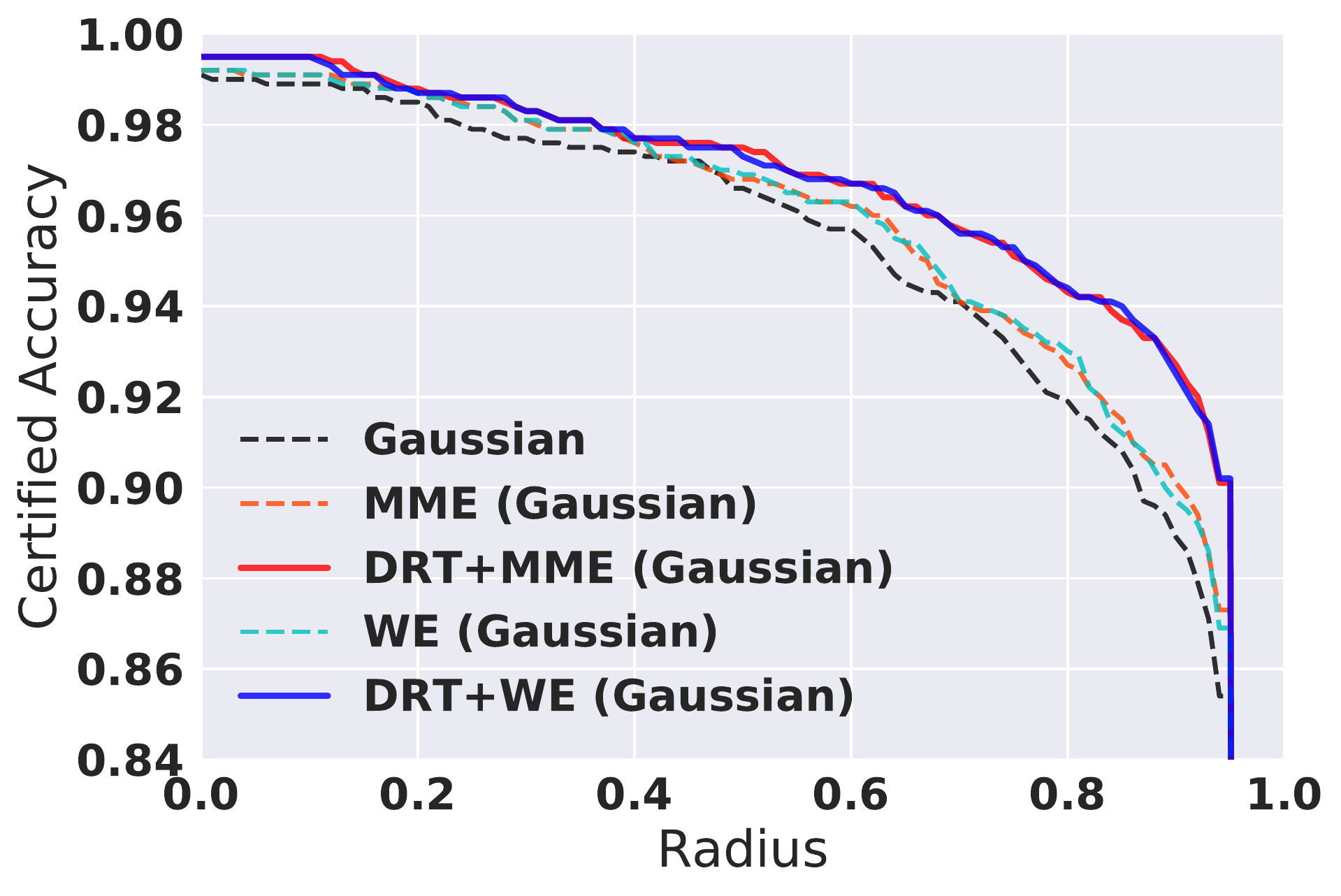}
    \caption{$\sigma=0.25$}
    \end{subfigure}
   \begin{subfigure}{.33\textwidth}
    \includegraphics[width=\textwidth]{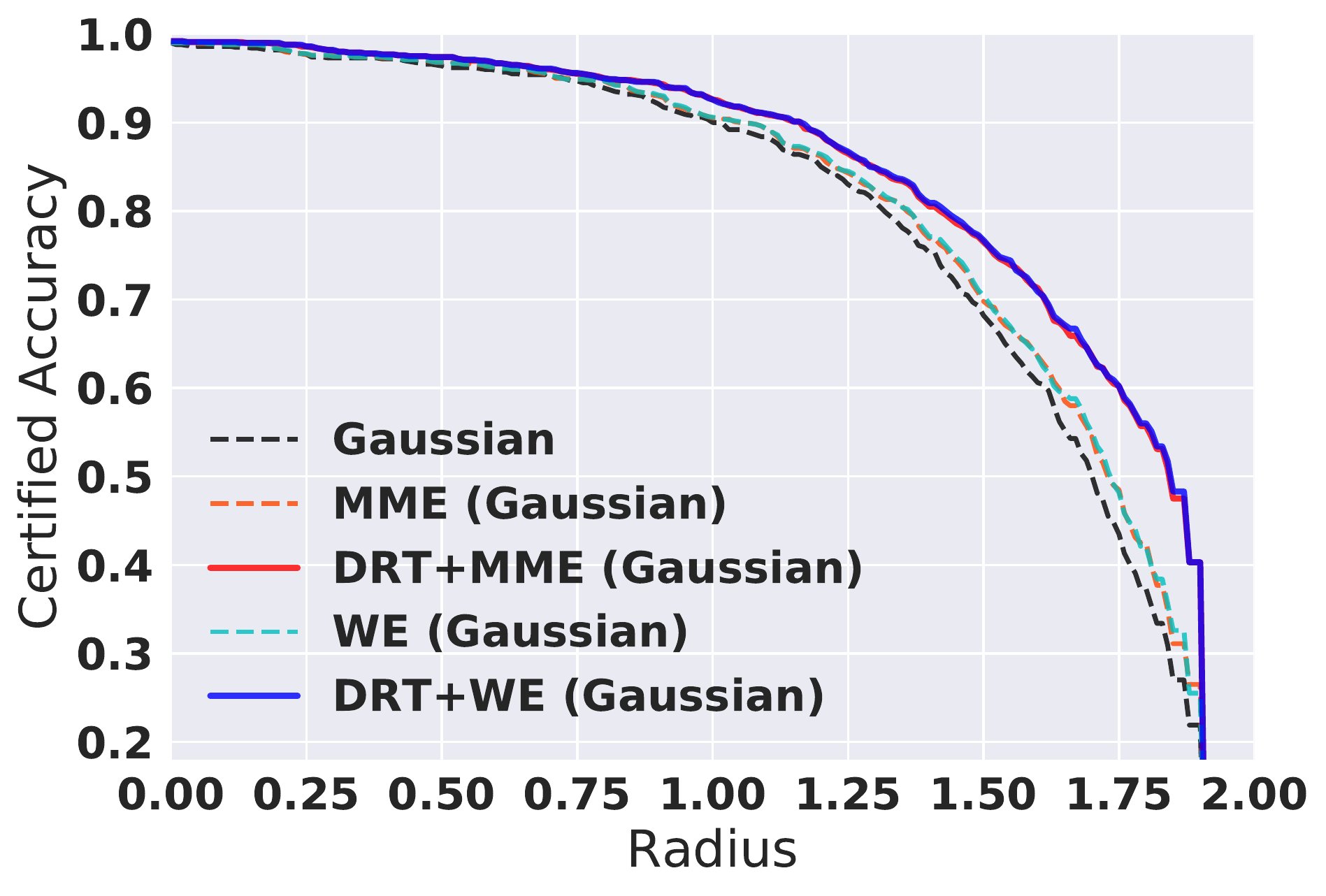}
    \caption{$\sigma=0.50$}
    \end{subfigure}
    \begin{subfigure}{.33\textwidth}
    \includegraphics[width=\textwidth]{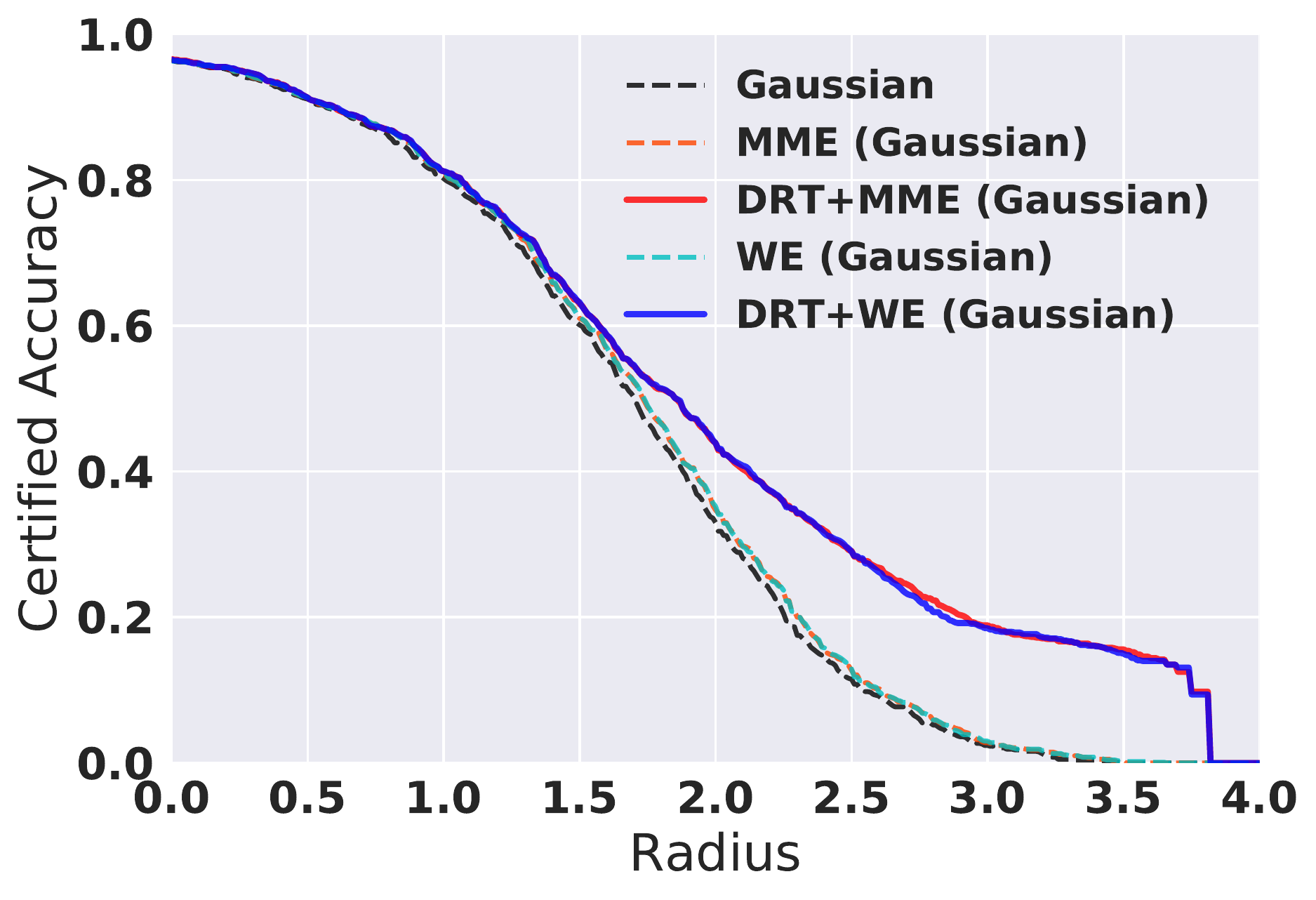}
    \caption{$\sigma=1.00$}
    \end{subfigure}
    \caption{Certified accuracy for ML ensembles with Gaussian smoothed base models, under smoothing parameter $\sigma \in \{0.25, 0.50, 1.00\}$ separately on MNIST.}
    \label{fig:mnist_plots}

    \begin{subfigure}{.33\textwidth}
    \includegraphics[width=\textwidth]{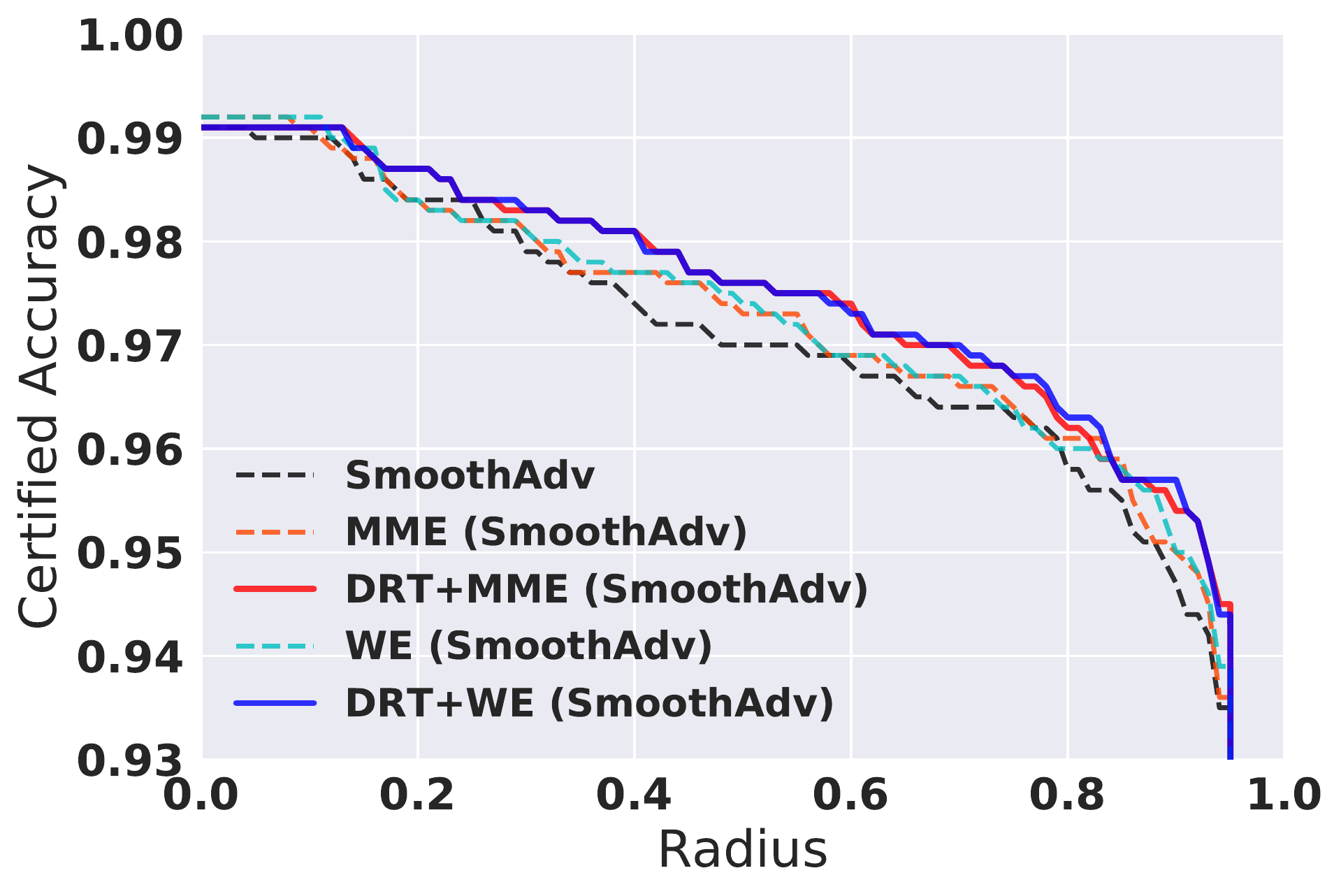}
    \caption{$\sigma=0.25$}
    \end{subfigure}
   \begin{subfigure}{.33\textwidth}
    \includegraphics[width=\textwidth]{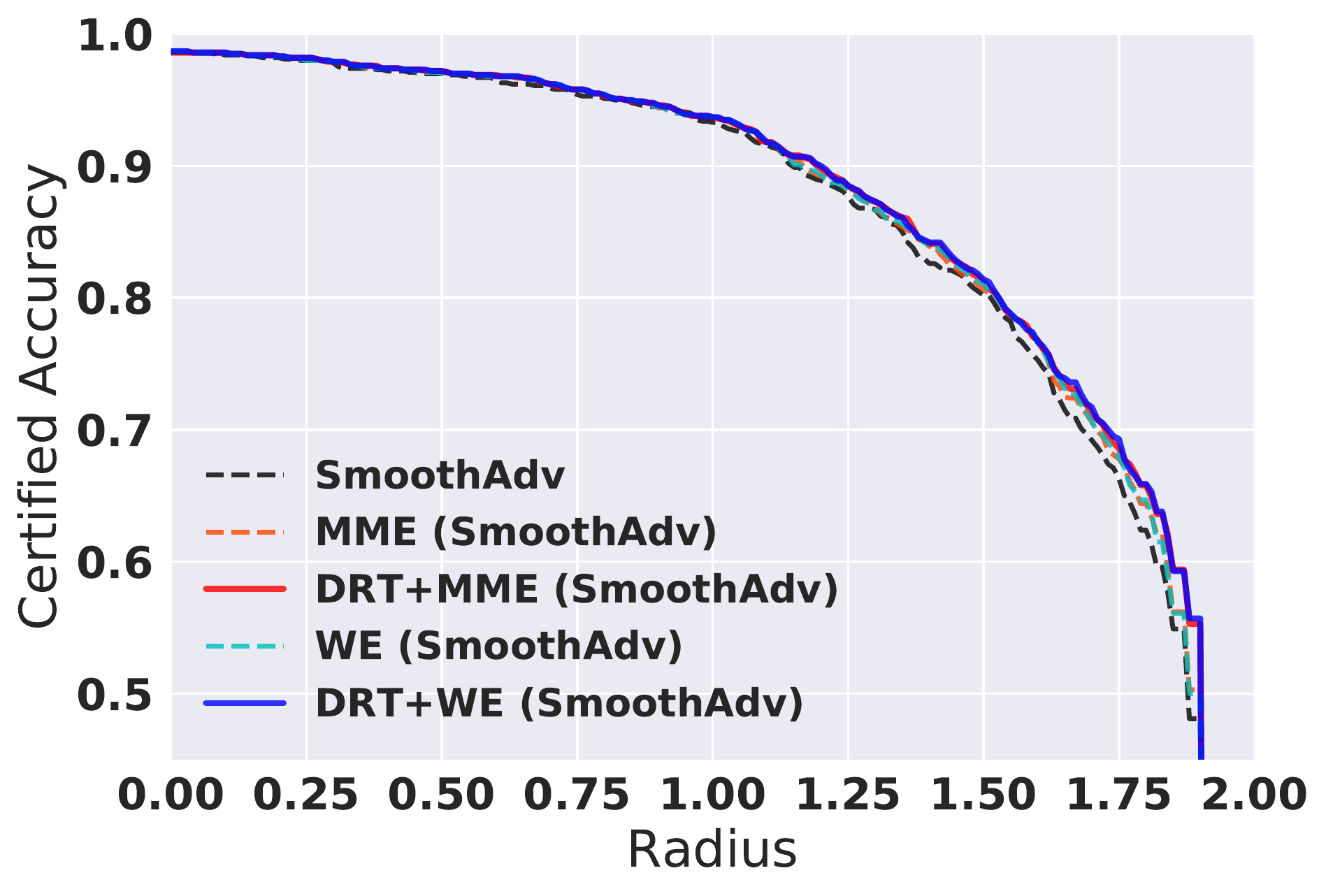}
    \caption{$\sigma=0.50$}
    \end{subfigure}
    \begin{subfigure}{.33\textwidth}
    \includegraphics[width=\textwidth]{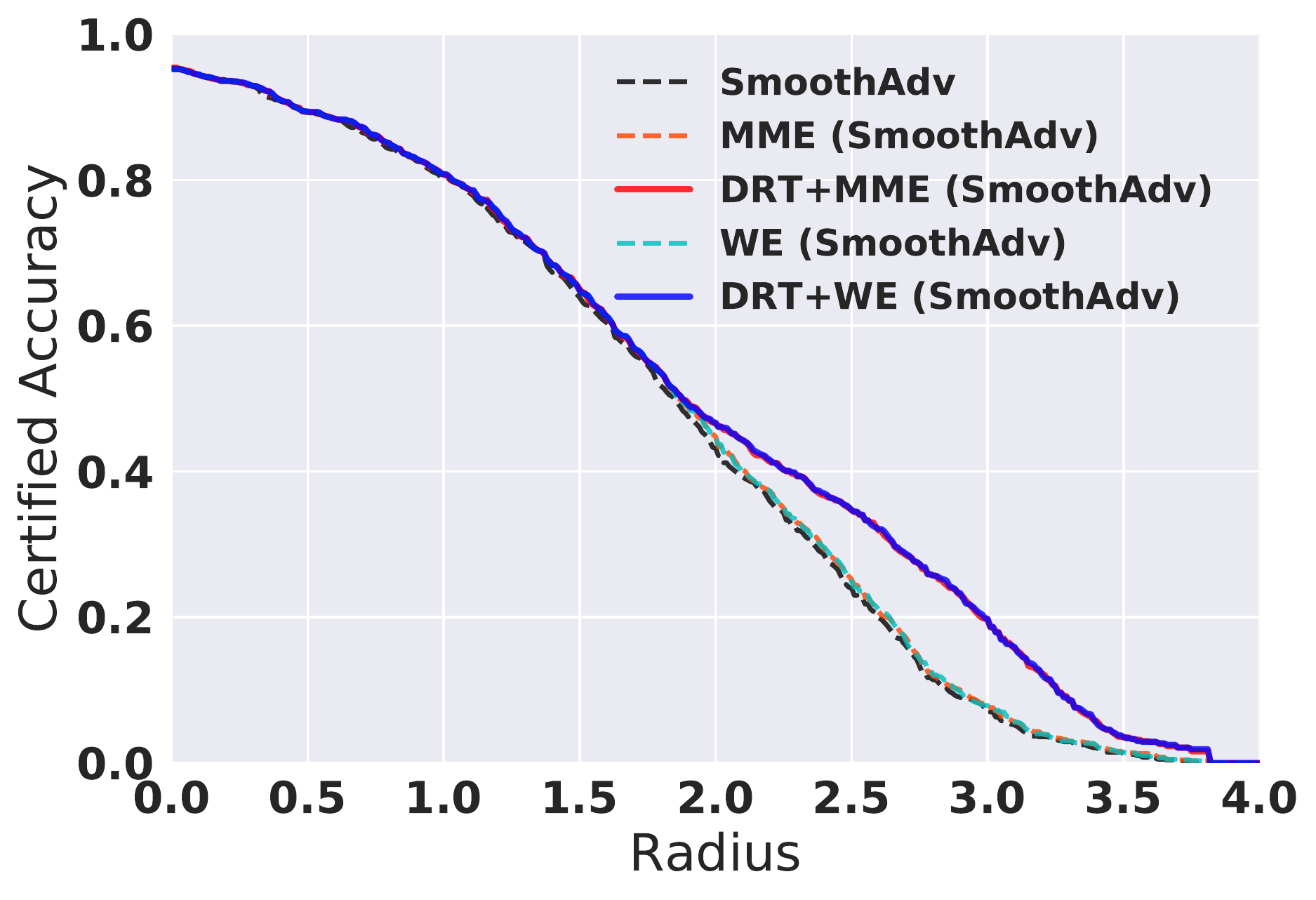}
    \caption{$\sigma=1.00$}
    \end{subfigure}
    \caption{Certified accuracy for ML ensembles with SmoothAdv base models, under smoothing parameter $\sigma \in \{0.25, 0.50, 1.00\}$ separately on MNIST. }
    \label{fig:mnist_advplots}
\end{figure}

\textbf{Trend of Certified Accuracy with Perturbation Radius.}
We visualize the trend of certified accuracy along with different perturbation radii on different smoothing parameters separately in \Cref{fig:mnist_plots} and \Cref{fig:mnist_advplots}. For each radius $r$, we present the best certified accuracy among all the trained models. We can notice that while simply applying MME or WE protocol could  slightly improve the certified accuracy, DRT could significantly boost the certified accuracy on different radii. 

\textbf{Average Certified Radius}. We report the Average Certified Radius (ACR)~\citep{zhai2019macer}: \text{ACR} = $\frac{1}{|\mathcal{S}_{\text{test}}|}\sum_{(x,y)\in\mathcal{S}_{\text{test}}} R(x,y)$, where $\mathcal{S}_{\text{test}}$ refers to the test set and $R(x, y)$ the certifed radius on testing sample $(x, y)$. We evaluate ACR of our DRT-trained ensemble trained with $\sigma \in \{0.25, 0.5, 1.0\}$ smoothing parameter and compare it with other baselines. Results are shown in \Cref{tab:acr-mnist}. 

We can clearly see that our DRT-trained ensemble could still achieve the highest ACR on all the smoothing parameter settings. Especially on $\sigma = 1.00$, our improvement is significant. 
\begin{table}[!htbp]
\centering
\caption{Average Certified Radius (ACR) of DRT-trained ensemble trained with different smoothing parameter $\sigma \in \{0.25, 0.50, 1.00\}$ on MNIST dataset, compared with other baselines. The grey rows present the performance of the proposed DRT approach. The brackets shows the base models we use.}
\begin{tabular}{l|c|c|c}
\toprule
Radius $r$                 & $\sigma=0.25$  & $\sigma=0.50$  & $\sigma=1.00$  \\ \hline
Gaussian~\citep{cohen2019certified}                   & 0.912          & 1.565          & 1.633          \\
SmoothAdv~\citep{salman2019provably}                  & 0.920          & 1.629          & 1.734          \\
MACER~\citep{zhai2019macer}                      & 0.918          & 1.583          & 1.520          \\ \hline
MME / WE (Gaussian)        & 0.915          & 1.585          & 1.669          \\
\rowcolor{tabgray} DRT + MME / WE (Gaussian)  & 0.923          & 1.637          & 1.745          \\
MME / WE (SmoothAdv)       & 0.926          & 1.678          & 1.765          \\
\rowcolor{tabgray} DRT + MME / WE (SmoothAdv) & \textbf{0.929} & \textbf{1.689} & \textbf{1.812} \\ \bottomrule
\end{tabular}
\label{tab:acr-mnist}
\end{table}

\begin{figure}[!htbp]
\centering
    \begin{subfigure}{.49\textwidth}
    \includegraphics[width=\textwidth]{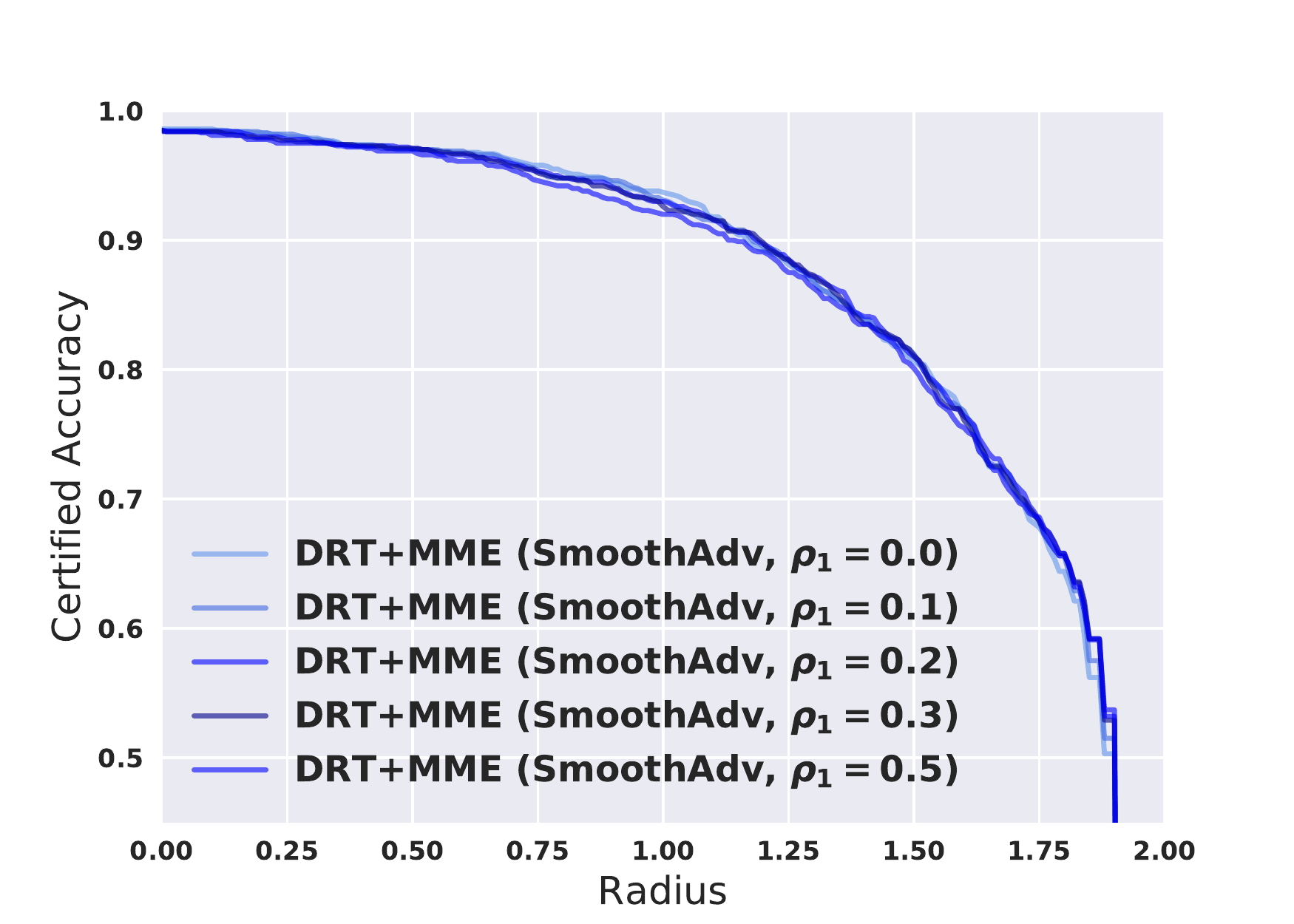}
    \caption{$\sigma=0.5$}
    \end{subfigure}
    \begin{subfigure}{.49\textwidth}
    \includegraphics[width=\textwidth]{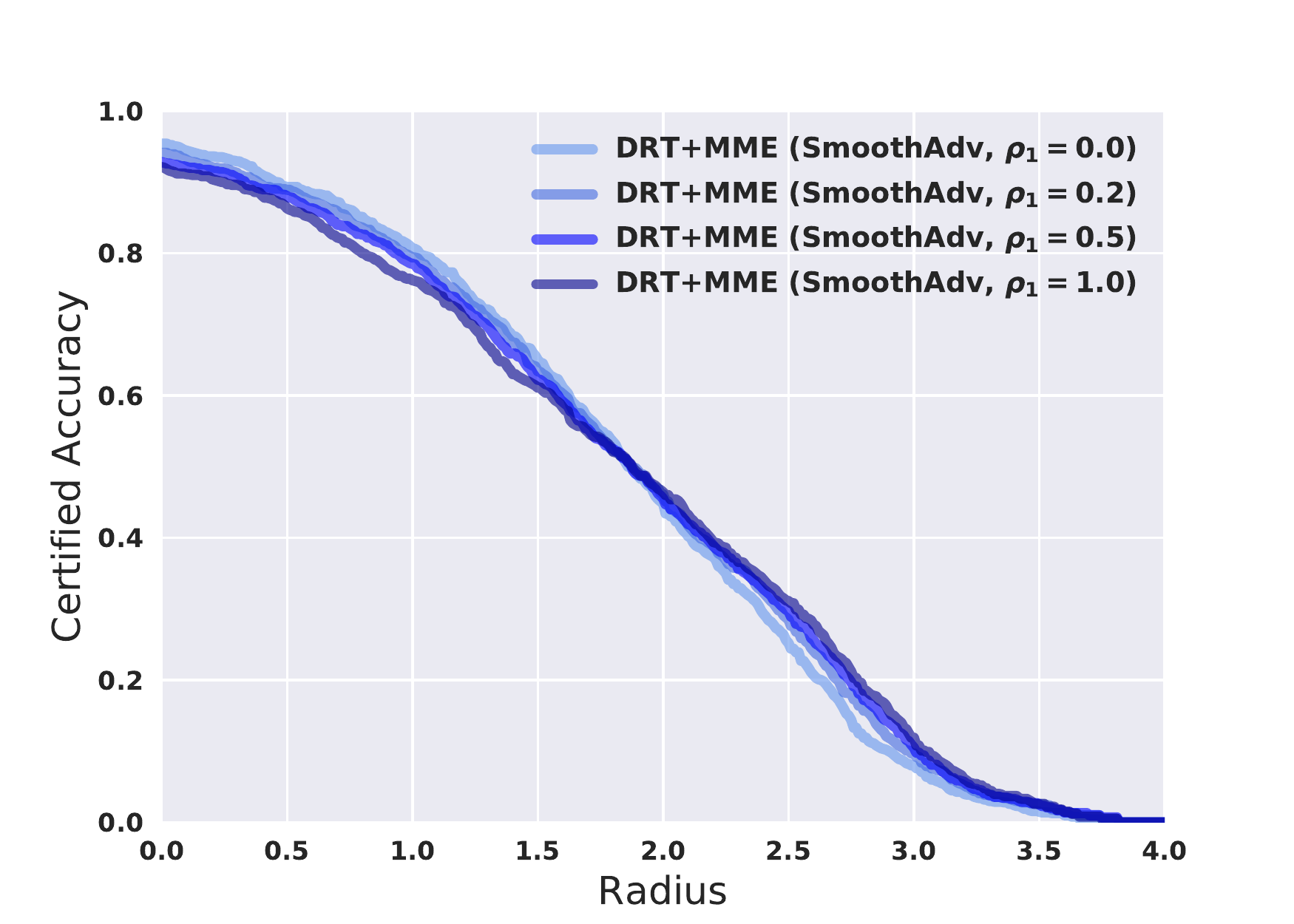}
    \caption{$\sigma=1.0$}
    \end{subfigure}
    \caption{\textbf{Effect of $\rho_1$}: Certified accuracy of DRT-based models with MME protocol trained by different GD Loss's weight $\rho_1$ on MNIST. Smoothing parameter $\sigma \in \{0.50, 1.00\}$. Training with large $\rho_1$ will lead to lower certified accuracy on small radii but higher certified accuracy on large radii.}
    \label{fig:mnist_rho}
\end{figure}

\textbf{Effects of $\rho_1$ and $\rho_2$.} 
We investigate the DRT hyper-parameters $\rho_1$ and $\rho_2$ corresponding to different smoothing parameter $\sigma \in \{0.25, 0.5, 1.0\}$. Here we put the detailed results for various hyper-parameter settings in \Cref{tab:bigtablemnist0.25,tab:bigtablemnist0.50,tab:bigtablemnist1.00} and bold the numbers with the highest certified accuracy on each radius $r$. From the experiments, we find that the GD loss's weight $\rho_1$ can have the major influence on the ensemble model's functionality: if we choose larger $\rho_1$, the model will achieve slightly 
lower certified accuracy on small radii, but higher certified accuracy on large radii. 
We also can not choose too large $\rho_1$ on small $\sigma$ cases (e.g., $\sigma = 0.25$). Otherwise, model's functionality will collapse. Here we show DRT-based models' \emph{certified accuracy} by applying different $\rho_1$ in \Cref{fig:mnist_rho}.

Alternatively, we find that the CM loss's weight $\rho_2$ can also have positive influence on model's performance: the larger $\rho_2$ we choose, the higher certified accuracy we could get. Choosing larger and larger $\rho_2$ does not harm model's functionality too much, but the improvement on certified accuracy will become more and more marginal. 

\textbf{Efficiency Analysis.} We regard the \emph{execution time per mini-batch} as our efficiency criterion. For MNIST with batch size equals to $256$, DRT with the Gaussian smoothing base model only requires \textcolor{red}{1.04s} to finish one mini-batch training to achieve the comparable results to the SmoothAdv method which requires \textcolor{red}{1.86s}. Moreover, DRT with the SmoothAdv base model requires \textcolor{red}{2.52s} per training batch but achieves much better results. The evaluation is on single NVIDIA GeForce GTX 1080 Ti GPU.  

\subsection{CIFAR-10}
\label{adx-subsec:cifar}
\textbf{Baseline Configuration.} 
  Following the literature~\citep{salman2019provably,jeong2020consistency,zhai2019macer}, 
  in each batch, each training sample is Gaussian augmented twice~(augmenting more times yields negligible difference as \citet{salman2019provably} show).
  We choose Gaussian smoothing variance $\sigma \in \{0.25,0.5,1.0\}$ for training and evaluation for all methods.
For SmoothAdv, we consider the attack to be $10$-step $L_2$ PGD attack with perturbation scale $\delta = 1.0$ without pretraining and unlabelled data augmentation. We also reproduced the similar results mentioned in baseline papers.

\textbf{Training Details.}
First, we use ResNet-110 architecture and train each base model for $150$ epochs. For the training optimizer, we use the SGD-momentum with the initial learning rate $\alpha = 0.1$. The learning rate is decayed for every $50$-epochs with decay ratio $\gamma = 0.1$. Then, we use DRT to finetune our model with small learning rate $\alpha$  for another $150$ epochs. We also explore different DRT hyper-parameters $\rho_1, \rho_2$ together with the initial learning rate $\alpha$, and report the best certified accuracy on each radius $r$ among all the trained ensemble models.

\begin{figure}[!t]
    \begin{subfigure}{.33\textwidth}
    \includegraphics[width=\textwidth]{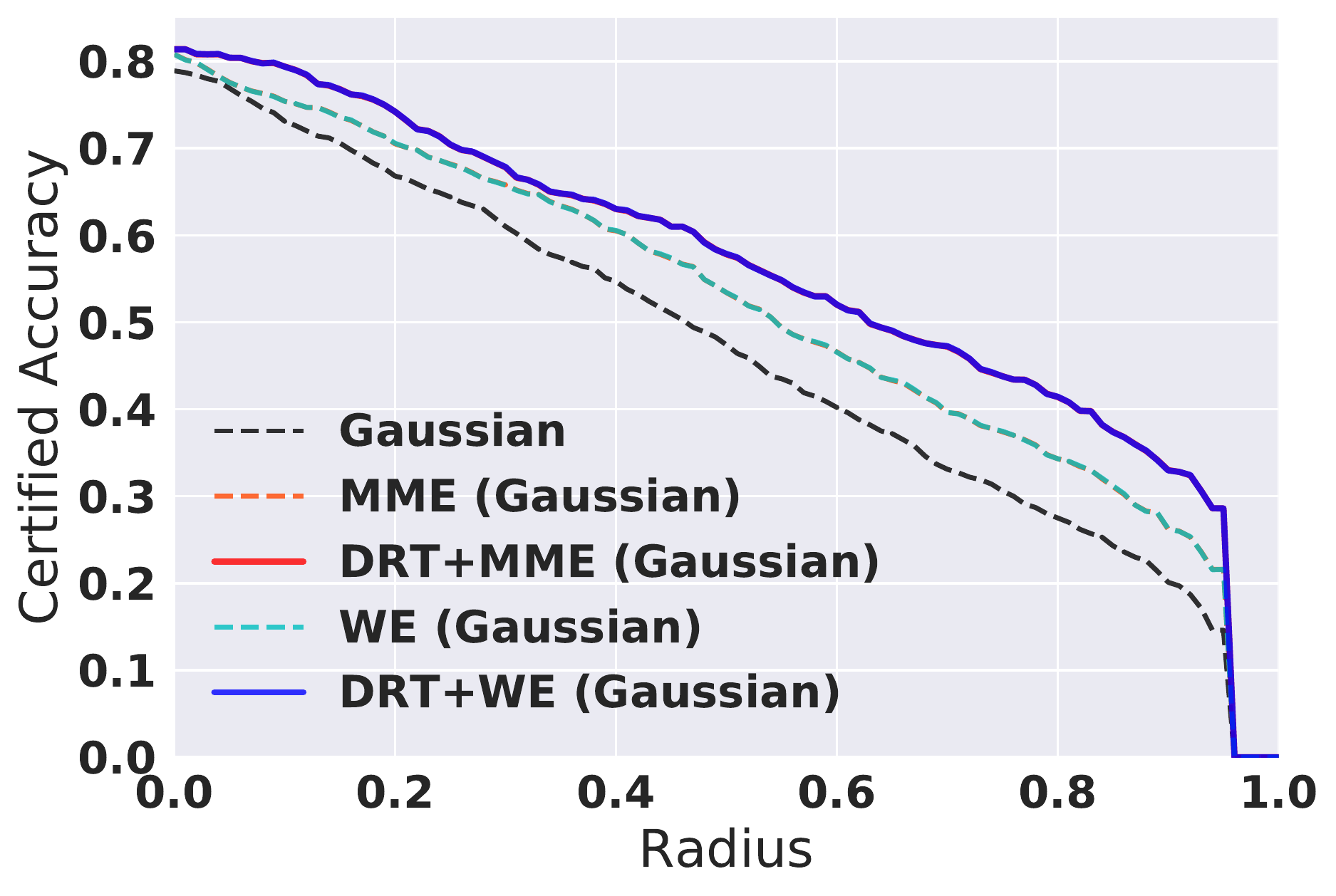}
    \caption{$\sigma=0.25$}
    \end{subfigure}
   \begin{subfigure}{.33\textwidth}
    \includegraphics[width=\textwidth]{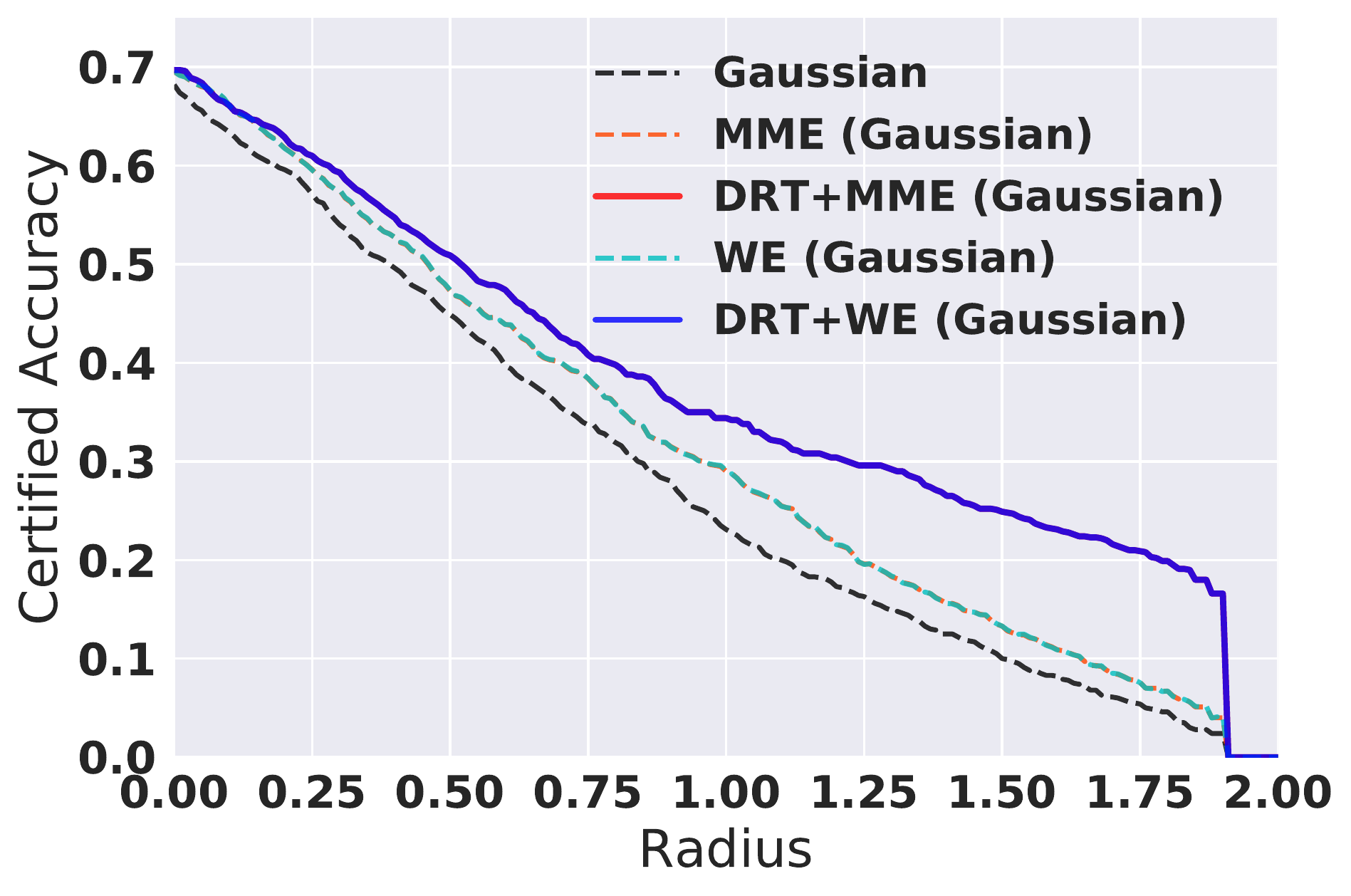}
    \caption{$\sigma=0.50$}
    \end{subfigure}
    \begin{subfigure}{.33\textwidth}
    \includegraphics[width=\textwidth]{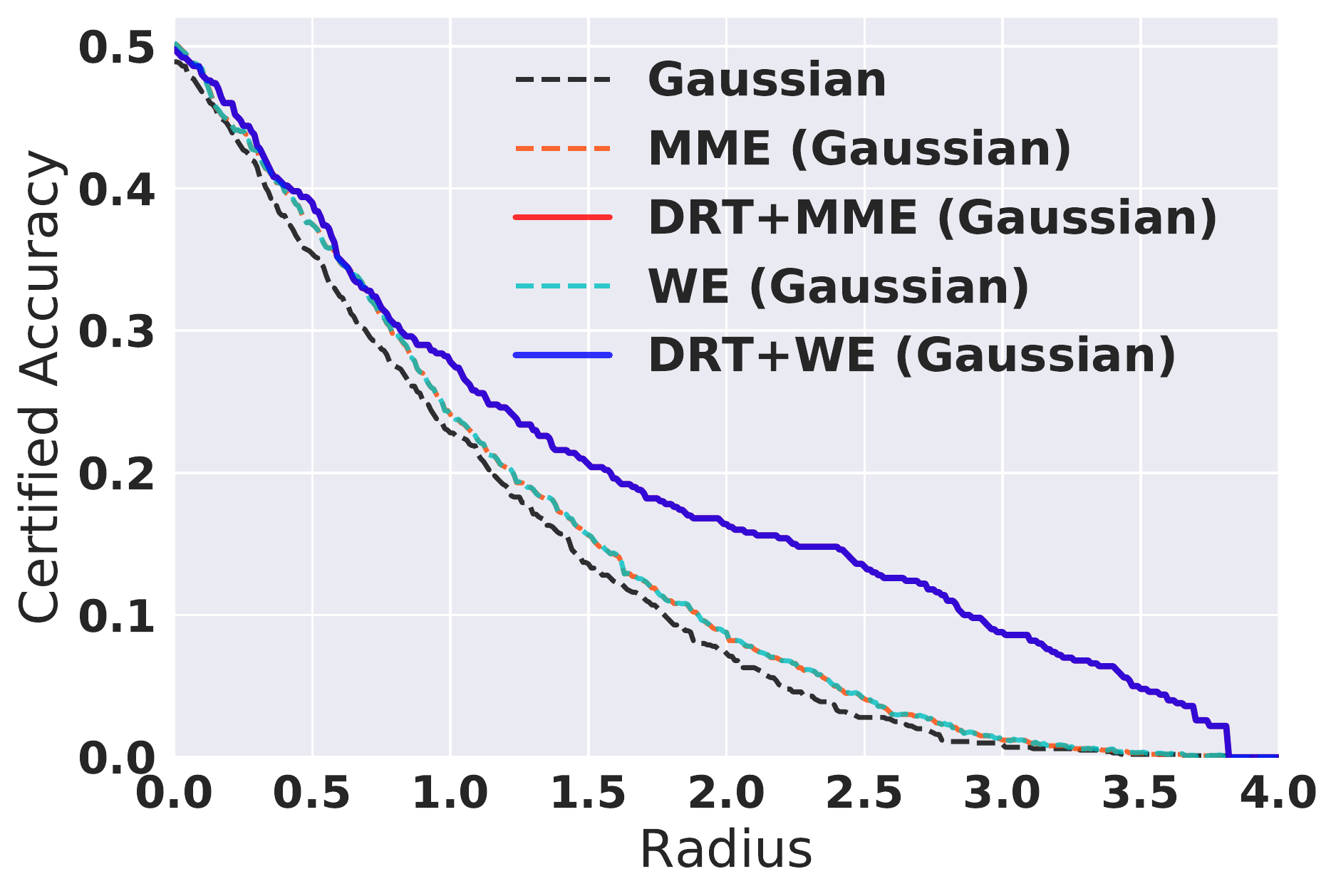}
    \caption{$\sigma=1.00$}
    \end{subfigure}
    \caption{Certified accuracy for ML ensembles with Gaussian smoothed base models, under smoothing parameter $\sigma \in \{0.25, 0.50, 1.00\}$ separately on CIFAR-10.}
    \label{fig:cifar_plots}

    \begin{subfigure}{.33\textwidth}
    \includegraphics[width=\textwidth]{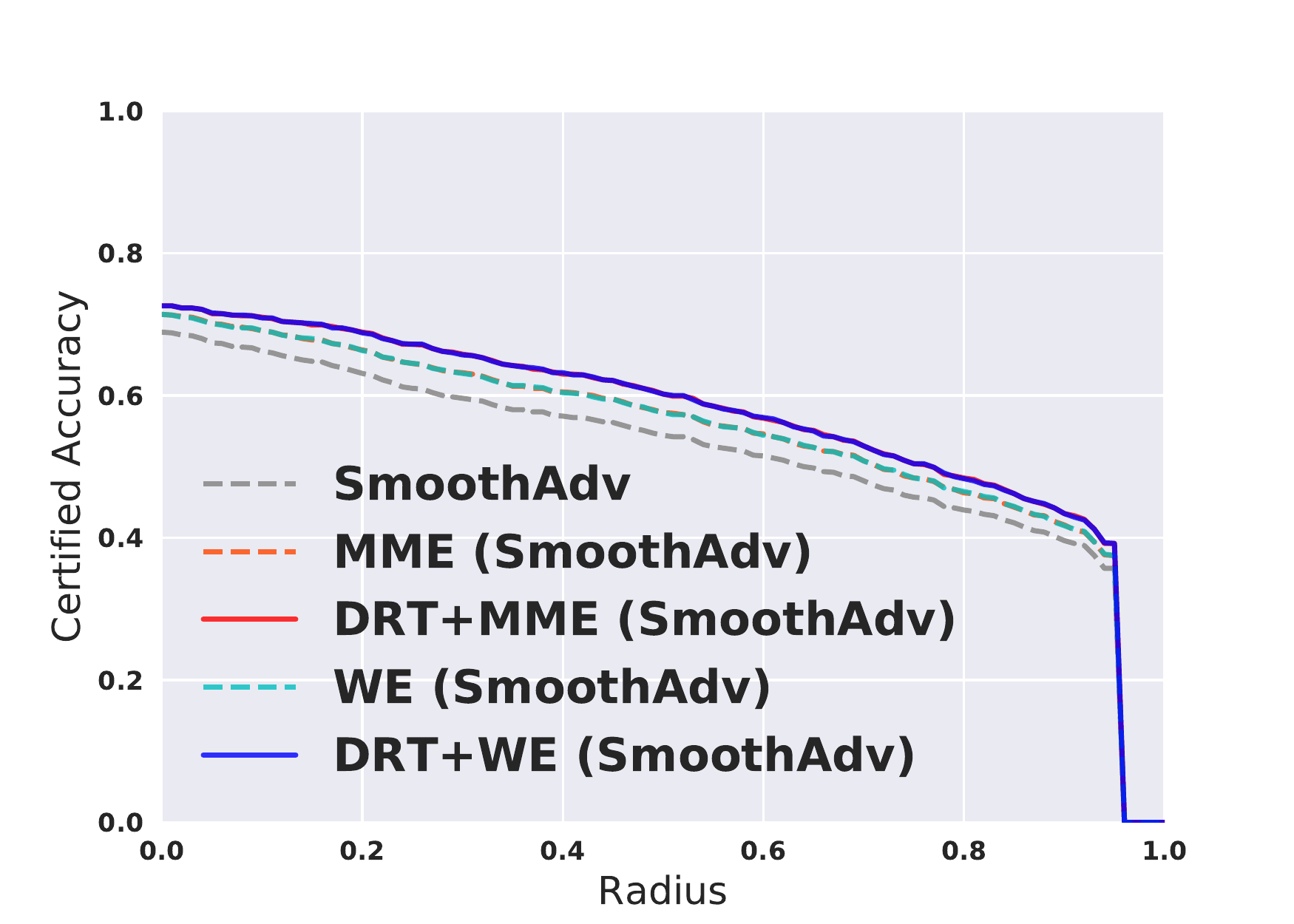}
    \caption{$\sigma=0.25$}
    \end{subfigure}
   \begin{subfigure}{.33\textwidth}
    \includegraphics[width=\textwidth]{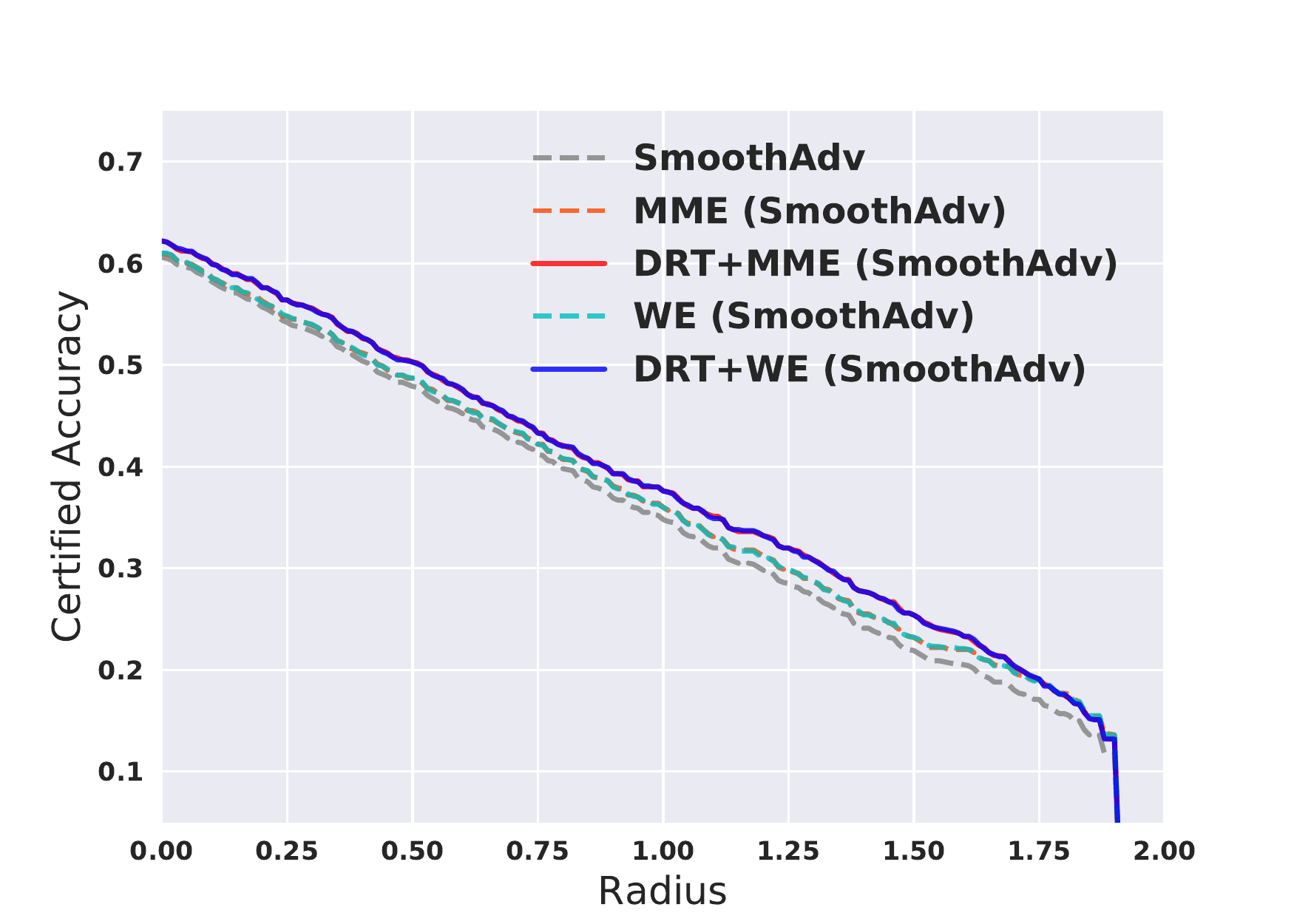}
    \caption{$\sigma=0.50$}
    \end{subfigure}
    \begin{subfigure}{.33\textwidth}
    \includegraphics[width=\textwidth]{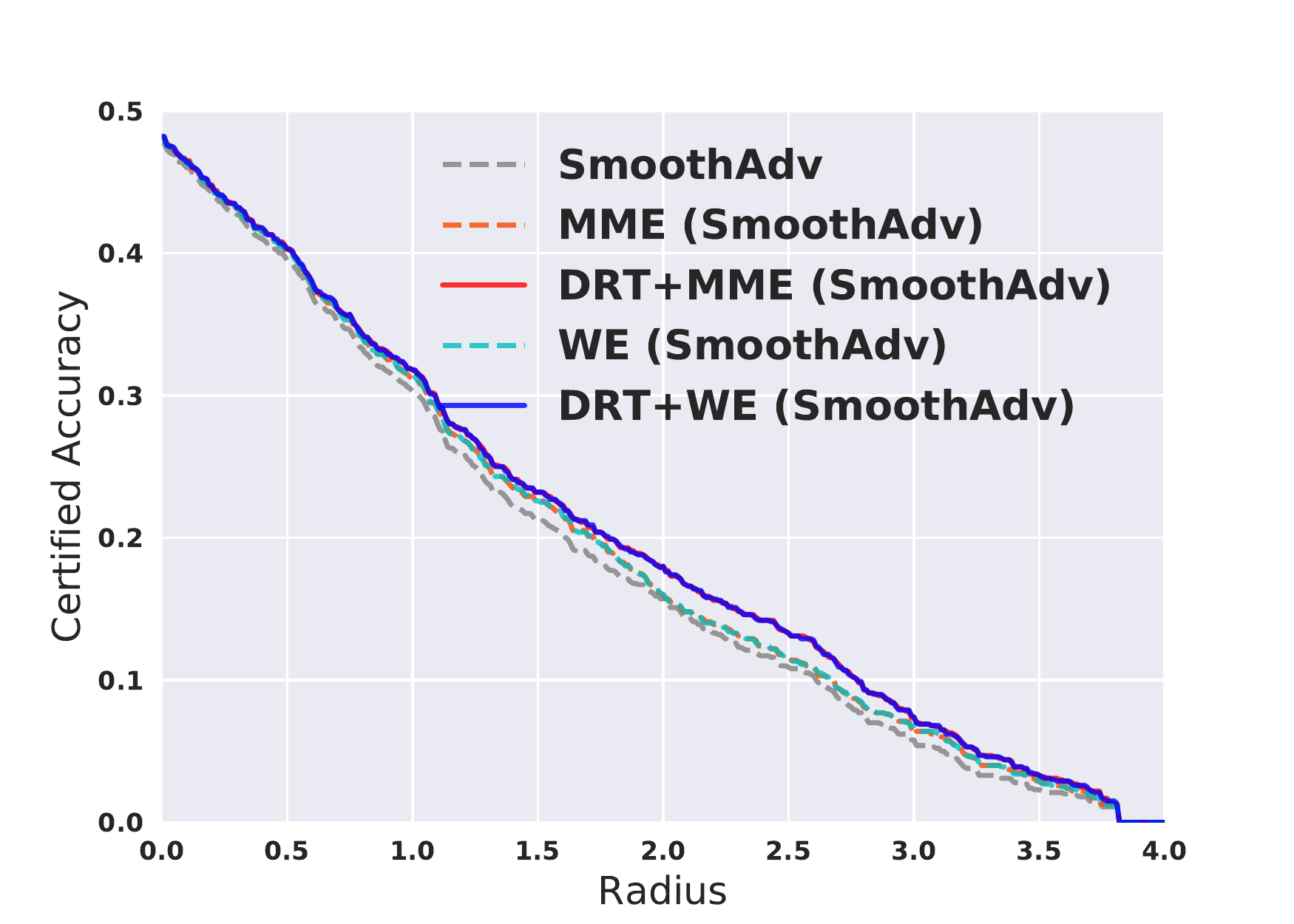}
    \caption{$\sigma=1.00$}
    \end{subfigure}
    \caption{Certified accuracy for ML ensembles with SmoothAdv base models, under smoothing parameter $\sigma \in \{0.25, 0.50, 1.00\}$ separately on CIFAR-10.}
    \label{fig:cifar_advplots}
\end{figure}

\textbf{Trend of Certified Accuracy with Perturbation Radius.}
We visualize the trend of certified accuracy along with different perturbation radii on different smoothing parameters separately in \Cref{fig:cifar_plots} and \Cref{fig:cifar_advplots}. For each radius $r$, we present the best certified accuracy among all the trained models. We can see the similar trends: Applying either MME or WE ensemble protocol will only give slight improvement while DRT can help make this improvement significant.

\begin{table}[!t]
\centering
\caption{Certified accuracy of DRT-$(\rho_1, \rho_2)$ under different radii $r$ on CIFAR-10 dataset. Smoothing parameter $\sigma=0.25$. The grey rows present the performance of the proposed \shortApproach approach. The brackets show the base models we use.}
\begin{tabular}{c|c|c|c|c|c|c}
\toprule
Radius $r$                             & $\rho_1$ & $\rho_2$ & $0.00$ & $0.25$ & $0.50$ & $0.75$ \\ \hline
Gaussian~\citep{cohen2019certified}                     & -        & -       & 78.9   & 64.4   & 47.4   & 30.6   \\
SmoothAdv~\citep{salman2019provably}                              & -        & -       & 68.9   & 61.0   & 54.4   & 45.7   \\ \hline
MME (Gaussian)                         & -        & -       & 80.8   & 68.2   & 53.4   & 37.4   \\ \hline
 \rowcolor{tabgray} & 0.1      & 0.5     & 81.4   & \textbf{70.4}   & 57.6   & 43.4   \\
     \rowcolor{tabgray}                                    & 0.2      & 0.5     & 78.8   & 69.2   & 57.8   & 43.8   \\
      \rowcolor{tabgray}                                   & 0.5      & 2.0     & 73.3   & 61.7   & 51.0   & 39.3   \\
     \rowcolor{tabgray}                            \multirow{-4}{*}{DRT + MME (Gaussian)}        & 0.5      & 5.0     & 66.2   & 57.1   & 46.2   & 34.4   \\ \hline
MME (SmoothAdv)                        & -        & -       & 71.4   & 64.5   & 57.6   & 48.4   \\ \hline
 \rowcolor{tabgray}  & 0.1      & 0.5     & 72.6   & 67.2   & \textbf{60.2}   & 50.3   \\
     \rowcolor{tabgray}                                    & 0.2      & 0.5     & 71.8   & 66.5   & 59.3   & 50.4   \\
       \rowcolor{tabgray}                                 \multirow{-3}{*}{DRT + MME (SmoothAdv)} & 0.5      & 0.5     & 68.2   & 64.3   & 58.2   & 48.9   \\ \hline
WE (Gaussian)                          & -        & -       & 80.7   & 68.3   & 53.6   & 37.5   \\ \hline
 \rowcolor{tabgray}    & 0.1      & 0.5     & \textbf{81.5}   & \textbf{70.4}   & 57.7   & 43.4   \\
    \rowcolor{tabgray}                                     & 0.2      & 0.5     & 78.8   & 69.3   & 57.9   & 44.0   \\
     \rowcolor{tabgray}                                    & 0.5      & 2.0     & 73.4   & 61.7   & 51.0   & 39.2   \\
   \rowcolor{tabgray}                                   \multirow{-4}{*}{DRT + WE (Gaussian)}   & 0.5      & 5.0     & 66.2   & 57.1   & 46.1   & 34.5   \\ \hline
WE (SmoothAdv)                         & -        & -       & 71.8   & 64.6   & 57.8   & 48.5   \\ \hline
 \rowcolor{tabgray}  & 0.1      & 0.5     & 72.6   & 67.0   & \textbf{60.2}   & 50.3   \\
     \rowcolor{tabgray}                                    & 0.2      & 0.5     & 71.9   & 66.5   & 59.4   & \textbf{50.5}   \\
     \rowcolor{tabgray}                                 \multirow{-3}{*}{DRT + WE (SmoothAdv)}    & 0.5      & 0.5     & 68.2   & 64.3   & 58.4   & 49.1   \\ \bottomrule
\end{tabular}
\label{tab:bigcifartable0.25}
\end{table}

\begin{table}[!t]
\centering
\caption{Certified accuracy of DRT-$(\rho_1, \rho_2)$ under different radii $r$ on CIFAR-10 dataset. Smoothing parameter $\sigma=0.50$. The grey rows present the performance of the proposed \shortApproach approach. The brackets show the base models we use.}
\scalebox{0.88}{
\begin{tabular}{c|c|c|c|c|c|c|c|c|c|c}
\toprule
Radius $r$                             & $\rho_1$              & $\rho_2$ & 0.00          & 0.25          & 0.50          & 0.75          & 1.00          & 1.25          & 1.50          & 1.75          \\ \hline
Gaussian~\citep{cohen2019certified}                               & -                     & -        & 68.2          & 57.1          & 44.9          & 33.7          & 23.1          & 16.3          & 10.0          & 5.4           \\
SmoothAdv~\citep{salman2019provably}                              & -                     & -        & 60.6          & 54.2          & 47.9          & 41.2          & 34.8          & 28.5          & 21.9          & 17.1          \\ \hline
MME (Gaussian)                         & -                     & -        & 69.5          & 59.6          & 47.3          & 38.4          & 29.0          & 19.6          & 13.3          & 7.6           \\ \hline
\rowcolor{tabgray} &   & 2.0      & \textbf{69.7} & 61.0          & \textbf{50.9} & 40.3          & 30.8          & 22.5          & 15.8          & 10.0          \\
     \rowcolor{tabgray}                                  &     \multirow{-2}{*}{0.2}                  & 5.0      & 68.0          & 59.9          & 50.0          & 40.8          & 30.1          & 22.1          & 15.2          & 9.6           \\ \cline{2-11} 
        \rowcolor{tabgray}                               &   & 2.0      & 67.8          & 58.5          & 49.0          & 39.9          & 31.6          & 23.4          & 16.1          & 10.2          \\
         \rowcolor{tabgray}                              &  \multirow{-2}{*}{0.5}                     & 5.0      & 65.5          & 58.4          & 49.0          & 40.1          & 31.2          & 23.6          & 16.5          & 10.2          \\ \cline{2-11} 
          \rowcolor{tabgray}                             &   & 2.0      & 64.5          & 55.8          & 47.5          & 39.4          & 31.1          & 23.6          & 14.8          & 9.3           \\
          \rowcolor{tabgray}                             &    \multirow{-2}{*}{1.0}                   & 5.0      & 62.2          & 54.1          & 46.5          & 38.8          & 29.7          & 22.8          & 16.6          & 11.0          \\ \cline{2-11} 
           \rowcolor{tabgray}                            & 1.5                   & 5.0      & 59.2          & 52.8          & 44.1          & 35.6          & 27.8          & 22.3          & 15.0          & 10.2          \\ \cline{2-11} 
           \rowcolor{tabgray}                            &   & 2.5      & 58.4          & 51.0          & 44.2          & 39.2          & 33.4          & 27.6          & 23.4          & 20.6          \\
           \rowcolor{tabgray}                            &    \multirow{-2}{*}{5.0}                   & 5.0      & 56.2          & 49.6          & 45.8          & 40.4          & 34.4          & 29.6          & 24.4          & 20.8          \\ \cline{2-11} 
            \rowcolor{tabgray}                           & & 2.5      & 52.0          & 46.8          & 42.0          & 36.2          & 32.4          & 27.8          & 23.4          & 19.7          \\
           \rowcolor{tabgray}                            &   \multirow{-2}{*}{10.0}                     & 5.0      & 51.2          & 47.5          & 42.5          & 38.1          & 33.7          & 28.9          & 24.9          & 20.9          \\ \cline{2-11} 
           \rowcolor{tabgray}                            & 15.0                  & 20.0     & 54.5          & 49.8          & 44.7          & 34.9          & 30.2          & 23.0          & 18.7          & 11.1          \\ \cline{2-11} 
            \rowcolor{tabgray}            \multirow{-13}{*}{DRT + MME (Gaussian)}               & 20.0                  & 30.0     & 52.2          & 46.2          & 40.2          & 34.4          & 29.4          & 22.6          & 17.8          & 12.8          \\ \hline
MME (SmoothAdv)                        & -                     & -        & 61.0          & 54.8          & 48.7          & 42.2          & 36.2          & 29.8          & 23.9          & 19.1          \\ \hline
\rowcolor{tabgray} & 0.2                   & 5.0      & 62.2          & 56.4          & 50.3          & 43.4          & 37.5          & 26.7          & 24.6          & 19.4          \\ \cline{2-11} 
         \rowcolor{tabgray}                              & 0.5                   & 5.0      & 61.9          & 56.2          & 50.3          & 43.5          & 37.6          & 31.8          & 24.8          & 19.6          \\ \cline{2-11} 
              \rowcolor{tabgray}                         & 1.0                   & 5.0      & 56.4          & 52.6          & 48.2          & \textbf{44.4} & 39.6          & 35.8          & \textbf{30.4} & 23.6          \\ \cline{2-11} 
               \rowcolor{tabgray}             \multirow{-4}{*}{DRT + MME (SmoothAdv)}           & 1.5                   & 5.0      & 56.0          & 50.8          & 47.2          & 44.2          & \textbf{39.8} & 35.0          & 29.4          & 24.0          \\ \hline
WE (Gaussian)                          & -                     & -        & 69.4          & 59.7          & 47.5          & 38.4          & 29.2          & 19.7          & 13.3          & 7.5           \\ \hline
  \rowcolor{tabgray} &   & 2.0      & \textbf{69.7} & \textbf{61.2} & 50.8          & 40.2          & 30.8          & 22.4          & 15.9          & 10.0          \\
                      \rowcolor{tabgray}                   &     \multirow{-2}{*}{0.2}                  & 5.0      & 68.0          & 59.9          & 50.1          & 40.8          & 30.1          & 22.1          & 15.4          & 9.7           \\ \cline{2-11} 
                      \rowcolor{tabgray}                   &   & 2.0      & 67.8          & 58.5          & 49.2          & 39.8          & 31.7          & 23.5          & 16.2          & 10.4          \\
                       \rowcolor{tabgray}                  &    \multirow{-2}{*}{0.5}                   & 5.0      & 65.5          & 58.4          & 49.1          & 40.3          & 31.3          & 24.2          & 16.4          & 10.3          \\ \cline{2-11} 
                        \rowcolor{tabgray}                 &   & 2.0      & 64.6          & 55.9          & 47.5          & 39.6          & 31.0          & 24.0          & 14.8          & 9.4           \\
                         \rowcolor{tabgray}                &      \multirow{-2}{*}{1.0}                 & 5.0      & 62.3          & 54.2          & 46.6          & 38.8          & 29.8          & 22.9          & 16.6          & 10.9          \\ \cline{2-11} 
                           \rowcolor{tabgray}              & 1.5                   & 5.0      & 59.2          & 52.8          & 44.2          & 35.8          & 27.8          & 22.4          & 15.0          & 10.3          \\ \cline{2-11} 
                            \rowcolor{tabgray}             &   & 2.5      & 58.4          & 51.1          & 44.2          & 39.2          & 33.3          & 27.8          & 23.2          & 20.6          \\
                             \rowcolor{tabgray}            &     \multirow{-2}{*}{5.0}                  & 5.0      & 56.2          & 49.7          & 45.8          & 40.3          & 34.2          & 29.6          & 24.5          & 20.8          \\ \cline{2-11} 
                             \rowcolor{tabgray}            &  & 2.5      & 52.0          & 46.9          & 42.0          & 36.4          & 32.5          & 27.8          & 23.5          & 19.7          \\
                               \rowcolor{tabgray}          &     \multirow{-2}{*}{10.0}                  & 5.0      & 51.2          & 47.6          & 42.4          & 38.1          & 33.6          & 28.9          & 24.9          & 20.8          \\ \cline{2-11} 
                              \rowcolor{tabgray}           & 15.0                  & 20.0     & 54.3          & 49.8          & 44.6          & 35.0          & 30.3          & 23.0          & 18.8          & 11.3          \\ \cline{2-11} 
          \rowcolor{tabgray}       \multirow{-13}{*}{DRT + WE (Gaussian)}                         & 20.0                  & 30.0     & 52.2          & 46.2          & 40.2          & 34.5          & 29.2          & 22.6          & 17.9          & 12.8          \\ \hline
WE (SmoothAdv)                         & -                     & -        & 61.1          & 54.8          & 48.8          & 42.3          & 36.2          & 29.6          & 24.2          & 19.0          \\ \hline
  \rowcolor{tabgray}  & 0.2                   & 5.0      & 62.2          & 56.3          & 50.3          & 43.4          & 37.5          & 26.9          & 24.7          & 19.3          \\ \cline{2-11} 
          \rowcolor{tabgray}           & 0.5                   & 5.0      & 61.9          & 56.2          & 50.2          & 43.4          & 37.9          & 31.8          & 25.0          & 19.6          \\ \cline{2-11} 
                   \rowcolor{tabgray}                      & 1.0                   & 5.0      & 56.4          & 52.6          & 48.2          & \textbf{44.4} & 39.5          & \textbf{36.0} & 30.3          & 23.6          \\ \cline{2-11} 
                                   \rowcolor{tabgray} 
\multirow{-4}{*}{DRT + WE (SmoothAdv)}                        & 1.5                   & 5.0      & 56.1          & 50.9          & 47.2          & 44.1          & \textbf{39.8} & 35.1          & 29.4          & \textbf{24.1} \\ \bottomrule
\end{tabular}}
\label{tab:bigcifartable0.50}
\end{table}

\begin{table}[!t]
\centering
\caption{DRT-$(\rho_1, \rho_2)$ model's certified accuracy under different radii $r$ on CIFAR-10 dataset. Smoothing parameter $\sigma=1.00$. The grey rows present the performance of the proposed \shortApproach approach. The brackets show the base models we use.}
\scalebox{0.83}{
\begin{tabular}{c|c|c|c|c|c|c|c|c|c|c|c}
\toprule
Radius $r$                             & $\rho_1$             & $\rho_2$ & 0.00 & 0.25 & 0.50 & 0.75 & 1.00 & 1.25 & 1.50 & 1.75 & 2.00 \\ \hline
Gaussian~\citep{cohen2019certified}                               & -                    & -        & 48.9 & 42.7 & 35.4 & 28.7 & 22.8 & 18.3 & 13.6 & 10.5 & 7.3  \\
SmoothAdv~\citep{salman2019provably}                              & -                    & -        & 47.8 & 43.3 & 39.5 & 34.6 & 30.3 & 25.0 & 21.2 & 18.2 & 15.7 \\ \hline
MME (Gaussian)                         & -                    & -        & 50.2 & 44.0 & 37.5 & 30.9 & 24.1 & 19.3 & 15.6 & 11.6 & 8.8  \\ \hline
      \rowcolor{tabgray}   & 0.5                  & 5.0      & 49.4 & 44.2 & 37.8 & 31.6 & 25.4 & 22.6 & 18.2 & 14.4 & 12.4 \\ \cline{2-12} 
             \rowcolor{tabgray}                                  & 1.0                  & 5.0      & 49.8 & \bf 44.4 & 39.0 & 31.6 & 25.6 & 22.6 & 18.2 & 15.0 & 12.0 \\ \cline{2-12} 
             \rowcolor{tabgray}                                  & 1.5                  & 5.0      & 48.0 & 42.4 & 36.4 & 30.4 & 26.2 & 22.0 & 18.4 & 15.4 & 12.8 \\ \cline{2-12} 
           \rowcolor{tabgray}                                    &  & 0.5      & 44.6 & 38.6 & 34.6 & 29.2 & 25.6 & 21.8 & 19.4 & 17.0 & 15.6 \\
           \rowcolor{tabgray}                                    &                      & 2.5      & 44.8 & 39.6 & 35.2 & 31.0 & 27.8 & 23.4 & 20.6 & 18.2 & 16.6 \\
            \rowcolor{tabgray}                                   &     \multirow{-3}{*}{5.0}                 & 10.0     & 45.4 & 40.4 & 36.8 & 30.4 & 26.0 & 21.8 & 19.0 & 15.8 & 13.6 \\ \cline{2-12} 
            \rowcolor{tabgray}                                   & 10.0                 & 20.0     & 44.4 & 40.8 & 36.2 & 31.2 & 27.4 & 21.2 & 18.8 & 17.2 & 13.6 \\ \cline{2-12} 
               \rowcolor{tabgray}                                & 15.0                 & 20.0     & 42.2 & 39.6 & 34.8 & 30.8 & 26.2 & 22.4 & 18.0 & 16.6 & 15.4 \\ \cline{2-12} 
               \rowcolor{tabgray}                     \multirow{-9}{*}{DRT + MME (Gaussian)}            & 20.0                 & 30.0     & 33.8 & 30.2 & 26.8 & 22.6 & 18.6 & 16.8 & 15.0 & 12.8 & 11.4 \\ \hline
MME (SmoothAdv)                        & -                    & -        & 48.2 & 43.7 & 40.1 & 35.4 & 31.3 & 26.2 & 22.6 & 19.5 & 16.2 \\ \hline
      \rowcolor{tabgray}   & 0.2                  & 5.0      & 48.2 & 43.9 & 40.1 & 35.4 & 31.5 & 26.7 & 22.9 & 19.8 & 16.8 \\ \cline{2-12} 
             \rowcolor{tabgray}                                  & 0.5                  & 5.0      & 48.1 & 43.8 & 40.3 & 35.7 & 31.8 & 26.9 & 23.1 & 20.1 & 17.5 \\ \cline{2-12} 
            \rowcolor{tabgray}                                   & 1.0                  & 5.0      & 46.2 & 43.4 & \bf 40.8 & \bf 37.0 & \bf 34.2 & 30.0 & \bf 26.8 & 23.8 & 20.1 \\ \cline{2-12} 
             \rowcolor{tabgray}                        \multirow{-4}{*}{DRT + MME (SmoothAdv)}          & 1.5                  & 5.0      & 47.8 & 43.4 & 39.5 & 35.4 & 31.6 & 26.7 & 23.1 & 20.4 & 18.1 \\ \hline
WE (Gaussian)                          & -                    & -        & \bf 50.4 & 44.1 & 37.5 & 30.9 & 24.2 & 19.2 & 15.9 & 11.8 & 8.9  \\ \hline
      \rowcolor{tabgray}   & 0.5                  & 5.0      & 49.5 & 44.3 & 37.8 & 31.8 & 25.6 & 22.5 & 18.2 & 14.4 & 12.3 \\ \cline{2-12} 
            \rowcolor{tabgray}                                   & 1.0                  & 5.0      & 49.8 & \bf 44.4 & 39.1 & 31.7 & 25.6 & 22.8 & 18.4 & 15.1 & 12.1 \\ \cline{2-12} 
           \rowcolor{tabgray}                                    & 1.5                  & 5.0      & 48.2 & 42.5 & 36.6 & 30.4 & 26.1 & 22.1 & 18.2 & 15.7 & 12.6 \\ \cline{2-12} 
          \rowcolor{tabgray}                                     &  & 0.5      & 44.6 & 38.6 & 34.7 & 29.1 & 25.8 & 21.8 & 19.6 & 17.1 & 15.6 \\
            \rowcolor{tabgray}                                   &                      & 2.5      & 44.8 & 39.6 & 35.4 & 31.0 & 27.9 & 23.4 & 20.6 & 18.1 & 16.4 \\
            \rowcolor{tabgray}                                   & \multirow{-3}{*}{5.0}                     & 10.0     & 45.4 & 40.3 & 36.8 & 30.4 & 26.2 & 21.8 & 19.1 & 15.8 & 13.6 \\ \cline{2-12} 
            \rowcolor{tabgray}                                   & 10.0                 & 20.0     & 44.5 & 40.8 & 36.2 & 31.3 & 27.4 & 21.2 & 18.9 & 17.2 & 13.5 \\ \cline{2-12} 
             \rowcolor{tabgray}                                  & 15.0                 & 20.0     & 42.2 & 39.7 & 34.8 & 30.8 & 26.1 & 22.4 & 18.0 & 16.7 & 15.4 \\ \cline{2-12} 
             \rowcolor{tabgray}                           \multirow{-9}{*}{DRT + WE (Gaussian)}         & 20.0                 & 30.0     & 33.8 & 30.4 & 26.8 & 22.8 & 18.6 & 16.9 & 15.0 & 12.7 & 11.2 \\ \hline
WE (SmoothAdv)                         & -                    & -        & 48.2 & 43.7 & 40.2 & 35.4 & 31.5 & 26.2 & 22.7 & 19.6 & 16.0 \\ \hline
       \rowcolor{tabgray}  & 0.2                  & 5.0      & 48.2 & 43.8 & 40.2 & 35.4 & 31.5 & 26.8 & 23.0 & 19.9 & 16.7 \\ \cline{2-12} 
             \rowcolor{tabgray}                                  & 0.5                  & 5.0      & 48.2 & 43.8 & 40.5 & 35.7 & 31.9 & 26.8 & 23.3 & 20.2 & 17.5 \\ \cline{2-12} 
             \rowcolor{tabgray}                                  & 1.0                  & 5.0      & 46.2 & 43.4 & 40.6 & \bf 37.0 & \bf 34.2 & \bf 30.1 & \bf 26.8 & \bf 23.9 & \bf 20.3 \\ \cline{2-12} 
              \rowcolor{tabgray}                           \multirow{-4}{*}{DRT + WE (SmoothAdv)}       & 1.5                  & 5.0      & 47.8 & 43.4 & 39.6 & 35.4 & 31.4 & 26.7 & 23.0 & 20.4 & 18.1 \\ \bottomrule
\end{tabular}}
\label{tab:bigcifartable1.00}
\end{table}

\textbf{Average Certified Radius}. We report the Average Certified Radius (ACR)~\citep{zhai2019macer}: \text{ACR} = $\frac{1}{|\mathcal{S}_{\text{test}}|}\sum_{(x,y)\in\mathcal{S}_{\text{test}}} R(x,y)$, where $\mathcal{S}_{\text{test}}$ refers to the test set and $R(x, y)$ the certifed radius on testing sample $(x, y)$. We evaluate ACR of our DRT-trained ensemble trained with $\sigma \in \{0.25, 0.5, 1.0\}$ smoothing parameter and compare it with other baselines. Results are shown in \Cref{tab:acr-cifar10}. 

Results shows that, DRT-trained ensemble has the highest ACR on almost all the settings. Especially on $\sigma = 1.00$, our improvement is significant. 
\begin{table}[!htbp]
\centering
\caption{Average Certified Radius (ACR) of DRT-trained ensemble trained with different smoothing parameter $\sigma \in \{0.25, 0.50, 1.00\}$ on CIFAR-10 dataset, compared with other baselines. The grey rows present the performance of the proposed DRT approach. The brackets shows the base models we use.}
\begin{tabular}{l|c|c|c}
\toprule
Radius $r$                 & $\sigma=0.25$  & $\sigma=0.50$  & $\sigma=1.00$  \\ \hline
Gaussian                   & 0.484          & 0.595          & 0.559          \\
SmoothAdv                  & 0.539          & 0.662          & 0.730          \\
MACER                      & \textbf{0.556} & 0.726          & 0.792          \\ \hline
MME / WE (Gaussian)        & 0.513          & 0.621          & 0.579          \\
DRT + MME / WE (Gaussian)  & 0.551          & 0.687          & 0.744          \\
MME / WE (SmoothAdv)       & 0.542          & 0.692          & 0.689          \\
DRT + MME / WE (SmoothAdv) & 0.545          & \textbf{0.760} & \textbf{0.868} \\ \bottomrule
\end{tabular}
\label{tab:acr-cifar10}
\end{table}

\textbf{Effects of $\rho_1$ and $\rho_2$.} 
We study the DRT hyper-parameter $\rho_1$ and $\rho_2$ corresponding to different smoothing parameters $\sigma \in \{0.25, 0.5, 1.0\}$ and put the detailed results in \Cref{tab:bigcifartable0.25,tab:bigcifartable0.50,tab:bigcifartable1.00}. We bold the numbers with the highest certified accuracy on each radius $r$. The results show similar conclusion to our understanding from MNIST. 

\textbf{Efficiency Analysis.} We also use the \emph{execution time per mini-batch} as our efficiency criterion. For CIFAR-10 with batch size equals to $256$, DRT with the Gaussian smoothing base model requires \textcolor{red}{3.82s} to finish one mini-batch training to achieve the competitive results to $10$-step PGD attack based SmoothAdv method which requires \textcolor{red}{6.39s}. All the models are trained in parallel on 4 NVIDIA GeForce GTX 1080 Ti GPUs.

\subsection{ImageNet} 

For ImageNet, we utilize ResNet-50 architecture and train each base model for $90$ epochs using SGD-momentum optimizer. The initial learning rate $\alpha$ is set to $0.1$. During training, the learning rate is decayed for every $30$-epochs with decay ratio $\gamma = 0.1$. We tried different Gaussian smoothing parameter $\sigma \in \{0.50, 1.00\}$, and consider the best hyper-parameter configuration for each $\sigma$. Then, we use DRT to finetune base models with the learning rate $\alpha = 5\times 10^{-3}$ for another $90$ epochs.
Due to the consideration of achieving high certified accuracy on large radii, we choose large DRT training hyper-parameter $\rho_1$ and $\rho_2$ in practice, which lead to relatively low benign accuracy. 

\section{Ablation studies}
\label{adx-subsec:ablation}

\subsection{The Effects of \LHSloss and \RHSloss}

\label{adx-subsec:seperate}

To explore the effects of individual Gradient Diversity and Confidence Margin Losses in DRT, we set $\rho_1$ or $\rho_2$ to 0 and tune the other for evaluation on MNIST and CIFAR-10. The results are shown in Table~\ref{tab:gdlvscml3} and~\ref{tab:gdlvscml2}. 
We observe that both GD and CM losses have positive effects on improving the certified accuracy while GD plays a major role on larger radii. By combining these two regularization losses together in DRT, the ensemble model achieves the highest certified accuracy under all radii.

\begin{table*}[!htbp]
\centering
\caption{Certified accuracy achieved by training with GD Loss (GDL) or Confidence Margin Loss (CML) only on MNIST dataset.}
\scalebox{0.76}{
\begin{tabular}{c|c|c|c|c|c|c|c|c|c|c|c}
\toprule
Radius $r$ & $0.00$ & $0.25$ & $0.50$ & $0.75$ & $1.00$ & $1.25$ & $1.50$ & $1.75$ & $2.00$ & $2.25$ & $2.50$ \\ \hline
MME (Gaussian) & 99.2                         & 98.4                         & 96.8                         & 94.9                         & 90.5                         & 84.3                         & 69.8      & 48.8 & 34.7 & 23.4 & 12.7                       \\
GDL + MME (Gaussian) & 99.2                         & 98.4                         & 96.9                         & 95.3                         & 92.3                         & 86.2                         & 76.0      & \bf 60.2 & 43.3 & 35.5 &  28.7                       \\
CML + MME (Gaussian) & 99.3                         & 98.4                         & 97.0                         & 95.0                         & 90.8                         & 84.0                         & 71.1      & 50.0 & 36.7 & 24.5 & 13.7                       \\
DRT + 
MME (Gaussian) & \bf 99.5                         & \bf 98.6                         & \bf 97.5                            & \bf 95.5                         & \bf 92.6                         & \bf 86.8                         & \bf 76.5       & \bf 60.2 & \bf 43.9 & \bf 36.0 & \bf 29.1 \\ \hline\hline 
WE (Gaussian)  & 99.2                         & 98.4                         & 96.9                         & 94.9                         & 90.6                         & 84.5                         & 70.4 & 49.0 & 35.2 & 23.7 & 12.9   \\
GDL + WE (Gaussian) & 99.3                         & 98.5                         & 97.1                         & 95.3                         & 92.3                         & 86.3                         & 76.3      &  59.8 & 43.4 & 35.1 &  \bf 29.0                       \\
CML + WE (Gaussian) & 99.3                         & 98.4                         & 97.0                         & 95.0                         & 90.8                         & 84.1                         & 71.1      & 50.3 & 37.0 & 24.6 & 13.7                       \\
DRT + WE (Gaussian)  & \bf 99.5                         & \bf 98.6                         & \bf 97.4                         & \bf 95.6                         & \bf 92.6                         & \bf 86.7                         & \bf 76.7         & \bf 60.2 & \bf 43.9 & \bf 35.8 & \bf 29.0                    \\
\bottomrule
\end{tabular}}
\label{tab:gdlvscml3}
\end{table*}

\begin{table*}[!t]
\centering
\caption{Certified accuracy achieved by training with GD Loss (GDL) or Confidence Margin Loss (CML) only on CIFAR-10 dataset.}
\scalebox{0.83}{
\begin{tabular}{c|c|c|c|c|c|c|c|c|c}
\toprule
Radius $r$ & $0.00$ & $0.25$ & $0.50$ & $0.75$ & $1.00$ & $1.25$ & $1.50$ & $1.75$ & $2.00$ \\ \hline
MME (Gaussian)   & 80.8   & 68.2   & 53.4   & 38.4   & 29.0   & 19.6   & 15.6   & 11.6   & 8.8    \\
GDL + MME (Gaussian)   & 81.0	& 69.0	& 55.6	& 41.9	& 30.4	& 24.8	& 20.1	& 16.9	& 14.7 \\
CML + MME (Gaussian)       & 81.2	& 69.4	& 54.4	& 39.6	& 29.2	& 21.6	& 17.0	& 13.1	& 12.8   \\
DRT + MME (Gaussian)      & \bf 81.4 & \bf 70.4	& \bf 57.8	& \bf 43.8	& \bf 34.4	& \bf 29.6	& \bf 24.9	& \bf 20.9	& \bf 16.6   \\ \hline\hline 
WE (Gaussian)   & 80.8   & 68.4   & 53.6   & 38.4   & 29.2   & 19.7   & 15.9   & 11.8   & 8.9    \\
GDL + WE (Gaussian)   & 81.0	& 69.1	& 55.6	& 41.8	& 30.6	& 25.2	& 20.2	& 16.9	& 14.9 \\
CML + WE (Gaussian)       & 81.1	& 69.4	& 54.6	& 39.7	& 29.4	& 21.7	& 17.2	& 13.2	& 12.8   \\
DRT + WE (Gaussian)      & \bf 81.5 & \bf 70.4	& \bf 57.9	& \bf 44.0	& \bf 34.2	& \bf 29.6	& \bf 24.9	& \bf 20.8	& \bf 16.4   \\
\bottomrule
\end{tabular}}
\label{tab:gdlvscml2}
\end{table*}

\subsection{Certified Robustness of single base model within DRT-trained ensemble}
\label{adx-subsec:singlemodels}
We also conduct ablation study on how the single base models' certified accuracy can be improved after applying DRT to the whole ensemble for both MNIST and CIFAR-10 datasets. Results are shown in in \Cref{tab:singlemodel-mnist} and \ref{tab:singlemodel-cifar}. We are surprised to find that the single base model within our DRT-trained ensemble are more robust compared to single base model within other baseline ensembles. Also, integrating them together could achieve higher robustness. 

\begin{table*}[!htbp]
\centering
\caption{Certified accuracy of single base model within DRT-trained ensemble on MNIST dataset.}
\scalebox{0.76}{
\begin{tabular}{c|c|c|c|c|c|c|c|c|c|c|c}
\toprule
Radius $r$ & $0.00$ & $0.25$ & $0.50$ & $0.75$ & $1.00$ & $1.25$ & $1.50$ & $1.75$ & $2.00$ & $2.25$ & $2.50$ \\ \hline
Single (Gaussian)               & 99.1                         & 97.9                         & 96.6                         & 94.7                         & 90.0                         & 83.0                         & 68.2  & 46.6 & 33.0 & 20.5 & 11.5 \\
DRT Single (Gaussian) & 99.0                         & \bf 98.6                         & 97.2                         & 95.4                         & 92.0                         & 85.6                         & 74.9      & 59.8 & 43.4 & 35.2 & 28.6                       \\
DRT + MME (Gaussian) & \bf 99.5                         & \bf 98.6                         & \bf 97.5                         & 95.5                         & \bf 92.6                         & \bf 86.8                         & 76.5      & \bf 60.2 & \bf 43.9 & \bf 36.0 &  \bf 29.1                       \\
DRT + WE (Gaussian) & \bf 99.5                         & \bf 98.6                         & 97.4                         & \bf 95.6                         & \bf 92.6                         & 86.7                         & \bf 76.7      & \bf 60.2 & \bf 43.9 & 35.8 & 29.0                       \\
 \hline\hline 
 Single (SmoothAdv) & 99.1                         & 98.4                         & 97.0                         & 96.3                         & 93.0                         & 87.7                         & 80.2   & 66.3 & 43.2 & 34.3 & 24.0                          \\
DRT Single (SmoothAdv)  & \bf 99.2                         & \bf 98.4                         & \bf 97.6                         & 96.6                         & 92.9                         & 88.1                         & 80.4 & 68.0 & 46.4 & 39.2 & 34.1   \\
DRT + MME (SmoothAdv) & \bf 99.2                         & \bf 98.4                         & \bf 97.6                         & \bf 96.7                         & 93.1                         & \bf 88.5                         & 83.2      &  \bf 68.9 & 48.2 & \bf 40.3 &  34.7                       \\
DRT + WE (SmoothAdv) & 99.1                         & \bf 98.4                         & \bf 97.6                         & \bf 96.7                         & \bf 93.4                         & \bf 88.5                         & \bf 83.3      & \bf 69.6 & \bf 48.3 & 40.2 & \bf 34.8                       \\
\bottomrule
\end{tabular}}
\label{tab:singlemodel-mnist}
\end{table*}

\begin{table*}[!t]
\centering
\caption{Certified accuracy of single base model within DRT-trained ensemble on CIFAR-10 dataset.}
\scalebox{0.83}{
\begin{tabular}{c|c|c|c|c|c|c|c|c|c}
\toprule
Radius $r$ & $0.00$ & $0.25$ & $0.50$ & $0.75$ & $1.00$ & $1.25$ & $1.50$ & $1.75$ & $2.00$ \\ \hline
 Single (Gaussian)  & 78.9                         & 64.4                         & 47.4                         & 33.7                         & 23.1                         & 18.3                         & 13.6  & 10.5 & 7.3         \\
DRT Single (Gaussian)   & 81.4   & 69.8   & 56.2   & 42.5   & 33.6   & 27.6   & 24.2   & 20.4   & 15.4    \\
DRT + MME (Gaussian)   & 81.4  	& \bf 70.4	& 57.8	& 43.8	& \bf 34.4	& \bf 29.6	& \bf 24.9	& \bf 20.9	& \bf 16.6 \\
DRT + WE (Gaussian)       & \bf 81.5	& \bf 70.4	& \bf 57.9	& \bf 44.0	& 34.2	& \bf 29.6	& \bf 24.9	& 20.8	& 16.4   \\ \hline\hline 
 Single (SmoothAdv) & 68.9                         & 61.0                         & 54.4                         & 45.7                         & 34.8                         & 28.5                         & 21.9   & 18.2 & 15.7 \\
DRT Single (SmoothAdv)   & 72.4   & 66.8   & 57.8   & 48.2   & 38.1   & 33.4   & 28.6   & 22.2   & 19.6    \\
DRT + MME (SmoothAdv)   & \bf 72.6	& \bf 67.2	& \bf 60.2	& 50.4	& 39.4	& 35.8	& \bf 30.4	& 24.0	& 20.1 \\
DRT + WE (SmoothAdv)       & \bf 72.6	& 67.0	& \bf 60.2	& \bf 50.5	& \bf 39.5	& \bf 36.0	& 30.3	& \bf 24.1	& \bf 20.3   \\
\bottomrule
\end{tabular}}
\label{tab:singlemodel-cifar}
\end{table*}

\subsection{Optimizing the Weights of DRT-trained Ensemble}
\label{adx-subsec:optimize-weights}

While we adapt average weights in our WE ensemble protocol in our experiments, we are also interested in how tuning the optimal weights could further improve the certified accuracy of our DRT-trained ensemble. We conduct this ablation study on both MNIST and CIFAR-10 datasets by grid-searching all the possible weights combination with step size as $0.1$. Results are shown in \Cref{tab:learn-weights-mnist} and \ref{tab:learn-weights-cifar}. (\textbf{AE} here refers to the average ensemble protocol and \textbf{WE} the weighted ensemble protocol by adapting the tuned optimal weights)

We can see that, by learning the optimal weights, the certified accuracy could be only slightly improved compared to the average weights setting, which indicates that, average weights can be a good choice in practice.

\begin{table*}[!htbp]
\centering
\caption{Comparison of the certified accuracy between Average Ensemble (AE) protocol and Weighted Ensemble (WE) protocol on MNIST dataset. Cells with \textbf{bold} numbers indicate learning optimal weights could achieve higher certified accuracy on corresponding radius $r$. }
\scalebox{0.76}{
\begin{tabular}{c|c|c|c|c|c|c|c|c|c|c|c}
\toprule
Radius $r$ & $0.00$ & $0.25$ & $0.50$ & $0.75$ & $1.00$ & $1.25$ & $1.50$ & $1.75$ & $2.00$ & $2.25$ & $2.50$ \\ \hline
DRT + AE (Gaussian) & 99.5                         &  98.6                         & 97.4                         & 95.6                         & 92.6                         & 86.7                         & 76.7      & 60.2 & 43.9 & 35.8 & 29.0                       \\
DRT + WE (Gaussian) &  99.5                         &  98.6                         & \bf 97.6                         & 95.6                         & \bf 92.7                         & 86.8                         & 76.7      & \bf 60.3 & \bf 44.0 & 36.0 & 29.3                       \\
 \hline\hline 
DRT + AE (SmoothAdv) & 99.1                         &  98.4                         &  97.6                         &  96.7                         &  93.4                         &  88.5                         &  83.3      &  69.6 &  48.3 & 40.2 &  34.8                       \\
DRT + WE (SmoothAdv) & \bf 99.1                         & 98.4                         & 97.6                         & \bf 96.8                         & \bf 93.5                         & 88.5                         & 83.3      &  \bf 69.7 & \bf 48.5 & 40.2 &  34.8                       \\
\bottomrule
\end{tabular}}
\label{tab:learn-weights-mnist}
\end{table*}

\begin{table*}[!t]
\centering
\caption{Comparison of the certified accuracy between Average Ensemble (AE) protocol and Weighted Ensemble (WE) protocol on CIFAR-10 dataset. Cells with \textbf{bold} numbers indicate learning optimal weights could achieve higher certified accuracy on corresponding radius $r$.}
\scalebox{0.83}{
\begin{tabular}{c|c|c|c|c|c|c|c|c|c}
\toprule
Radius $r$ & $0.00$ & $0.25$ & $0.50$ & $0.75$ & $1.00$ & $1.25$ & $1.50$ & $1.75$ & $2.00$ \\ \hline
DRT + AE (Gaussian)       & 81.5	& 70.4	& 57.9	& 44.0	& 34.2	& 29.6	&  24.9	& 20.8	& 16.4   \\
DRT + WE (Gaussian)   & 81.5  	& 70.4	& 57.9	& 44.0	& \bf 34.3	& 29.6	& \bf 25.0	& \bf 20.9	& \bf 16.5 \\
\hline\hline 
DRT + AE (SmoothAdv)       & 72.6	& 67.0	& 60.2	& 50.5	& 39.5	& 36.0	& 30.3	& 24.1	& 20.3   \\
DRT + WE (SmoothAdv)   & 72.6	& \bf 67.1	& 60.2	& 50.5	& 39.5	& \bf 36.1	& 30.3	& 24.1	& \bf 20.4 \\
\bottomrule
\end{tabular}}
\label{tab:learn-weights-cifar}
\end{table*}

\subsection{Comparison with Other Gradient Diversity Regularizers}

\label{adx-subsec:adpgal}

We notice that out of the \textit{certifiably} robust ensemble field, there exist two representatives of  gradient diversity promoting regularizers:  ADP~\citep{pang2019improving} and  GAL~\citep{kariyappa2019improving}. 
They achieved notable improvements on \textit{empirical} ensemble robustness. 
For an ensemble consisting of base models $\{F_i\}_{i=1}^N$ and input $\vx$ and ground truth label $y$, the ADP regularizer is defined as $$\mathcal{L}_{\text{ADP}}(\vx, y)=\alpha \cdot \sum_{i=1}^{N}H(\mathrm{mean}(\{f_i(\vx)\})) + \beta \cdot \log(\mathbb{ED})$$ where $H(\cdot)$ refers to the Shannon Entropy Loss function and $\mathbb{ED}$ the square of the spanned volume of base models' logit vectors.

GAL regularizer minimizes the cosine similarity value between base models' loss gradient vectors, which is defined as: \\ $$\mathcal{L}_{\text{GAL}}(\vx, y) = \log\left(\sum_{1 \leq i < j \leq N} \exp\left(\cos\langle\nabla_{\vx} \ell_{F_i}, \nabla_{\vx} \ell_{F_j}\rangle\right)\right).$$

Under the smoothed ensemble training setting, the final training loss is represented by
$$\mathcal{L}_{\text{train}}(\vx, y) =  \sum_{i\in [N]} \mathcal{L}_{\mathrm{std}}(\vx + \rvepsilon, y)_i + \{\mathcal{L}_{\text{ADP}}(\vx + \rvepsilon, y)\text{ or }\mathcal{L}_{\text{GAL}}(\vx + \rvepsilon, y)\}$$
where we consider standard training loss $\mathcal{L}_{\mathrm{std}}(\vx_0+\rvepsilon, y_0)_i$ of each base model $F_i$ to be the standard cross-entropy loss.

Table~\ref{tab:adpgal} shows the certified accuracy of \{ADP, GAL, DRT\}-trained ensemble under different radii with WE protocol on MNIST and CIFAR-10 dataset. We  notice that DRT outperforms both  ADP and GAL significantly in terms of the \textit{certified} accuracy on different datasets.
\begin{table}
\small
\centering
\caption{{Certified accuracy} of \{ADP, GAL, DRT\}-based Gaussian smoothed ensemble under different radii with WE protocol.}
\begin{tabular}{l|c|c|c|c|c|c|c|c|c|c|c}
\toprule
    \bf MNIST $r$ & $0.00$ & $0.25$ & $0.50$ & $0.75$ & $1.00$ & $1.25$ & $1.50$ & $1.75$ & $2.00$ & $2.25$ & $2.50$ \\
\midrule
    ADP & \bf 99.5 & 98.2 & 97.2 & 95.2 & 92.2 & 85.8 & 73.4 & 53.2 & 36.9 & 24.7 & 13.3\\
    GAL &  \bf 99.5 & 98.3 & 97.2 & 95.1 & 92.4 & 86.1 & 73.2 & 54.4 & 36.2 & 24.7 & 13.9 \\
    \rowcolor{tabgray} DRT & \bf 99.5 & \bf 98.6 & \bf 97.4 & \bf 95.6 & \bf 92.6 & \bf 86.7 & \bf 76.7 & \bf 60.2 & \bf 43.9 & \bf 35.8 & \bf 29.0\\
\bottomrule
\end{tabular}
\vspace{1em}
\begin{tabular}{l|c|c|c|c|c|c|c|c|c}
\toprule
    \bf CIFAR-10 $r$ & $0.00$ & $0.25$ & $0.50$ & $0.75$ & $1.00$ & $1.25$ & $1.50$ & $1.75$ & $2.00$ \\
\midrule

    ADP & \bf 83.0 & 68.0 & 52.2 & 38.2 & 28.8 & 20.0 & 16.8 & 14.2 & 11.0\\
    GAL & 82.2 & 67.6 & 53.6 & 38.8 & 27.6 & 20.2 & 15.4 & 13.6 & 10.6 \\
    \rowcolor{tabgray} DRT &  81.5 & \bf 70.4 & \bf 57.9 & \bf 44.0 & \bf 34.2 & \bf 29.6 & \bf 24.9 & \bf 20.8 & \bf 16.4 \\
\bottomrule
\end{tabular}
\label{tab:adpgal}
\end{table}



\end{document}